%% file: main.tex
\documentclass{article}

\usepackage{microtype}
\usepackage{graphicx}
\usepackage{subcaption}
\usepackage{booktabs} % for professional tables
\usepackage{wrapfig}

\usepackage{hyperref}

\usepackage[preprint]{icml2026}

\usepackage{amsmath}
\usepackage{amssymb}
\usepackage{mathtools}
\usepackage{amsthm}
\usepackage{multirow}

\usepackage[capitalize,noabbrev]{cleveref}

\theoremstyle{plain}

\theoremstyle{definition}

\theoremstyle{remark}

\newcommand{\gray}[1]{\textcolor{gray}{#1}}

\usepackage{upgreek}
\usepackage{enumitem}
\usepackage[table]{xcolor}
\usepackage[most]{tcolorbox}
\usepackage{tikz}
\usepackage{multirow}

\definecolor{ourcell}{HTML}{E5E5E5}  % light pink: FCEBF7
\definecolor{colorcommentbg}{HTML}{F8F2F5}  % comment background
\definecolor{colorcommentframe}{HTML}{62AED4}  % comment title background 
\definecolor{colorcommentborderline}{HTML}{2A60D9}  % pink: AD2B75  % comment borderlines 
\definecolor{maintext}{HTML}{000000}
\definecolor{commenttext}{HTML}{23579A}

\usepackage{pifont}
\newcommand{\cmark}{\ding{51}} % 对勾
\newcommand{\xmark}{\ding{55}} % 叉

\newcommand{\typetag}[2]{%
\tikz[baseline=(X.base)]\node[
    draw=none,
    fill=#1,
    rounded corners=2pt,
    minimum width=1.8em,
    minimum height=1.0em,
    inner sep=0pt,
    align=center
] (X) {\footnotesize\textbf{\strut #2}};%
}

\definecolor{typeQ}{HTML}{A4CA91}
\newcommand{\TypeQ}[1][]{%
  \if\relax\detokenize{#1}\relax
    \typetag{typeQ}{Q}%
  \else
    \typetag{typeQ}{Q$_\mathrm{#1}$}%
  \fi
}

\definecolor{typeP}{HTML}{B4D2EC}

\newcommand{\TypeP}[1][]{%
  \if\relax\detokenize{#1}\relax
    \typetag{typeP}{P}%
  \else
    \typetag{typeP}{P$_\mathrm{#1}$}%
  \fi
}

\definecolor{typeM}{HTML}{CBB697}
\newcommand{\TypeM}[1][]{%
  \if\relax\detokenize{#1}\relax
    \typetag{typeM}{M}%
  \else
    \typetag{typeM}{M$_\mathrm{#1}$}%
  \fi
}

\definecolor{typeL}{HTML}{E7A431}
\newcommand{\TypeL}[1][]{%
  \if\relax\detokenize{#1}\relax
    \typetag{typeL}{L}%
  \else
    \typetag{typeL}{L$_\mathrm{#1}$}%
  \fi
}

\newdimen\RowHt
\newcommand{\vbar}[1]{%
  \multirow[c]{#1}{*}{\rule{0.6pt}{\dimexpr #1\RowHt\relax}}%
}

\newcounter{takeawaycounter}

\newenvironment{takeaway}{%
    \refstepcounter{takeawaycounter}%
    \begin{tcolorbox}[
        fonttitle=\bfseries,
        enhanced jigsaw,
        colback={white},
        coltext=maintext,
        boxrule=0pt,
        arc=0pt,
        outer arc=0pt,
        opacityframe=0,
        left=1pt,
        right=1pt,
        top=1pt,
        bottom=2pt,
        boxsep=2pt,
        borderline north={1pt}{0pt}{colorcommentborderline},
        borderline south={1pt}{0pt}{colorcommentborderline}
    ]
    {\color{colorcommentborderline}\textbf{Takeaway \thetakeawaycounter:}}%
}{%
    \end{tcolorbox}
}

\usepackage[textsize=tiny]{todonotes}

\icmltitlerunning{Attribution-Guided and Coverage-Maximized Pruning for Structural MoE Compression}

\begin{document}

\twocolumn[
  \icmltitle{
  Attribution-Guided and Coverage-Maximized Pruning for \\ Structural MoE Compression
  }

  \begin{icmlauthorlist}
    \icmlauthor{Yifu Ding}{sklccse,cs,ntu}
    \icmlauthor{Jiacheng Wang}{sklccse,ai}
    \icmlauthor{Ge Yang}{sklccse,ai}
    \icmlauthor{Yongcheng Jing}{ntu} \\
    \icmlauthor{Jinyang Guo}{sklccse,ai}
    \icmlauthor{Xianglong Liu}{sklccse,cs}
    \icmlauthor{Dacheng Tao}{ntu}
  \end{icmlauthorlist}

  \icmlaffiliation{sklccse}{State Key Laboratory of Complex \& Critical Software Environment, Beihang University}
  \icmlaffiliation{cs}{School of Computer Science and Engineering, Beihang University}
  \icmlaffiliation{ai}{School of Artificial Intelligence, Beihang University}
  \icmlaffiliation{ntu}{Nanyang Technological University}

  \icmlcorrespondingauthor{Xianglong Liu}{xlliu@buaa.edu.cn}

  \icmlkeywords{Mixture-of-Experts, Model Compression, Structural Pruning, Multimodal Large Language Models}

  \vskip 0.3in
]



\printAffiliationsAndNotice{This work is completed during Yifu Ding's research attachment at NTU. }  % no special notice (required even if empty)

\begin{abstract}

\input{sections/0_abstract}

\end{abstract}

\input{sections/1_introduction}
\input{sections/2_related_work}

\input{sections/3_motivation}

\input{sections/4_method}

\input{sections/5_experiments}
\input{sections/6_conclusion}

\section*{Acknowledgement}
This work was supported by the National Natural Science Foundation of China (Nos. 62476018, 62525601), the Academic Excellence Foundation of BUAA for PhD Students, and the Fundamental Research Funds for the Central Universities. 

Dr Tao’s research is partially supported by NTU RSR and Start Up Grants. 

% \bibliography{example_paper}
\bibliography{references}
\bibliographystyle{icml2026}

%%%%%%%%%%%%%%%%%%%%%%%%%%%%%%%%%%%%%%%%%%%%%%%%%%%%%%%%%%%%%%%%%%%%%%%%%%%%%%%
%%%%%%%%%%%%%%%%%%%%%%%%%%%%%%%%%%%%%%%%%%%%%%%%%%%%%%%%%%%%%%%%%%%%%%%%%%%%%%%
% APPENDIX
%%%%%%%%%%%%%%%%%%%%%%%%%%%%%%%%%%%%%%%%%%%%%%%%%%%%%%%%%%%%%%%%%%%%%%%%%%%%%%%
%%%%%%%%%%%%%%%%%%%%%%%%%%%%%%%%%%%%%%%%%%%%%%%%%%%%%%%%%%%%%%%%%%%%%%%%%%%%%%%

\input{sections/10_appendix}

\end{document}

%% file: sections/0_abstract.tex
% Mixture-of-Experts (MoE) architectures enable massive model scaling, yet their deployment is hindered by heavy memory pressure and computational overhead. 
% MoE models scale efficiently but remain costly to deploy due to memory and inference overhead.
Mixture-of-Experts (MoE) models scale compute efficiently, yet they remain expensive to deploy due to substantial memory footprint and inference overhead. Prior methods mainly operate at the expert level, either removing whole experts or ranking experts by importance. However, such expert-wise decisions are too coarse to identify redundancy, and often misallocate pruning budgets and limits compression.
%  This issue worsens in large MoEs with dynamic routing and heterogeneous experts.
% \jy{Previous compression methods typically remove entire experts or measure the expert-wise importance, which are too coarse to identify inner expert redundancy, and leads to insufficient compression as the MoE scales to hundreds of experts due to the expert heterogeneous. } 
% \yc{Existing structural pruning failed to effectively and accurately estimate expert-wise importance when scales to large MoEs. }
% \ycj{Highlight the failure from expert scaling}
% Specifically, the massive search space makes traditional loss-ablation impractical, while highly dynamic activation patterns render router statistics unreliable indicators of actual expert contribution. 
% 前面先体现问题重点，大的MoE即专家比较多的case。考虑说一下这个是不是没人做过。最好再补半句深入说一下为啥这种cases pruning比较难。
% -> 之后说为了解决前面提到的为什么比较难这个事情，我们begin by分析
% -> motivated by这个分析中的哪里，我们设计了哪个方法模块
To alleviate this dilemma, we 
% for the first time 
observe that information in MoE experts is highly concentrated in a few channels, leaving substantial redundancy even in ``high importance'' experts. 
% \ycj{Is this the first time we discover this fact?}.  %., suggesting that even ``important`` experts can be aggressively pruned. 
% \ycj{Motivated by XXX}
% Motivated by this insight that even high-score experts exhibit significant channel-level redundancy, % 已合到上句
Accordingly, we propose a structural pruning framework tailored for MoEs, reforming the prune-ratio objective to maximizing channel-score coverage via an efficient attribution-based approximation. 
% Attribution-guided Expert-wise Slimming framework, which shifts the pruning objective from parameter ratios to maximizing saliency coverage supported by efficient attribution-based loss approximation to bypass expensive calibrations. % and utilizes alignment-aware redistribution to ensure the pruned model seamlessly integrates with hardware-efficient quantization.
% Therefore, we propose the Attribution-guided and Coverage-Maximized Expert-wise Pruning framework, which shifts the optimization objective 
% from allocating prune ratios 
% to maximizing saliency coverage, supported by efficient attribution-based loss approximation, and alignment-aware redistribution to integrated with quantization techniques. 
% Extensive evaluations on DeepSeek-MoE and Qwen-MoE demonstrate that our framework consistently outperforms state-of-the-art baselines in both commonsense and reasoning tasks.
Experiments on DeepSeek and Qwen MoEs retain accuracy under 50\% or 25\% pruning joinly with 4-bit quantization, reducing the memory footprint of Qwen3-30B-A3B by 5.27$\times$, and outperforming state-of-the-art baselines under diverse benchmarks.\footnote{Our code is available at \url{https://github.com/yifu-ding/MoE-Slimming}. } 
% achieving significant parameter reduction while maintaining competitive model accuracy. 

%% file: sections/1_introduction.tex
% \section{Introduction}

% \ycj{Narrow the analysis to MoEs with a large number of experts?}

% 背景: MoE Pruning (不要太长，短一些)

% 现有 Pruning Methods 类别 -> Two Limitations (loss ablation and gate weight) -> 引出总结的核心挑战，remain unexplored / barely studied

% 核心观察说成是我们第一个发现了导致上述挑战的原因？引出observation

% Motivated by XX discovery, in this paper, we strive to fill this gap 说明我们的目标是什么，要实现什么样的效果，这一段强调我们是第一个XXX。这段不要太长，是一个总说

% 说具体方法，一段写不下写两段

% Our contribution is therefore the first XXX

% 写出来实验的亮点：量大、效果好（提升%，相对/绝对提升）

\section{Introduction}
Mixture-of-Experts (MoE) architectures have become a dominant paradigm for scaling language models, offering high parameter capacity while maintaining manageable computation by activating only a subset of experts for each token~\cite{openmoe, qwen3technicalreport, deepseek-r1}. To effectively deploy large modern MoEs and accelerate the inference, structural pruning, which removes entire channels or experts to yield hardware-efficient dense smaller model, offers a promising solution~\cite{llm-pruner,disp-llm,an2024fluctuation,guo2024joint}. Quantization, which reduces model bit-widths, is another complementary efficiency approach~\cite{gong2025lowbit,lv2026llmcplus}.
% Yet, effectively slimming MoEs remains a significant challenge, due to their heterogeneous experts and unique sparse activations.
In contrast to dense models, which use a single FFN per-layer shared by all tokens, MoEs comprise multiple experts with token-dependent routing. Experts are activated at vastly different frequencies and exhibit non-uniform internal redundancy~\cite{mc-moe,zhang2024diversify}. 
Consequently, pruning decisions are tightly related to data-dependent activation. 
% However, existing methods that apply uniform pruning~\cite{moe-pruner} can over-prune critical experts while wasting budget on low-contribution experts. 

% \jy{Finding the best prune ratios across heterogeneous experts becomes more difficult as modern MoEs scale to hundreds of experts, compared to traditional MoE that only consists of few experts.} 
% \jy{Loss-based methods} ablates in expert-wise manner become computationally infeasible as the number of experts grows, while \jy{routing statistics} collecting router output and post-softmax probabilities are cheap, but only reflect expert selection frequency and how they aggregate, ignoring their actual contribution. 
% \jy{Moreover, both methods fail to profile expert internal redundancy, in other words, how much capacity can be safely removed based on expert heterogeneous.} 
% As a result, accurate and scalable capacity allocation across experts and channels in large MoEs remains underexplored. 

\input{figures/1-intro}
Allocating good prune ratios across heterogeneous experts becomes substantially harder as modern MoEs scale to hundreds of experts, compared to earlier MoEs with only a handful of experts. Expert-wise, loss-based ablations~\cite{zhang2024diversify,lu2024notall} require evaluating each expert separately, so the cost scales linearly with the number of experts and becomes impractical at scale~\cite{yang2024moei2,bai2025diep}. 
Routing statistics~\cite{he2025towards,Lee2024STUNSP,xie2024moepruner} are cheap to collect, but they only capture selection frequency and aggregation proportion, rather than the experts’ true contribution. Moreover, both methods make decisions at the expert level, treating each expert as a whole unit and failing to characterize its internal redundancy, namely \textit{how much capacity can be safely removed} under significant expert heterogeneity. As a result, accurate and scalable capacity allocation across experts in large MoEs remains underexplored.

In this paper, we rethink MoE structural pruning based on our observation that MoE information is highly concentrated in a small fraction of channels, making expert-level importance too coarse to capture internal redundancy. 
To the best of our knowledge, we are the first to show that even ``high-importance'' experts may not require large capacity. 
This motivates a score-coverage-maximized allocation that prioritizes high contributed structures and avoids wasting capacity on low-score tails. 

We propose \textbf{Attribution-guided and Coverage-Maximized Expert-wise Pruning}, a framework tailored for MoE slimming. As shown in~\cref{fig:intro}, instead of allocating prune ratios directly from expert-level importance, we maximize \textit{channel score coverage} under a global budget, which better aligns with the highly concentrated and unbalanced information distribution in modern MoEs. 

Our framework consists of three components, as shown in~\cref{fig:intro}: 
(1) \textit{Attribution-guided Loss Approximation} (ALA) efficiently estimates expert importance layerwisely, without exhaustive ablation. 
(2) \textit{Coverage-maximized Budget Allocation} (CBA) uses ALA scores and performs coverage-driven capacity allocation under a global budget, retaining high-contribution channels while pruning low-score tails. 
(3) \textit{Alignment-Aware Redistribution} (AAR) adjusts dimensions after the initial allocation to satisfy low-bit kernel constraints, ensuring seamless integration with quantized storage and efficient inference. 
% 原来的版本，太长。
% To achieve this goal, the proposed framework contains three components. First, our \textbf{Maximized Coverage Budget Allocation} (MCBA) transform the optimization problem from the best prune ratio for each expert to the maximized channel score coverage under the importance constraints, which consists of inter-layer and intra-layer stages that find the maximized cumulative saliency scores across all experts. 
% Second, since the expert-wise loss evaluation is time-consuming with large number of experts, we employ an \textbf{Attribution-Based Loss Approximation} (ABLA) to leverage the first-order Taylor surrogate to approximate the loss of ablating every expert, enabling estimating the impact across all experts simultaneously. 
% Third, to bridge the gap between theoretical sparsity and actual implementation via low-bit quantization, we further introduce an \textbf{Alignment-Aware Redistribution} (AAR) mechanism by applying Hamilton apportionment to adjust the matrices to satisfy hardware low-bit GEMM kernel constraints to seamlessly integrates to low-bit quantization techniques and enjoy the further compression and speedup. 

Our framework achieves impressive results on representative MoE architectures, including DeepSeek and Qwen MoEs, across diverse downstream benchmarks. 
On general knowledge benchmarks, it delivers over $5\times$ compression with an average accuracy drop of at most 1\%. On reasoning benchmarks, the compressed models consistently approach or even surpass the original counterpart across various models and tasks.
% Our framework achieves more than $5\times$ compression by combining structural pruning and quantization, while approaching or even surpassing the original model in accuracy. Specifically, on GSM8K, Qwen1.5-MoE-A2.7B retains 58.2 accuracy versus 61.5 for the unpruned model, and on MATH500, Qwen3-30B-A3B reaches 95.0 under 50\% sparsity, outperforming the original model. 
These results demonstrate the effectiveness of our fine-grained, expert-wise pruning framework and provide a practical path toward efficient MoE deployment.

% \ycj{Focus on what problem you are the first to address, rather than propose a XX method} 
The main contributions are summarized as follows:
\begin{itemize}
    \item We observe that MoE information is concentrated in a small fraction of channels, making expert-level importance too coarse to capture expert internal redundancy. 

    \item We are the first to introduce \emph{channel score coverage} as a pruning objective, reformulating capacity allocation as maximizing coverage under a global budget to avoid wasting capacity on low-score tails.

    \item We propose an \emph{attribution-guided loss approximation} to enable scalable importance expert estimation with $20\times$ fewer GPU hours, and \emph{alignment-aware redistribution} for satisfy kernel shape constraints, allowing kernel-friendly storage and efficient inference. 
    
    \item Experiments on DeepSeek and Qwen MoEs deliver over $5\times$ compression with strong accuracy, with under 1\% drop on general knowledge, and 94.5 on MATH500 for Qwen3-30B-A3B under aggressive 50\% pruning.  
\end{itemize}

%% file: figures/1-intro.tex
\begin{figure}[t]
 \begin{center}
\centerline{\includegraphics[width=0.96\columnwidth]{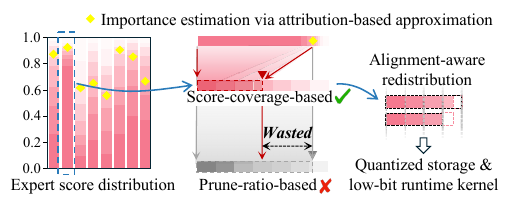}}
    \vspace{-0.05in}
    \caption{
    Overview of our pruning framework: estimating expert importance via an attribution-based approximation (left), maximizing score coverage to avoid wasting capacity (middle), and applying alignment-aware redistribution for compact storage and kernel-friendly low-bit inference (right). 
    }
    \label{fig:intro}
    \vspace{-0.4in}
  \end{center}
\end{figure}

%% file: sections/2_related_work.tex
\section{Related Works}
\label{sec:related_work}
Due to space limitations, a more comprehensive discussion is provided in Appendix~\cref{app:sec:related_work}. 

\noindent\textbf{MoE Compression.}
For efficient deployment of large MoEs, prior work explores: (i) \emph{Expert trimming} and \emph{expert skipping} to reduce runtime computation~\cite{liu2024eep,bai2025diep,lu2024notall,Chen2025EACMoEEA,huang2025modes}. (ii) \emph{Expert slimming} to compress each expert via pruning, quantization, or low-rank factorization~\cite{yang2024moei2,xie2024moepruner,Chen2025CollaborativeCF,guo2024joint,chen2024dbllm}.  Concurrent to this work, an anonymized submission (provided in the supplementary material) studies MoE pruning with a focus on structural pruning along the hidden dimension~\cite{anon_icml2026_hidden_intermediate_prune_distill}. (iii) \emph{Expert merging} to combine similar experts~\cite{zhao2025puzzlemoe,c-prune}. 
Most approaches operate at the granularity of whole experts or apply uniform compression for each expert, while only limited work explores heterogeneous compression across experts, e.g., different low-rank ranks~\cite{yang2024moei2} and mixed-precision bitwidth assignments~\cite{Chen2025CollaborativeCF}).

\paragraph{Expert Importance Estimation.}
A key challenge in MoE compression is estimating per-expert importance. Existing approaches commonly rely on router outputs (gate weights, token usage)~\cite{he2025towards,Lee2024STUNSP,mc-moe}, activation-based metrics~\cite{dong2025domain,zhao2025puzzlemoe}, performance-based criteria (e.g., loss or accuracy degradation under ablation)~\cite{liu2024eep,yang2024moei2}, or learnable scalars~\cite{bai2025diep}. 
%% 这一段：previous works limitations 
% However, these are often inadequate for MoE slimming. They primarily operate at the expert level and do not capture within-expert information concentration, making it difficult to quantify how much capacity can be safely removed. As a result, they are better suited for coarse expert trimming than for fine-grained, budgeted allocation in heterogeneous expert pruning. 
However, these signals are often inadequate for MoE slimming because they operate at the expert level and ignore expert internal information concentration, making them only suitable for expert trimming rather than fine-grained expert slimming. 
%% our methods 基于上述方法的缺点的解决思路：
Our approach advances prior works by replacing costly expert-wise ablation with efficient approximation, and further goes beyond ranking experts to channel-level budgets allocation via global score-coverage maximization.

%% file: sections/3_motivation.tex
\section{Pre-analysis: \emph{The Inherent Difficulty of Expert-Level Importance Estimation}}

MoEs sparsely route tokens across experts, and experts contribute unequally to final performance, making expert importance estimation a key problem in MoE compression. 
% Existing methods typically rely on heuristic proxies from router outputs or expert statistics, but these signals are often coarse and unreliable for guiding fine-grained slimming decisions. 
% In the following, we revisit common metrics and their limits, and motivate our approach by highlighting a fundamental mismatch between expert contribution and internal redundancy. 
However, existing methods typically rely on router outputs or expert statistics, which are often coarse and unreliable for fine-grained slimming allocations. Below, we revisit common metrics and their limitations, and motivate our approach by highlighting a fundamental mismatch between expert contribution and internal redundancy.

% \ycj{mention: will motivate the proposed methodology in Sect.~XX}

% \ycj{Titles of Sect. 3.1 \& 3.2 not consistent}
%可能没有那么整体感，我们后面meeting时候可以再讨论下。

\subsection{Limitation of Heuristic Metrics}
\label{sec:limitation-of-existing-metrics}

% \ycj{section title -> Inter-Expert Difficulty?}

Existing metrics suffer from two key limitations: (1) \textbf{router outputs} (e.g., routing weights or token usage) only quantify token engagement but do not indicate whether an expert’s output is beneficial or harmful; (2) \textbf{raw statistics} (e.g., weight, activation or gradients) exhibit layer-dependent magnitude across layers, and also poorly correlate with the actual contribution of experts within a layer.

\input{figures/1-router-NLL}

\paragraph{Routers can have wrong choices. }
\label{sec:router-can-have-wrong-choices}

Some prior works estimate expert importance using router outputs, such as post-softmax probabilities or the number of tokens routed to each expert. 
However, \cref{fig:observation2-a} shows that these routing statistics can be seriously misaligned with an expert's true contribution, measured by expert-wise ablated Negative Log-Likelihood (NLL). Concretely, we plot expert-wise ablated $\Delta$NLL (bars) alongside router probabilities and token usage on Qwen1.5-MoE-A2.7B: (a) plots the top 50 experts ranked by router weight and (b) ranked experts by token usage. Empirically, both router probabilities and token usage exhibit weak correlation with $\Delta$NLL. Highly prioritized or frequently activated experts can cause only a minor loss increase when removed (blue bars), and some even reduce the loss (orange bars below zero). This suggests that routing signals mainly reflect selection and how the experts' output are aggregated, rather than whether an expert is beneficial, and the selected expert can be noisy or even harmful. 

\begin{takeaway}
\label{tw:router}
Router-derived statistics (post-softmax weights, token usage) only reflect the engagement of experts, but do not tell the actual contribution of experts. 
\end{takeaway}

\input{figures/1-raw-stats}
\paragraph{Incomparability of raw statistics across layers or experts. }
\label{subsec:imcomparability_of_raw_statistics}
% - 展示不同层、不同 expert 的 activation norm, weight norm, gradient norm 分布的统计差异。
% - 引用一些理论 paper（比如关于深层网络梯度尺度、表示分布随深度变化的文献）
%   说明「跨层/跨 expert 直接比较这些量不合理」。
Beyond router signals, another common heuristic estimate expert importance from raw forward or backward statistics (e.g., weights, activations, or gradients). However, these quantities are not comparable across layers and often not informative across experts within a layer, making them unreliable as direct importance proxies. 
(1) \emph{Cross-layer magnitude bias. }
In \cref{fig:observation2}(a), the layer-wise mean$\pm$std of raw weights and activations exhibits a depth-dependent trend, whereas gradients decay with depth. Such behavior arises due to residual accumulation, normalization, and so on. In contrast, layerwise $\Delta$NLL (blue markers) follows a different pattern and does not align with any raw statistic, indicating that magnitudes are inherently layer-dependent and unsuitable for cross-layer comparison. 
(2) \emph{Intra-Layer non-correlation. }
\cref{fig:observation2}(b) shows a similar issue within a single layer: after sorting experts by expert-wise ablated $\Delta$NLL (bars), the corresponding weight, activation, and gradient statistics (mean$\pm$std) exhibit no meaningful relation to loss impact, failing to distinguish helpful experts. 
% demonstrates that raw statistics can be uninformative even within a single layer. We sort experts by expertwise-ablated $\Delta$NLL (bars), and find that the corresponding weight, activation, and gradient (mean$\pm$std) exhibit no meaningful relation with the loss impact. In other words, raw magnitudes fail to distinguish helpful from harmful experts, even when the comparison is restricted to experts within the same layer. 

\begin{takeaway}
\label{tw:raw-statistics}
Raw statistics (weights, activations, gradients) exhibit weak cross-layer and intra-layer correlation with actual loss when removing the expert, fail to reliably represent the expert importance. 
\end{takeaway}

\subsection{Mismatch between Redundancy and Contribution}
\label{subsec:importance_neq_redundancy}
Some prior work estimates expert importance by measuring the loss increase when an expert is entirely removed. While this yields an expert-wise ranking, it does not indicate how much capacity can be safely removed within each expert.

\paragraph{Visualization of channel redundancy. }
\label{subsec:channel_level_concentration} 
To examine how information is distributed inside each expert, in \cref{fig:expertwise_saliency_concentration}(a), we sort channels by their scores (see Appendix~\cref{app:sec:channel-score-metric}) in descending order, and plot the cumulative fraction of the score covered by the top-$k\%$ channels. 
The results reveal pronounced heterogeneity in intra-expert redundancy: for some experts, nearly 40\% channels conveys negligible information, whereas other experts exhibit much weaker concentration. 
% This indicates that internal redundancy is expert-specific, that cannot be captured by expert-level importance signals alone. 
This expert-specific redundancy cannot be captured by expert-level importance signals alone.

\input{figures/2_expretwise_saliency_concentration}

\paragraph{Contribution can be recovered by few channels. }
\label{sec:big-loss-can-be-retrieved}
Given the concentration patterns, whole-expert ablation becomes an overly coarse proxy for redundancy. 
\cref{fig:expertwise_saliency_concentration}(b) shows that even when removing all channels causes a noticeable $\Delta$NLL, the loss may drop rapidly as only a small fraction of channels is restored. 
% In many cases, retaining the top 20-30\% channels recovers most of the information at the layer's output. 
Therefore, whole-expert ablation mainly serve as a binary signal about whether an expert is important, rather than quantifying how much redundancy exists within the expert or how pruning budgets should be allocated across channels. 
% Moreover, when loss is measured at the model’s final output, the impact of removing a single expert can be partially compensated by downstream layers, making the ablation results less reliable. Meanwhile, the loss metric can be sensitive to implementation details such as the loss definition, tokenization, and calibration data. 

\begin{takeaway}
\label{tw:contribution-vs-redundancy}
Expert-wise ablation measures expert-level contribution ($\Delta$NLL), but do not reflect the internal redundancy: channel scores can be highly concentrated in a small fraction of channels. 
\end{takeaway}

In conclusion, there is still no accurate and scalable metric that quantifies such redundancy across layers and experts, which motivates a fine-grained pruning strategy to determine the actual capacity of each expert.

%% file: figures/1-router-NLL.tex
\begin{figure}[t]
  % \vskip 0.2in
  \begin{center}
    \centerline{\includegraphics[width=1.0\columnwidth]{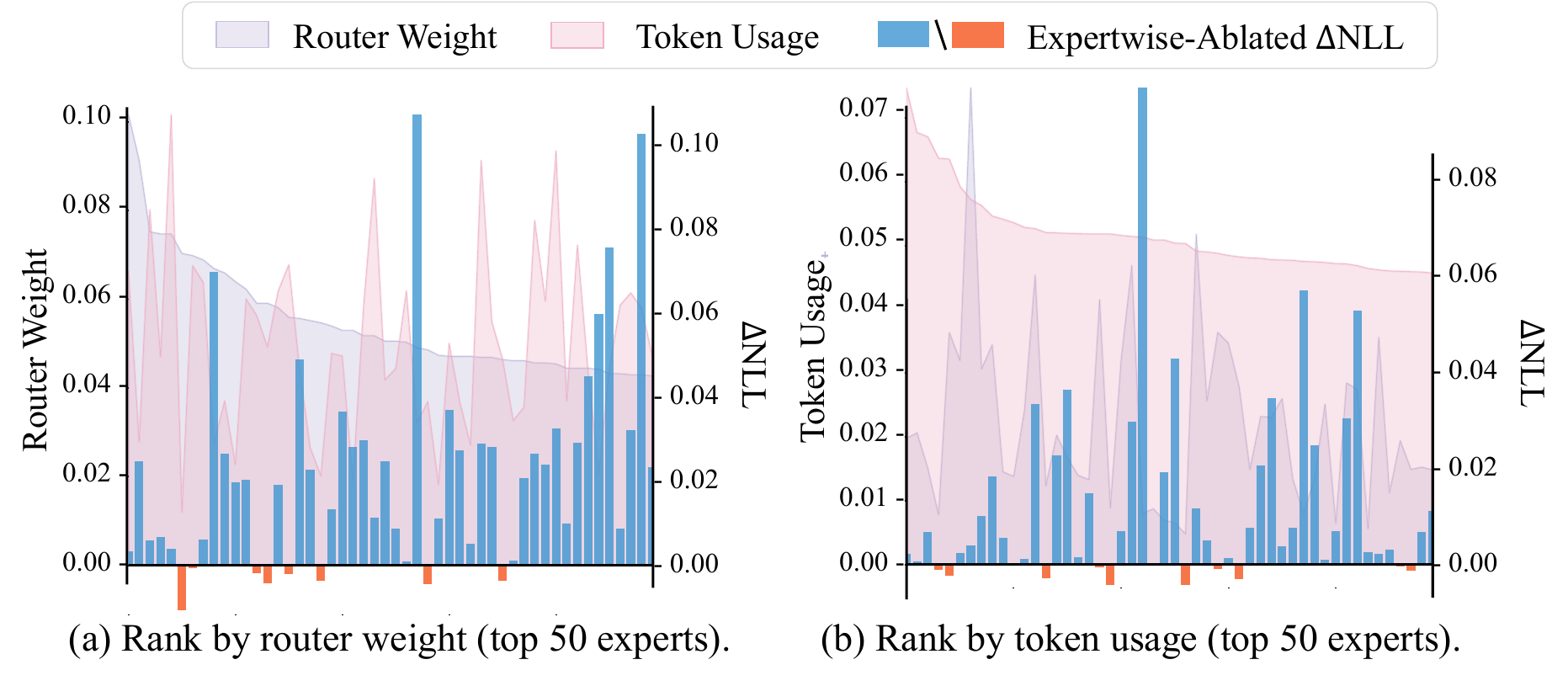}}
    \vspace{-0.06in}
    \caption{
Misalignment between router outputs and expert-wise ablated NLL. (a) and (b) rank the top 50 experts by router weight and token usage. The NLL (bars) demonstrates a weak correlation with router outputs. Notably, the orange bars highlight that even selected experts can provide negative contributions. 
% , since ablating them yields a lower NLL despite high routing priority and usage.
    }
    \label{fig:observation2-a}
    \vspace{-0.4in}
  \end{center}
\end{figure}

% \begin{figure}[ht]
%   % \vskip 0.2in
%   \begin{center}
%     \centerline{\includegraphics[width=1.0\columnwidth]{figures/images/observation2.pdf}}
%     \caption{
%     Layerwise statistics for different metrics. 
%     }
%     \label{fig:observation2}
%     \vspace{-0.2in}
%   \end{center}
% \end{figure}

% \begin{figure}[t]
%   % \vskip 0.2in
%   \begin{center}
%     \centerline{\includegraphics[width=0.95\columnwidth]{figures/images/observation2-a.png}}
%     \caption{
%     % Loss change $\Delta L$ (left) and loss $L$ (right) compared with router weights (top) and token usage (bottom). 
%     Router weights (top) and token usage (bottom) of each expert and the loss when trimming it entirely. 
%     }
%     \label{fig:observation2-a}
%     \vspace{-0.2in}
%   \end{center}
% \end{figure}

% \begin{figure}[t]
%   % \vskip 0.2in
%   \begin{center}
%     \centerline{\includegraphics[width=0.85\columnwidth]{figures/images/observation2-b.png}}
%     \caption{
%     Loss curves of $\Delta L$ under varying prune ratios for each expert.
%     }
%     \label{fig:observation2-b}
%     \vspace{-0.2in}
%   \end{center}
% \end{figure}

%% file: figures/1-raw-stats.tex
\begin{figure}[t]
 \begin{center}
    \centerline{\includegraphics[width=0.96\columnwidth]{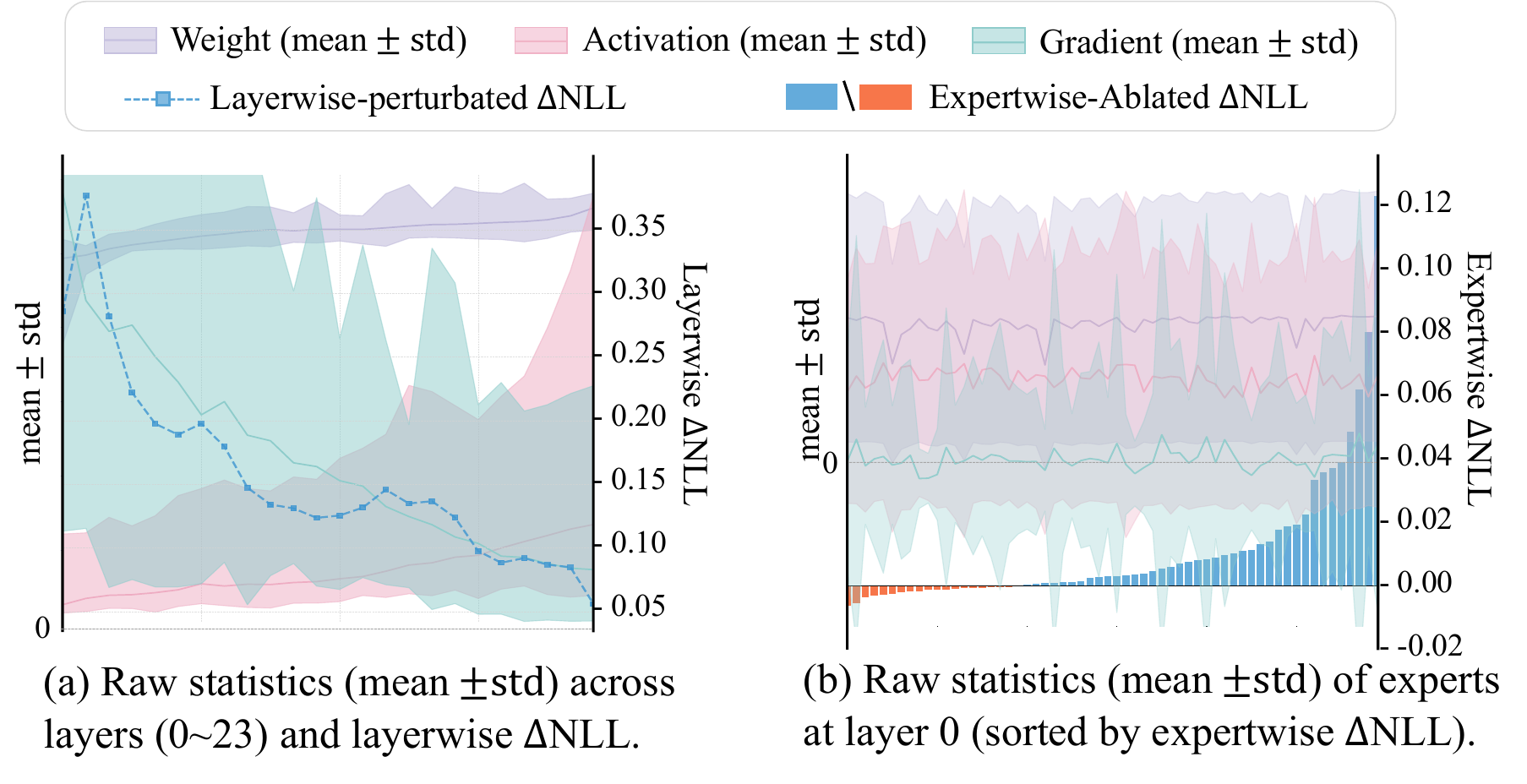}}
    \vspace{-0.08in}
    \caption{
    {Incomparability of raw statistics (weight, activation and gradient) across layers and experts.} (a) shows raw statistics of weights and activations grow monotonically while gradients decay with depth; and (b) reveals intra-layer uncorrelation between these statistics with actual $\Delta \mathrm{NLL}$ when ablated. 
    % . Experts at layer 0 sorted by their actual $\Delta \mathrm{NLL}$ when ablated, where the corresponding raw statistics show no meaningful correlation to the actual loss. 
    }
    \label{fig:observation2}
    \vspace{-0.35in}
  \end{center}
\end{figure}

% \begin{figure}[ht]
%   % \vskip 0.2in
%   \begin{center}
%     \centerline{\includegraphics[width=1.0\columnwidth]{figures/images/observation1-horizontal.pdf}}
%     \caption{
%     Layerwise statistics for different metrics. 
%     }
%     \label{fig:observation1}
%     \vspace{-0.2in}
%   \end{center}
% \end{figure}

% \begin{figure}[t]

%  \begin{center}
%     \centerline{\includegraphics[width=1.0\columnwidth]{figures/images/observation1-b.png}}
%     \caption{
%     The negative log-likelihood (NLL) of Expert-wise trimming solely over the tokens for which that expert receives the top-1 router weight. It shows bars for weights (top), activations (middle) and gradients (bottom), overlaid with loss $L$ and loss change $\Delta L$.
%     }
%     \label{fig:observation1-b}
%     \vspace{-0.2in}
%   \end{center}
% \end{figure}

%% file: figures/2_expretwise_saliency_concentration.tex
\begin{figure}[t]
  % \vskip 0.2in
  \begin{center}
     \vspace{-0.1in}
    \centerline{\includegraphics[width=0.99\columnwidth]{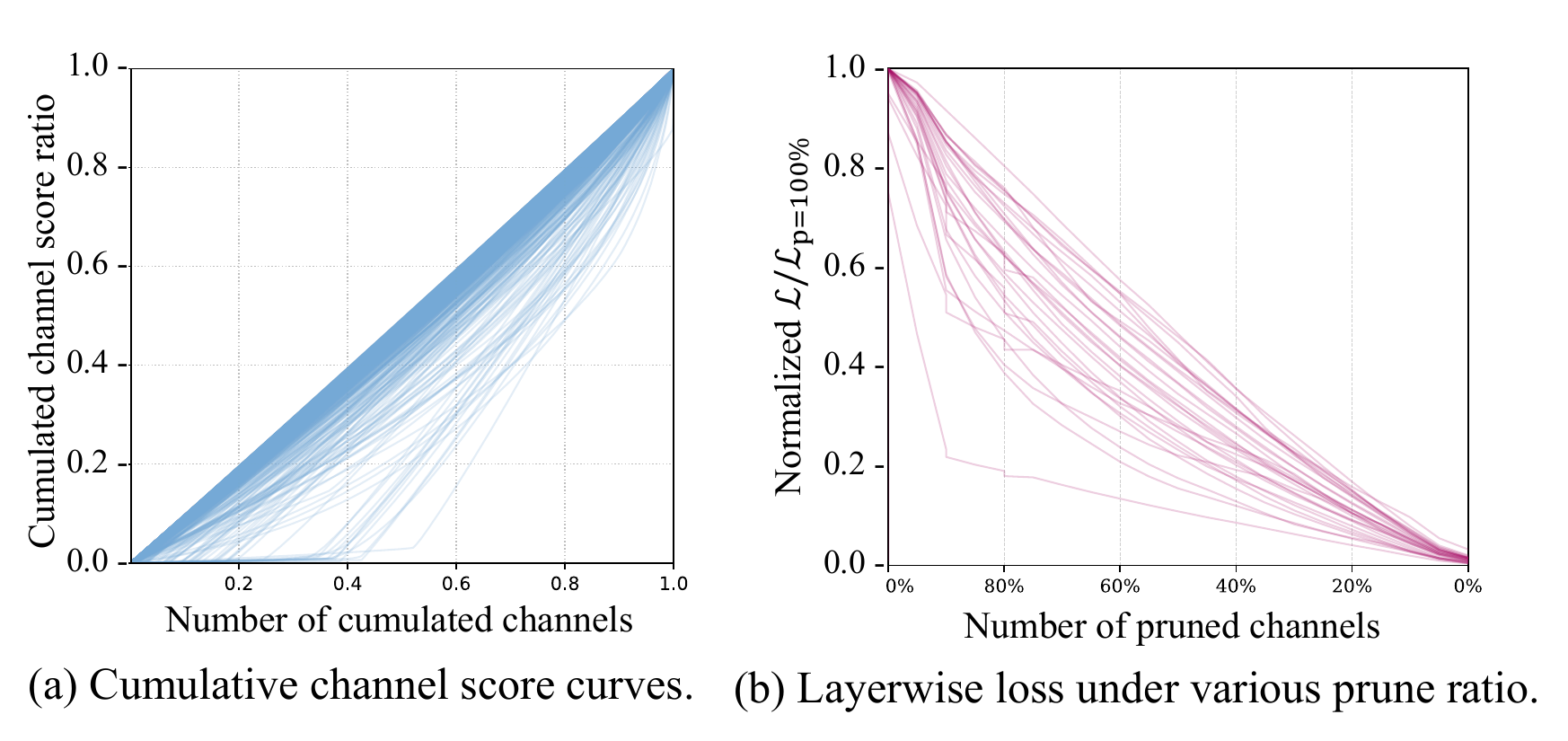}}
     \vspace{-0.08in}
    \caption{
(a) Cumulative channel score distribution, which reveal that many experts possess highly centralized channels.
(b) Layerwise output loss under various prune ratio, for some experts the loss drops rapidly after keeping only a small fraction of channels.
    }
    \label{fig:expertwise_saliency_concentration}
    \vspace{-0.35in}
  \end{center}
\end{figure}

%% file: sections/4_method.tex
\input{figures/3_overview_fig}

\section{Proposed Method}  % Coverage-Maximized Budget Allocation
\label{sec:method}

In this section, we present a structural-pruning method for MoE slimming.  % with three components. 
An overview is shown in \cref{fig:overview}. 
% Left: Attribution-based Loss Approximation (\cref{sec:method2:attribution_based_expert_wise_coverage}) provides an efficient estimator of expert contribution, avoiding expensive expert-wise ablation. 
% Middle: Maximum Coverage Budget Allocation (\cref{sec:method1:maximum_coverage}) allocates channel budgets by optimizing score coverage, addressing the strong intra-expert concentration observed in \cref{fig:expertwise_saliency_concentration}.
% Right: Alignment-Aware Redistribution (\cref{sec:method3:alignment_aware}) adjusts the allocated budgets via Hamilton apportionment to satisfy kernel-shape constraints and better integrate with low-bit quantization.

\subsection{Attribution-Guided Loss Approximation}
\label{sec:method2:attribution_based_expert_wise_coverage}

\paragraph{Rationale.}
% Expert contributions in MoE models are highly unbalanced and input-dependent, so effective slimming should allocate more channels to more contributive experts. 
As \cref{sec:router-can-have-wrong-choices} show, routing outputs and raw statistics can be weakly aligned with an expert's true impact on the final output. 
% another option is expert-wise ablation to measure the loss change for removing each expert, yet such evaluation is prohibitively expensive when repeated over many experts and layers (see \cref{sec:calib-time-comparisons} for calibration time comparisons).
% Expert-wise ablation directly measures the loss change from removing each expert, but becomes prohibitively expensive across many experts and layers (see \cref{sec:calib-time-comparisons}). 
% This motivates the following question: \textit{how can we obtain an accurate yet efficient proxy of expert-wise loss impact to serve as the input to our budget allocation algorithm?} 
This raises a key question: \textit{how can we obtain an accurate yet efficient proxy of loss impact for expert-wise prune budget allocation?} 

In this subsection, we propose an Attribution-based Loss Approximation (ALA) to estimate expert contributions, producing a scalable expert-wise loss proxy that initializes our coverage-maximized pruning algorithm in \cref{sec:method1:maximum_coverage}.

\paragraph{Derivation. }
\label{sec:attribution_derivation}
% - 对每个 expert 计算 output * grad_output 的 attribution，
%   给出一阶泰勒展开推导，说明它近似 expert 对 loss 的一阶影响。
% - 得到每层内的一组 expert saliency A_{\ell,e}，再归一成权重 w_{\ell,e}。
% \dyf{[Can be shorter. Move to Appendix. ]}
% To accurately and efficiently measure the importance of experts, we propose an Attribution-Based Approximation to estimate the contribution for all experts together in once forward. 

Let $h_\ell \in \mathbb{R}^{d}$ be the input hidden state of the MoE block in layer $\ell$, and the output can be written as 
\begin{equation}
y_\ell \;=\; \sum_{e \in \mathcal{E}_\ell} g_{\ell,e}(h_\ell)\, z_{\ell,e}.
\end{equation}
where $z_{\ell,e}=f_{\ell,e}(h_\ell)$ and $g_{\ell,e}(h_\ell)\ge 0$ are the expert output and the router probability for selected experts $e\in\mathcal E_\ell$. 
Removing expert $e$ corresponds to setting $z_{\ell,e}=0$, which perturbs the layer output by 
\begin{equation}
\Delta y_\ell^{(e)} = - g_{\ell,e}\, z_{\ell,e}.
\end{equation}
Let $\mathcal{L}$ denote the loss evaluated against the original layer output. We approximate the loss change using a first-order Taylor expansion around $y_\ell$. 
% \begin{equation}
% \mathcal L(y_\ell+\Delta y)\approx \mathcal L(y_\ell)+\left(\frac{\partial \mathcal L}{\partial y_\ell}\right)^\top \Delta y.
% \end{equation}
Thus, the loss change induced by removing expert $e$ is
\begin{equation}
\Delta \mathcal L^{(e)}
\approx
\left(\frac{\partial \mathcal L}{\partial y_\ell}\right)^\top \Delta y_\ell^{(e)}
=
-\left(\frac{\partial \mathcal L}{\partial y_\ell}\right)^\top (g_{\ell,e} z_{\ell,e}).
\end{equation}
Using the chain rule, the loss gradient with respect to the expert output satisfies $\frac{\partial \mathcal L}{\partial z_{\ell,e}}=g_{\ell,e}\frac{\partial \mathcal L}{\partial y_\ell}$. Put this into the first-order expansion yields the final approximation form
\begin{equation}
\Delta \mathcal L_{\ell}^{(e)} \approx -\left(\frac{\partial \mathcal L_{\ell}}{\partial z_{\ell,e}}\right)^\top z_{\ell,e}, 
\end{equation}
which serves as a proxy of expert contribution, and we compute for all experts within a layer in one backward pass.

\paragraph{Implementation and efficiency.}
We collect $\Delta \mathcal L_{\ell}^{(e)}$ on a calibration set of roughly 3M tokens using an exponential moving average (EMA). We perturb all experts at layer $\ell$ by uniformly scaling their activation outputs with a small factor, and then apply a simple square-root smoothing to the loss, obtaining the expert-wise importance prior $\boldsymbol \phi$. 
% In final, we apply a simple square-root smoothing on the losses. 
% After repeating the same calibration on all layers, we obtain the final smoothed loss and proceed to prune budget allocation in the following section. 

\input{tables/motivation/calib_time_cost}
\label{sec:calib-time-comparisons}
We compare calibration time in~\cref{tab:calib_time_cost} against expert-wise ablation under the same data amount and iterations. Our method reduces 14-26$\times$ time cost due to a smaller search space. While heuristics such as greedy search~\cite{cao2024condense} or genetic algorithms~\cite{liu2024eep} can reduce ablation cost, they can probably fall into local optimum. Meanwhile, it is noticeable that none of them has been validated on MoEs with hundreds of experts. 
% \paragraph{Calibration Time Comparisons. }

% Task loss can be a relatively straightforward metric to measure the impact of removing experts. Put the performance aside, they can be pretty time-consuming if we estimate all experts in all layers for larger MoE models with hundreds of experts. Methods are proposed to accelerate the performance, such as greedy search, genetic algorithm. However, they can probably fall into local optimum, and the time cost is exponentially increasing when the search space is growing. It is noticeable that none of them has been validated on MoE with hundreds of experts. 

\subsection{Coverage-Maximized Budget Allocation}
\label{sec:method1:maximum_coverage}
\input{algorithms/4_layer_coverage}
Motivated by the mismatch between experts contribution and their internal redundancy observed in \cref{subsec:importance_neq_redundancy}, we propose a new objective that directly rewards retaining the concentrated, high-contribution channels.  
% prune-ratio based objectives become poorly aligned with the actual structural redundancy in MoEs. 
% This calls for an objective that directly rewards retaining the concentrated, high-contribution channels, avoiding wasting capacity on low-scores tails. 
% In the following, we demonstrate our Coverage-Maximized Budget Allocation algorithm. 

\paragraph{Unified coverage formulation for inter- and intra-layer allocation.}
Consider a group $\mathcal{G}$, which can be either all layers or all experts within a single layer. Each layer or expert $g\in\mathcal{G}$ contains channels $c\in\mathcal{C}_g$ with non-negative scores $s_{g,c}\ge 0$. Sorting channels by $s_{g,c}$ in descending order, then we notate them as $s_{g,(1)}\ge \cdots \ge s_{g,(|\mathcal C_g|)}$. And then, we precompute the prefix sums of $n$ channels as
\begin{equation}
\mathcal{S}_g(n)=\sum_{i=1}^{n} s_{g,(i)}, \qquad
\mathrm{S}^{tot}_g=\mathcal{S}_g(|\mathcal{C}_g|),
\end{equation}
% which induces the coverage function $\rho_g(n)=\mathcal{S}_g(n)/\mathrm{S}^{tot}_g$.
Given precomputed prefix sums, the coverage ratio for top-$n$ channels is computed directly as $\rho_g(n) = \mathcal{S}_g(n) / \mathrm{S}^{tot}_g$.

The core idea of our algorithm is to change the objective of \textit{prune ratio} allocation to the \textit{channel score coverage} allocation. 
% Based on the importance estimation given by our ALA, we maximize the total scores by keeping high score channels. 
Given a global prune target $p$, we have total channel budget $N_{\mathrm{budget}}(p)=(1-p) N^{tot}=(1-p)\sum_{g\in\mathcal G}|\mathcal C_g|$. 
We allocate channels by searching for the largest target coverage vector $\boldsymbol{\rho}\in[0,1]^{|\mathcal{G}|}$ that maximizes total covered score while retaining the minimum number of channels:
\begin{equation}
% & N_g(\rho_g)=\min\left\{n \,\middle|\, \mathcal{S}_g(n)\ge \rho_g \, \mathrm{S}^{tot}_g \right\},
% \\
N(\boldsymbol\rho)=\sum_{g\in\mathcal G} N_g(\rho_g) = \sum_{g\in\mathcal G} \min\left\{n \,\middle|\, \mathcal{S}_g(n)\ge \rho_g \, \mathrm{S}^{tot}_g \right\}, 
\end{equation}
where each $\rho_g\in \boldsymbol{\rho}$ corresponds to the coverage ratio of $g$, $N(\boldsymbol\rho)$ is the minimal number of channels needed to reach coverage $\boldsymbol \rho$. 
Since $\mathcal S_g(n)$ is monotone in $n$, $N(\boldsymbol\rho)$  can be obtained efficiently via binary search (Appendix \cref{algo:layer_min_channels}). 

The pipeline of CBA is shown in \cref{algo:coverage_search}. 
We initialize $\boldsymbol\rho$ using non-negative importance prior $\boldsymbol \phi \in \mathbb R^{|\mathcal G|}$ derived from ALA, and a single scaling factor $\alpha$ (line 6 in~\cref{algo:coverage_search}). 
% \begin{equation}
% \boldsymbol\rho(\alpha)=\min(\alpha\,\boldsymbol \phi,\,1).
% \label{eq:coverage_alpha}
% \end{equation}
We apply binary search over $\alpha$ to find the largest $\alpha^\star$ such that $N(\boldsymbol\rho(\alpha^\star))\le N_{\mathrm{budget}}$, which yields the final budgets for each item $g\in\mathcal G$: $N_g^\star=N_g(\rho_g(\alpha^\star))$. 

\paragraph{Inter-layer vs. intra-layer instantiation.}
The procedure above is identical for inter-layer and intra-layer allocation, which only differ in the definition of group $\mathcal G$ and the initialization of importance estimation $\boldsymbol \phi$: 
(1) \textit{Inter-layer allocation.} $\mathcal{G}$ consists of all layers, and $\mathcal C_g$ ($\forall g\in\mathcal {G}$) includes all channels in one layer. We set $\boldsymbol \phi$ using the layerwise loss, and our algorithm producing budgets $N_\ell^\star$ for all layers. 
(2) \textit{Intra-layer allocation.} $\mathcal{G}=\{(\ell,1),\ldots,(\ell,E)\}$ contains all experts at layer $\ell$, and $\mathcal C_g$ means channels for expert $g\in\mathcal G$. $\boldsymbol \phi_\ell$ is derived from our ALA (\cref{sec:method2:attribution_based_expert_wise_coverage}), and run the same search under the layer budget $N_\ell^\star$ to obtain $N_{\ell,e}^\star, \forall e\in \mathcal E_{\ell}$. 

Overall, our CBA algorithm takes the ALA outcome as budget initialization, and translates them into kept channels by maximizing the accumulated scores within each expert. 
% , producing a complete allocation $\{N_{\ell,e}^\star\}$ under the global pruning constraint. 
As illustrated in~\cref{fig:overview}, unlike prune-by-ratio baselines (gray) that can waste capacity on low-score tails, our method retains only high-score channels to maximize score coverage (red). Time breakdown of CBA is provided in Appendix~\cref{tab:build_masks_time_breakdown}, and more details are in Appendix~\cref{app:sec:detail_of_maximum_coverage_algo}. 

\subsection{Alignment-Aware Redistribution} \label{sec:method3:alignment_aware}

\paragraph{Rationale.}
To remain compatible with low-bit quantization after pruning, inference backends (e.g., BitsAndBytes) require the input dimensions to be multiples of a hardware-friendly block size. Otherwise, frameworks may (i) trigger warnings and fall back to slower generic implementations, (ii) suffer degraded throughput. For example, Qwen3-30B-A3B drops from 14.21 tokens/s  when dimensions are aligned to 128, while 10.23 tokens/s when not aligned, see \cref{tab:align_vs_nonalign}. And (iii) incur padding that wastes storage and compute while conveying no information.  Qwen3-30B-A3B would have $4.1\%$ padded channels, corresponding to $\approx4.0\times10^8$ wasted parameters (see Appendix~\cref{app:sec:redundant_channels}). 
\input{tables/motivation/compare_align}

Our coverage-based allocation produces per-expert channel budgets $N_{\ell,e}$ optimal for score coverage, but may violate low-bit GEMM constraints that require channel dimensions to be multiples of a block size $a$ (e.g., $64$ or $128$).
We therefore apply an Alignment-Aware Redistribution (AAR) that converts $\{N_{\ell,e}\}$ into aligned budgets $\{N^\mathrm{aligned}_{\ell,e}\}$ while approaching as close as possible to the original allocation. 

\paragraph{Downward alignment. } First, we drop extremely small experts by a minimal channel threshold $m$: experts with $N_{\ell,e}<m$ are set to zero and excluded from redistribution. Because overly slim experts can convey more noise than information. 
% if the expert is slimmed too thin, it possibly conveys more noise than useful information. 
For each remaining expert, we apply downward alignment to the nearest multiple of $a$, producing a kernel-compatible base budget by $N^{\mathrm{base}}_{\ell,e}=\big\lfloor \tilde N_{\ell,e}/a \big\rfloor \cdot a$.
% , which may release a small amount of budget due to rounding. 
% Let $N_\ell^\star$ be the layer budget after coverage allocation. As shown in \cref{fig:overview}, 
The released quota after rounding in layer $\ell$ is $R_\ell = N_\ell^\star - \sum_e N^{\mathrm{base}}_{\ell,e}$ (red slices), corresponding to $q_\ell=\lfloor R_\ell/a \rfloor$ additional $a$-blocks that can be reassigned. 

% \paragraph{Hamilton largest-remainder apportionment.} We regard this redistribution as a Hamilton appointment problem. 
% Each expert's fractional remainder induced by alignment is
% \begin{equation}
% r_{\ell,e}=\frac{\tilde N_{\ell,e}-N^{\mathrm{base}}_{\ell,e}}{a},\quad \tilde N_{\ell,e} =
% \begin{cases}
% 0, & N_{\ell,e} < m,\\
% N_{\ell,e}, & N_{\ell,e} \ge m,
% \end{cases}
% \end{equation}
% which quantifies how much the expert was rounded down relative to its original budget.
% We then assign the $q_\ell$ available blocks to the experts with the largest remainders, giving each selected expert one extra $a$ block. This ensures all active experts are aligned for low-bit kernels, and preserves the coverage-based allocation as closely as possible. 

\paragraph{Hamilton largest-remainder apportionment.}
We perform alignment-aware redistribution via Hamilton's largest-remainder rule.
After rounding each active expert's channel budget down to the nearest multiple of $a$, we obtain per-expert fractional remainders $r_{\ell,e}\in[0,1)$ that quantify how close each expert is to the next $a$-block.
We then allocate the $q_\ell$ available $a$-blocks to the experts with the largest remainders, which can be written as
\begin{equation}
\label{eq:hamilton_perm_cap1_main}
b_{\ell,e}
=
\mathbb{I}\left[e \in \{\pi(1),\ldots,\pi(q_\ell)\}\right],
\end{equation}
where $\pi$ sorts experts in descending order of $r_{\ell,e}$.
The final aligned channel budgets are
\begin{equation}
\label{eq:k_final_minimal_main}
N'_{\ell,e}
=
N^{\mathrm{base}}_{\ell,e} + a \cdot b_{\ell,e}.
\end{equation}
This preserves the coverage-based allocation as closely as possible while ensuring divisibility by $a$ for efficient low-bit kernels.
The complete redistribution procedure and all implementation details are provided in Appendix~\cref{app:sec:hamilton_apportionment}.

%% file: figures/3_overview_fig.tex
\begin{figure*}[ht]
  \begin{center}
    \vspace{-0.05in}
    \centerline{\includegraphics[width=0.93\textwidth]{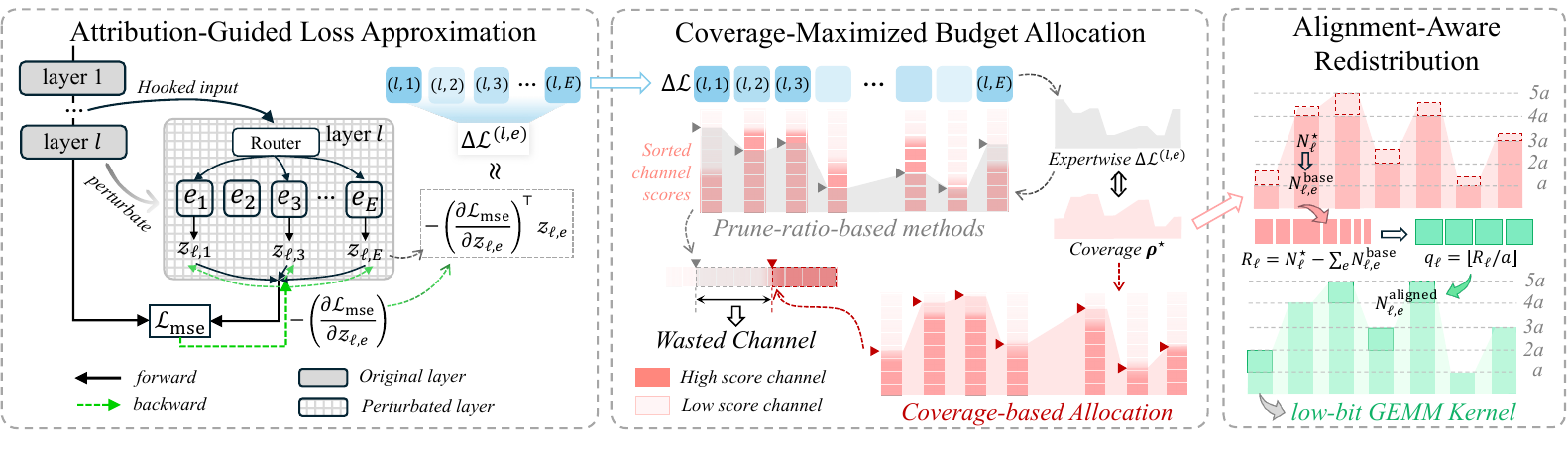}}
    \vspace{-0.05in}
    \caption{
    The overview of Attribution-Guided \& Coverage-Maximized Expert-wise Pruning framework for MoE models. 
    }
    \label{fig:overview}
    \vspace{-0.35in}
  \end{center}
\end{figure*}

%% file: tables/motivation/calib_time_cost.tex
\begin{table}[h]
\centering
\vspace{-0.05in}
\caption{Comparisons of time costs (GPU hours) between loss-based importance estimation using expert-wise ablation and ours. }
\vspace{-0.05in}
\label{tab:calib_time_cost}
\resizebox{0.99\columnwidth}{!}{
\begin{tabular}{l c c r r}
\toprule
\textbf{Model} & \textbf{Layers} & \textbf{Experts} & \textbf{Expert-ablated (h)} & \textbf{Ours (h)} \\
\midrule
DeepSeek-MoE-16B      & 28 & 64  & 23.67  & 1.70 \\
DeepSeek-V2-Lite      & 27 & 64  & 20.05  & 1.43 \\
Qwen1.5-MoE-A2.7B     & 24 & 60  & 21.77  & 1.54 \\
Qwen3-30B-A3B         & 48 & 128 & 130.10 & 5.23 \\
\bottomrule
\end{tabular}
\vspace{-0.45in}
}
\end{table}

%% file: algorithms/4_layer_coverage.tex
\begin{algorithm}[t]
  \vspace{-0.05in}
  \caption{Coverage-Maximized Allocation Search}
  % \vspace{-0.05in}
  \label{algo:coverage_search}
  \begin{algorithmic}[1]
    \STATE {\bfseries Input:} Score allocation weights $\boldsymbol \phi\in\mathbb{R}_+^{|\mathcal G|}$; prefix sums $\{\mathcal S_g(n)\}_{g\in\mathcal G}$; total scores $\{\mathrm S^{tot}_g\}_{g\in\mathcal G}$; channel budget $N_{\mathrm{budget}}$; total channels $N^{tot}$; tolerance $\varepsilon$.
    \STATE {\bfseries Output:} Channel budgets $\{N_g^\star\}_{g\in\mathcal G}$

    \STATE $\alpha_{\min} \leftarrow 0$, $\alpha_{\max} \leftarrow 1$
    \WHILE{$\alpha_{\min} < \alpha_{\max}$}
      \STATE $\alpha \leftarrow (\alpha_{\min} + \alpha_{\max})/2$
      \STATE $\boldsymbol{\rho} \leftarrow \min\big(\alpha\,\boldsymbol \phi,\,1\big)$ \label{alg:rho_alpha}
      \STATE $N(\boldsymbol{\rho}) \leftarrow \sum_{g\in\mathcal G} \min \left\{ n \,\middle|\, \mathcal{S}_g(n) \ge \rho_g(\alpha)\,\mathrm S^{tot}_g \right\}$ \label{alg:N_alpha}

      \IF{$\bigl| N(\boldsymbol{\rho}) - N_{\mathrm{budget}} \bigr| \le \varepsilon \, N^{tot}$}
        \STATE $N_g^\star \leftarrow \min \left\{ n \,\middle|\, \mathcal{S}_g(n) \ge \rho_g(\alpha)\,\mathrm S^{tot}_g \right\}$, $\forall g\in\mathcal G$
        % \STATE $N^\star \leftarrow \sum_{g\in\mathcal G} \min\left\{n \,\middle|\, \mathcal{S}_g(n)\ge \rho_g \, \mathrm{S}^{tot}_g \right\}$
        \STATE {\bfseries break}
      \ENDIF

      \IF{$N(\boldsymbol{\rho}(\alpha)) > N_{\mathrm{budget}}$}
        \STATE $\alpha_{\max} \leftarrow \alpha$
      \ELSE
        \STATE $\alpha_{\min} \leftarrow \alpha$
      \ENDIF
    \ENDWHILE
    \STATE {\bfseries return} $\{N_g^\star\}_{g\in\mathcal G}$
    % \STATE {\bfseries return}  $N^\star$
  \end{algorithmic}
  \vspace{-0.05in}
\end{algorithm}
\vspace{-0.05in}

%% file: tables/motivation/compare_align.tex
\begin{table}[h]
\centering
\vspace{-0.10in}
\caption{Throughput and latency of Qwen MoE models with and without channel alignment (under 50\% sparsity).}
\vspace{-0.05in}
\label{tab:align_vs_nonalign}
\resizebox{\columnwidth}{!}{
\begin{tabular}{lccc}
\toprule
\textbf{Model} & \textbf{Alignment ($a$)} & \textbf{Tokens/s} & \textbf{Latency (ms)} \\
\midrule
\multirow{3}{*}{Qwen1.5-MoE-A2.7B}
  & --         & 31.41  & 31.84 \\
  & 64   & 33.92  & 29.48 \\
  & 128  & 37.12 & 26.94  \\
\midrule
\multirow{3}{*}{Qwen3-30B-A3B}
  & --         & 10.23  & 97.75 \\  % 12.05, 82.97
  & 64   & 12.98  & 77.03\\
  & 128  & 14.21  &  70.38 \\
\bottomrule
\end{tabular}
\vspace{-0.3in}
}
\end{table}

%% file: sections/5_experiments.tex
\section{Experiments}

% Comparison Methods: 
% EAC-MoE~\cite{Chen2025EACMoEEA}, we report its $\alpha=0.3$ setting with 11\% pruning ratio and $\alpha=0.7$ setting with more aggressive 38\% pruning ratio, and average bitwidth 3.03. % quant+prune
% \citeauthor{he2025towards} combines expert trimming and slimming, and we report the 25\% layer/block drop with AWQ quantization in their original paper. 
% % EES and ODP results are followed EAC-MoE~\cite{Chen2025EACMoEEA}. 
% MoE-I$^2$~\cite{yang2024moei2} proposes low-rank decomposition approach.  
% PuzzleMoE~\cite{zhao2025puzzlemoe} focus on expert merging and customized CUDA kernel. 
% MoNE~\cite{Zhang2025MoNERR} is MoE pruning method which replaces redundant experts with lightweight counterparts. 

\subsection{Experimental Setup}
\label{sec:exp:setup}

\paragraph{Models and Compared Methods.} We evaluate our method on representative open-source MoEs covering different scales, including DeepSeek-MoE-16B~\cite{deepseek-moe}, DeepSeek-V2-Lite~\cite{deepseekv2}, Qwen1.5-MoE-A2.7B~\cite{qwen1.5-moe}, and Qwen3-30B-A3B-Thinking~\cite{qwen3technicalreport}. 
We compare against recent LLM or MoE compression methods, including Wanda~\cite{wanda} using unstructural pruning, MoNE~\cite{Zhang2025MoNERR}) with structural pruning, both of which denoted as \TypeP[x\%]. EAC-MoE~\cite{Chen2025EACMoEEA} and \citeauthor{he2025towards} jointly combine pruning and quantization (\TypeP[x\%]~\TypeQ[yb]). MoE-I$^2$~\cite{yang2024moei2} proposes low-rank decomposition (\TypeL[x\%]), and PuzzleMoE~\cite{zhao2025puzzlemoe} applies expert merge (\TypeM[x\%]). Here, $\mathrm x\%$ represents the parameter reduction ratio, and $\mathrm{yb}$ means it uses y-bit quantization. Detailed introductions can be found in~\cref{app:sec:compared-methods}. 

\textbf{Implementations.} We adopt 25\% channel pruning with 4-bit quantized via alignment, noted as {Ours$_Q$} (\TypeP[25\%] \TypeQ[4b]), and more aggressive 50\% pruning without quantization or alignment, noted as Ours (\TypeP[50\%]). We generate pruning allocation using C4~\cite{c4} for general benchmarks for knowledge, GSM8K~\cite{gsm8k} or OpenCodeReasoning~\cite{ahmad2025opencodereasoning} for math and code, followed by lightweight fine-tuning on Alpaca~\cite{alpaca}. 
% We report accuracy on standard reasoning and knowledge tasks (e.g., ARC, HellaSwag, PIQA, BoolQ, WinoGrande, MMLU) and exact match or pass@1 on math/code benchmarks (GSM8K, HumanEval, MATH500, AIME25) under their default protocols. 
Extended configurations are provided in Appendix~\cref{app:sec:exp:setup}.

\input{tables/acc/csqa_tasks}

\input{tables/acc/math_code_tasks}

\input{tables/acc/csqa_tasks_2}

\subsection{Overall Results}
\label{sec:exp:main_results}

\paragraph{Results on General Tasks.}
\label{sec:exp:general_tasks}
\cref{tab:overall-results-qwen} and \cref{tab:overall-results-ds} report  zero-shot accuracy on knowledge tasks together with storage with Qwen MoEs and DeepSeek MoEs, respectively. Under the quantization-aware setting, Ours$_Q$ consistently preserves or improves accuracy while substantially reducing storage across all models. In particular, on Deepseek-MoE-16B, Qwen1.5-MoE-A2.7B and Qwen3-30B-A3B, Ours$_Q$ even surpasses the original model after lightweigt fine-tuning on average performance, with more than $5\times$ storage reduction by jointly using structural pruning and quantization. On Deepseek-V2-Lite, Ours$_Q$ also achieves better performance even under more aggressive compression ratio compared to Wanda and MoNE. 
Overall, results show that our attribution-guided, coverage-maximized allocation achieves strong compression with negligible accuracy loss, and alignment-aware redistribution allows us to integrate with low-bit quantization to achieve further storage saving. 
We further compare channel-level pruning with expert-level pruning baselines at matched storage budgets in Appendix~\cref{app:sec:rebuttal_channel_vs_expert}, where the Pareto frontier in~\cref{fig:pareto_frontier} shows that the advantage becomes larger as the compression budget tightens.

\paragraph{Results on Reasoning Benchmarks.}
We also report accuracy and pass@1 performance on math and code reasoning tasks in~\cref{tab:math-code-1}. On Qwen1.5-MoE-A2.7B, our method in both quantization and non-quantization settings largely retains GSM8K accuracy (58.20 vs.\ 61.50) while improving HumanEval from 34.20 to 38.14. In contrast, previous method~\cite{c-prune} which also allocate different prune ratio on experts based on similarity clustering degrades reasoning accuracy, especially on GSM8K. 
For the larger Qwen3-MoE-30B-A3B, our method remains robust at higher difficulty, which reaches 95.0 on MATH500 under 50\% sparsity, indicating that attribution-guided coverage allocation can preserve the critical intermediate representation space while reduce noisy informations during complex reasoning even under aggressive structural compression.

\input{figures/9-memory-2-models}

\paragraph{Storage and Memory Reduction.}
\label{sec:exp:ablation:breakdown_on_p_q_ft}
We report the storage footprint and peak memory usage during runtime in~\cref{fig:speed-memory}. 
Our method yields substantial memory savings. Applying \TypeP[50\%] nearly halves peak memory on all MoEs, e.g., from 57.24GB to 32.02GB on Qwen3-30B-A3B. 
Combining \TypeP[25\%] with \TypeQ[4b] achieves the smallest storage, reducing it by over $3\times$, although the throughput may drop slightly due to on-the-fly dequantization. Results for additional MoEs are reported in Appendix~\cref{app:sec:exp:alignment_speedup_and_memory}. 
% Combining \TypeP[25\%]~\TypeQ[4b] achieves the smallest storage (about 5.1 to 5.3$\times$ reduction) and further reduces peak memory, reaching 7.64GB on Qwen1.5-MoE-A2.7B (3.62$\times$ reduction) and 17.90GB on Qwen3-30B-A3B (3.20$\times$ reduction), enabling inference on desktop GPUs with limited (24GB) memory. 

\input{tables/ablation/prune_allocation_method}

\subsection{Ablation Studies}
\label{sec:exp:ablation}

% \paragraph{Comparison of Sparsity Allocation Strategies.}
\label{sec:prune_allocation_method}
% \textbf{Effect of CBA.} 
\cref{tab:sparsity_allocation} compares inter-layer and intra-layer sparsity allocation under a 50\% pruning budget. Simple heuristics (uniform or U-shaped schedules) consistently underperform data-driven strategies, indicating that the expert importance in MoE is highly unbalanced. 
Coverage-based allocation strategy improves both inter- and intra-layer results. For inter-layer allocation, coverage initialized with smoothed loss performs best (40.0 on ARC-c, 58.2 on GSM8K), approaching the non-pruned model (40.4 and 61.5). 
\input{tables/ablation/rebuttal_smoothing_compact}
For intra-layer allocation, coverage initialized with the attribution-based proxy also outperforms others. 
These results confirm that coverage-based allocation is robust under aggressive pruning, and attribution-approximated loss yields stronger importance estimates and better performance.

\paragraph{Smoothing of layerwise loss.}
The square-root smoothing is a simple monotone-concave transform for compressing the dynamic range of layerwise losses. 
As shown in~\cref{tab:smoothing_main}, all smoothed variants outperform the unsmoothed baseline, and square-root gives the best average. Definitions of all smoothing functions and full per-task results are provided in Appendix~\cref{app:sec:rebuttal_smoothing}.

\input{tables/ablation/rebuttal_aar_reallocation}

\paragraph{AAR residual reallocation strategy.}
\label{sec:exp:aar}
In AAR reallocation, we compare two criteria: \emph{largest removed channels} (l-r-c), which prioritizes structural capacity recovery, and \emph{largest removed scores} (l-r-s), which targets score-weighted importance loss. \cref{tab:aar_reallocation} suggests that channel count and importance-weighted loss are strongly correlated under coverage-based pruning, both of which yield comparable accuracy across all block sizes $a\in\{64,128,256\}$. 
% We adopt l-r-c as the default for its simplicity. 

\subsection{Visualization and Analysis}

\input{figures/4_coverage_keep_ratio_colorbar}

\paragraph{Coverage ratio vs. Prune ratio. }
To examine the coverage-based allocation, we visualize the layer-wise raw loss, score coverage ratio and channel keep ratio in~\cref{fig:coverage_keep_ratio_colorbar}. 
% It shows that different layers exhibit different degrees of score centralization. 
Our method consistently retains a high fraction of cumulative channel scores by maximizing score coverage (about 90\% to 99\% under \TypeP[25\%]), while the kept channels vary widely from 60\% to 93\%. This highlights that layers or experts with highly concentrated scores can preserve most information with relatively few channels. 
% In contrast, flat score distributions require more channels to reach the same coverage, suggesting weaker concentration and possibly lower redundancy. 
We also provide expert-level visualizations, channel score distributions, and the resulting sparsity allocation in Appendix~\cref{app:sec:visual-loss-qwen1.5}.

\paragraph{Robustness across routing architectures.}
\label{sec:exp:routing}
Our method is not tied to a specific routing design. The main-text experiments already cover Qwen-style standard top-$k$ routing and the more constrained DeepSeek routing with load balancing. We further switch the activated experts from top-$2$ to top-$1$ on Qwen1.5-MoE-A2.7B and DeepSeek-V2-Lite. As shown in~\cref{tab:routing_topk_main}, accuracy drops under top-$1$ routing, but our method still preserves most of the original performance at $50$\% pruning under both settings. Full results and router entropy statistics are provided in Appendix~\cref{app:sec:rebuttal_routing}, confirming that expert heterogeneity persists across routing dynamics, layer depths, and data sources.

\input{figures/rebuttal_routing_pq_compact}

\paragraph{Additional analyses.}
Appendix~\cref{app:sec:rebuttal_calibration} evaluates calibration-corpus sensitivity and shows that general tasks remain stable across general-domain corpora, while math and code benefit from matched calibration data. We also provide wider pruning--quantization sweeps, second-order attribution comparisons, and AAR hyperparameter studies in Appendix~\cref{app:sec:rebuttal_pq_sweep,app:sec:rebuttal_second_order,app:sec:rebuttal_hyperparams}, which support the same accuracy--efficiency and robustness trends.

% \paragraph{Visualization of loss, scores and sparsity allocation at expert-level. }
% \input{figures/8_stacked_bar-qwen3}

% To visualize the channel scores and the resulting pruning allocation, \cref{fig:qwen3-stacked-bar} plots the cumulated expert-wise channel scores as blue stacked bars. The x-axis is the experts within a layer.
% % , and if scores in a certain expert centralized within limited number of channels, it has more darker bars.
% Experts whose scores are concentrated in a small subset of channels exhibit darker segments, whereas expert with more lighter bars have dis-centralized score distribution.
% \cref{fig:qwen3-stacked-bar}(a) shows that when we set a global maximum but uniform coverage ratio for all experts within the layer as the intra-layer allocation target, we will have less channels for experts with centralized score distribution, while give more channels for those with dis-centralized distribution. Furthermore, when we set the attribution scores as the coverage target (yellow diamond markers in \cref{fig:qwen3-stacked-bar}(b)), which are regarded as the contribution of each expert on the final output, it results in a more diverge budget allocation, indicating non-uniform importance or redundancy across experts. More visualizations are provided in Appendix~\cref{app:sec:visual-loss-qwen1.5}.

%% file: tables/acc/csqa_tasks.tex
\begin{table*}[h]
\centering
\vspace{-0.05in}
\caption{Comparison on Qwen MoE models. MMLU is evaluated under 5-shot setting, while other tasks are evaluated zero-shot.
% {Ours} represent results of not quantized model without alignment, and {Ours${_\mathrm Q}$} is that of 4-bit quantization using BitsandBytes NF4 format, aligning with $a=64$. 
}
\vspace{-0.05in}
\label{tab:overall-results-qwen}
\resizebox{0.83\textwidth}{!}{
\begin{tabular}{l c r cccccccc}
\toprule
\textbf{Method} & \textbf{Type} & \textbf{Storage} & \textbf{ARC-c} & \textbf{ARC-e} &
\textbf{Hella} & \textbf{PiQA} & \textbf{BoolQ} & \textbf{Wino} & \textbf{MMLU} & \textbf{Avg} \\
\midrule

\multicolumn{11}{c}{\textbf{Qwen1.5-MoE-A2.7B}} \\
Baseline & -- & 28.67GB & 40.41 & 69.44 & 77.17 & 80.79 & 79.57 & 69.77 &  61.08 & 68.32 \\ 
Wanda & \TypeP[25\%] & 28.67GB & 40.70 & 69.07 & 76.48 & 79.87 & 79.54 & 69.14 & 57.55 & 67.48\\
Wanda & \TypeP[50\%] & 28.67GB & 38.48 & 65.99 & 71.92 & 78.07 & 76.57 & 66.61 & 53.82 & 64.49\\
Wanda & \TypeP[25\%]~\TypeQ[4b] & 7.10GB & 41.64 & 69.74 & 76.48 & 79.38 & 79.36 & 69.22 & 60.34 & 68.02 \\
MoNE & \TypeP[25\%] & 22.40GB & 42.15 & 75.34 & 54.47 & 77.47 & 74.86 & 72.10 & 56.19 & 64.65 \\
MoNE & \TypeP[50\%] & 16.17GB & 32.17 & 64.81 & 43.52 & 70.13 & 69.05 & 63.61 & 45.67 & 55.57 \\
MoNE & \TypeP[25\%]~\TypeQ[4b] & 5.60GB & 40.36 & 59.10 & 63.06 & 81.44 & 69.37 & 66.49 & 55.39 & 62.17 \\
\citeauthor{he2025towards} & \TypeP[50\%] & 16.17GB & 26.96 & 42.21 & 45.46 & 68.23 & 63.55 & 53.35 & 31.52 & 47.33 \\
\citeauthor{he2025towards} & \TypeP[25\%]~\TypeQ[4b] & 5.60GB & 38.91 & 60.69 & 71.22 & 77.91 & 68.41 & 63.85 & 52.73 & 61.96 \\
% EAC-MoE & \TypeP[11\%]~\TypeQ[3.03b]  & 6.69GB & 42.15 & 66.92 & 75.81 & 80.41 & 76.88 & 69.22 & 58.72 & 67.16 \\ 
EAC-MoE & \TypeP[38\%]~\TypeQ[3.03b] & 4.35GB & 41.30 & 66.71 & 74.97 & 79.16 & 75.35 & 68.75 & 55.96 & 61.77 \\
MoE-I$^2$ & \TypeL[53.98\%] & 15.18GB & 41.13 & 71.68 & {53.08} & -- & 75.08 & 66.54 & -- & 57.82 \\ 
PuzzleMoE & \TypeM[25\%] & 22.40GB   & 40.9 & 73.4 & 57.3 & 79.7  & 79.2 & 69.6 & 60.4 & 65.8 \\
PuzzleMoE & \TypeM[50\%] & 16.17GB & 40.7 & 73.5 & 56.5 & 79.4 & 78.6 & 69.4 & 60.0 & 65.4 \\ 
C-Prune & \TypeP[25\%]~\TypeQ[4b] & 5.60GB & 40.00 & 62.70 & 63.06 & 78.92 & 77.12 & 67.75 & 56.68 & 63.75 \\
\cellcolor{ourcell}\textbf{Ours} & \cellcolor{ourcell}{\TypeP[50\%]} & \cellcolor{ourcell}\textbf{16.17GB} & \cellcolor{ourcell}\textbf{40.53} & \cellcolor{ourcell}\textbf{70.45} & \cellcolor{ourcell}\textbf{73.11} & \cellcolor{ourcell}\textbf{77.04} & \cellcolor{ourcell}\textbf{78.93} & \cellcolor{ourcell}\textbf{65.27} & \cellcolor{ourcell}\textbf{50.67} & \cellcolor{ourcell}\textbf{65.14} \\ 
\cellcolor{ourcell}\textbf{Ours${_\mathrm Q}$} & \cellcolor{ourcell}{\TypeP[25\%]~\TypeQ[4b]}  & \cellcolor{ourcell}\textbf{5.60GB} & \cellcolor{ourcell}\textbf{45.13} & \cellcolor{ourcell}\textbf{73.19} & \cellcolor{ourcell}\textbf{75.01}  & \cellcolor{ourcell}\textbf{79.00} & \cellcolor{ourcell}\textbf{84.22} & \cellcolor{ourcell}\textbf{68.67} & \cellcolor{ourcell}\textbf{58.29}  & \cellcolor{ourcell}\textbf{69.07} \\ 
\midrule

\multicolumn{11}{c}{\textbf{Qwen3-MoE-30B-A3B}} \\
Baseline & -- & 61.06GB & 52.70 & 79.30 & 78.50 & 79.60 & 88.70 & 72.85 & 77.80 & 75.64  \\
Wanda & \TypeP[25\%] & 61.06GB & 26.45 & 33.71 & 25.43 & 52.12 & 38.50 & 51.14 & 22.95 & 35.76 \\
Wanda & \TypeP[50\%] & 61.06GB & 24.57 & 27.23 & 26.48 & 53.75 & 40.55 & 50.12 & 22.93 & 35.09\\
PuzzleMoE & \TypeM[25\%] & 46.56GB & 51.6 & 78.9 & 58.3 & 79.3 & 88.2 & 70.4 & 76.6 & 71.9 \\
PuzzleMoE & \TypeM[50\%] & 32.07GB & 51.0 & 78.5 & 57.1 & 78.9 & 88.0 & 70.1 & 75.1 & 71.2 \\ 
\cellcolor{ourcell}\textbf{Ours} & \cellcolor{ourcell}{\TypeP[50\%]} & \cellcolor{ourcell}\textbf{32.07GB} & \cellcolor{ourcell}\textbf{52.13} & \cellcolor{ourcell}\textbf{78.03} & \cellcolor{ourcell}\textbf{77.19} & \cellcolor{ourcell}\textbf{79.33} & \cellcolor{ourcell}\textbf{83.36} & \cellcolor{ourcell}\textbf{71.59} & \cellcolor{ourcell}\textbf{61.27} & \cellcolor{ourcell}\textbf{70.81} \\ 
\cellcolor{ourcell}\textbf{Ours${_\mathrm Q}$} & \cellcolor{ourcell}{\TypeP[25\%]~\TypeQ[4b]}  & \cellcolor{ourcell}\textbf{11.64GB} & \cellcolor{ourcell}\textbf{57.94} & \cellcolor{ourcell}\textbf{81.94} & \cellcolor{ourcell}\textbf{78.87}  & \cellcolor{ourcell}\textbf{81.18} & \cellcolor{ourcell}\textbf{84.22} & \cellcolor{ourcell}\textbf{72.45} & \cellcolor{ourcell}\textbf{73.04}  & \cellcolor{ourcell}\textbf{75.66}  \\ 
\bottomrule
\end{tabular}
\vspace{-0.25in}
}
\end{table*}

%% file: tables/acc/math_code_tasks.tex
\begin{table}[h]
\centering
\vspace{-0.1in}
\caption{Reasoning benchmarks with math and code tasks. }
% GSM8K and HumanEval are evaluated under 8-shot setting, while other benchmarks are evaluated under 0-shot.} 
% , including GSM8K, HumanEval, and MBPP on DeepSeek-V2-Lite and Qwen1.5-MoE-A2.7B, and MATH500, AIME25, and LiveCodeBench-v5 on Qwen3-30B-A3B-Thinking. }
\vspace{-0.05in}
\label{tab:math-code-1}
\resizebox{0.9\columnwidth}{!}{
\begin{tabular}{lcccc}
\toprule
\textbf{Method} & \textbf{Type} & \textbf{GSM8K} & \textbf{HumanEval} & \textbf{MBPP} \\
\midrule

% \multicolumn{5}{c}{\textbf{Qwen1.5-MoE-A2.7B}} \\
% Baseline & -- & 61.50 & 34.20  & 36.60 \\
% C-Prune & \TypeP[20\%] & 39.40 & 32.90 & -- \\
% \cellcolor{ourcell}\textbf{Ours} & \cellcolor{ourcell}\TypeP[50\%] & \cellcolor{ourcell}\textbf{58.20} & \cellcolor{ourcell}\textbf{38.14} & \cellcolor{ourcell}\textbf{36.58}  \\
% \cellcolor{ourcell}\textbf{Ours$_Q$} & \cellcolor{ourcell}\TypeP[25\%]~\TypeQ[4b] & \cellcolor{ourcell}\textbf{58.53} & \cellcolor{ourcell}\textbf{36.59} & \cellcolor{ourcell}\textbf{38.52} \\
% \midrule

% \multicolumn{5}{c}{\textbf{DeepSeek-V2-Lite}} \\
% Baseline & -- & 30.94 & 32.30  & 43.2 \\
% C-Prune & \TypeP[20\%] & 26.45 & 18.90 & -- \\
% \cellcolor{ourcell}\textbf{Ours} & \cellcolor{ourcell}\TypeP[50\%] & \cellcolor{ourcell}\textbf{33.66} & \cellcolor{ourcell}\textbf{28.66} & \cellcolor{ourcell}\textbf{44.36} \\
% \cellcolor{ourcell}\textbf{Ours$_Q$} & \cellcolor{ourcell}\TypeP[25\%]~\TypeQ[4b] & \cellcolor{ourcell}\textbf{29.34} & \cellcolor{ourcell}\textbf{23.78} & \cellcolor{ourcell}\textbf{37.35} \\

\multicolumn{5}{c}{\textbf{Qwen1.5-MoE-A2.7B}} \\
Baseline & -- & 61.5 & 34.2 & 36.6 \\
C-Prune & \TypeP[20\%] & 39.4 & 32.9 & -- \\
\cellcolor{ourcell}\textbf{Ours} & \cellcolor{ourcell}\TypeP[50\%] & \cellcolor{ourcell}\textbf{58.2} & \cellcolor{ourcell}\textbf{38.1} & \cellcolor{ourcell}\textbf{36.6} \\
\cellcolor{ourcell}\textbf{Ours$_Q$} & \cellcolor{ourcell}\TypeP[25\%]~\TypeQ[4b] & \cellcolor{ourcell}\textbf{58.5} & \cellcolor{ourcell}\textbf{36.6} & \cellcolor{ourcell}\textbf{38.5} \\
\midrule

\multicolumn{5}{c}{\textbf{DeepSeek-V2-Lite}} \\
Baseline & -- & 30.9 & 32.3 & 43.2 \\
C-Prune & \TypeP[20\%] & 26.4 & 18.9 & -- \\
\cellcolor{ourcell}\textbf{Ours} & \cellcolor{ourcell}\TypeP[50\%] & \cellcolor{ourcell}\textbf{33.7} & \cellcolor{ourcell}\textbf{28.7} & \cellcolor{ourcell}\textbf{44.4} \\
\cellcolor{ourcell}\textbf{Ours$_Q$} & \cellcolor{ourcell}\TypeP[25\%]~\TypeQ[4b] & \cellcolor{ourcell}\textbf{29.3} & \cellcolor{ourcell}\textbf{23.8} & \cellcolor{ourcell}\textbf{37.4} \\

\bottomrule

\\[-2.4ex]

\toprule
\textbf{Method} & \textbf{Type} & \textbf{MATH500} & \textbf{AIME25} & \textbf{LCB}  \\
\midrule

\multicolumn{5}{c}{\textbf{Qwen3-MoE-30B-A3B}} \\
Baseline & -- & 92.8  & 76.7 &  62.6 \\
\cellcolor{ourcell}\textbf{Ours} & \cellcolor{ourcell}\TypeP[50\%] & \cellcolor{ourcell}\textbf{95.0} & \cellcolor{ourcell}\textbf{64.0} & \cellcolor{ourcell}\textbf{--} \\
\cellcolor{ourcell}\textbf{Ours$_Q$} & \cellcolor{ourcell}\TypeP[25\%]~\TypeQ[4b] & \cellcolor{ourcell}\textbf{94.5} & \cellcolor{ourcell}\textbf{75.0} &  \cellcolor{ourcell}\textbf{51.0} \\
\bottomrule

\end{tabular}
}
\vspace{-0.2in}
\end{table}

% \begin{table}[h]
% \centering
% \caption{Results on reasoning and instruction-following tasks, including MATH500, AIME25, Livecodebench-v5, and IFEval. }
% \label{tab:math-code-2}
% \resizebox{0.90\columnwidth}{!}{
% \begin{tabular}{lccccc}
% \toprule
% \textbf{Method} & \textbf{Type} & \textbf{MATH500} & \textbf{AIME25} & \textbf{LCB} & \textbf{IFEval} \\
% \midrule

% \multicolumn{6}{c}{\textbf{Qwen3-MoE-30B-A3B}} \\
% Baseline & -- & 92.8  & 76.7 &  62.6 \\
% \cellcolor{ourcell}\textbf{Ours} & \cellcolor{ourcell}\TypeP[50\%] & \cellcolor{ourcell}\textbf{95.0} & \cellcolor{ourcell}\textbf{64.0} &  \cellcolor{ourcell}{--} \\
% \cellcolor{ourcell}\textbf{Ours$_Q$} & \cellcolor{ourcell}\TypeP[25\%]~\TypeQ[4b] & \cellcolor{ourcell}\textbf{94.5} & \cellcolor{ourcell}\textbf{75.0} &  \cellcolor{ourcell}\textbf{51.0} \\

% \bottomrule
% \end{tabular}
% }
% \end{table}

%% file: tables/acc/csqa_tasks_2.tex
\begin{table*}[h]
\centering
\vspace{-0.05in}
\caption{Comparison on Deepseek MoE models. MMLU is evaluated under 5-shot setting, while other tasks are evaluated zero-shot.
% {Ours} represent results of not quantized model without alignment, and {Ours${_\mathrm Q}$} is that of 4-bit quantization using BitsandBytes NF4 format, aligning with $a=64$. 
}
\vspace{-0.05in}
\label{tab:overall-results-ds}
\resizebox{0.83\textwidth}{!}{
\begin{tabular}{l c r cccccccc}
\toprule
\textbf{Method} & \textbf{Type} & \textbf{Storage} & \textbf{ARC-c} & \textbf{ARC-e} &
\textbf{Hella} & \textbf{PiQA} & \textbf{BoolQ} & \textbf{Wino} & \textbf{MMLU} & \textbf{Avg} \\
\midrule

\multicolumn{11}{c}{\textbf{Deepseek-MoE-16B}} \\
Baseline & -- & 32.7GB & 47.53 & 73.19 & 77.43 & 80.52 & 72.57 & 69.93 & 38.18 & 65.62  \\
Wanda & \TypeP[25\%] & 32.7GB & 44.54 & 72.26 & 57.58 & 78.51 & 74.16 & 70.24 & 33.63 & 61.56 \\
Wanda & \TypeP[50\%] & 32.7GB & 44.37 & 71.46 & 54.09 & 77.86 & 75.11 & 69.85 & 32.99 & 60.82 \\
MoNE & \TypeP[25\%] & 25.04GB & 22.70 & 25.08 & 25.89 & 53.37 & 37.83 & 49.57 & 23.12 & 33.94 \\
MoNE & \TypeP[50\%] & 17.34GB & 22.53 & 24.58 & 25.57 & 49.51 & 37.83 & 49.57 & 23.12 & 33.24 \\
\citeauthor{he2025towards} & \TypeP[50\%] & 17.34GB & 28.67 & 41.41 & 53.67 & 68.17 & 38.65 & 55.17 & 23.47 & 44.17 \\
\citeauthor{he2025towards} & \TypeP[25\%]~\TypeQ[4b]  & 7.70GB & {44.00} & 70.75 & {74.50} & {78.50} & 66.00 & 67.30 & 27.90 & 59.70 \\ 
EAC-MoE & \TypeP[11\%]~\TypeQ[3.03b] & {7.19GB} & {46.16} & 73.15 & 75.55 & 79.54 & 75.02 & 70.24 & 37.45 & 65.30 \\ 
EAC-MoE & \TypeP[38\%]~\TypeQ[3.03b] & 4.47GB & {45.05}  & 71.55 & 72.86 & 77.20 & 72.51 & 66.46 & 33.33 & 62.70 \\
PuzzleMoE & \TypeM[25\%] & 25.04GB & 44.0 & 75.7 & 57.2 & 78.7 & 73.1 & 70.6 & 37.2 & 62.36 \\ 
PuzzleMoE & \TypeM[50\%] & 17.34GB & 43.0 & 75.2 & 56.3 & 78.4 & 74.5 & 70.3 & 36.9 & 62.08 \\ 

\cellcolor{ourcell}\textbf{Ours} & \cellcolor{ourcell}\textbf{\TypeP[50\%]} & \cellcolor{ourcell}\textbf{17.34GB} & \cellcolor{ourcell}\textbf{42.58} & \cellcolor{ourcell}\textbf{71.38} & \cellcolor{ourcell}\textbf{70.50} & \cellcolor{ourcell}\textbf{77.97} & \cellcolor{ourcell}\textbf{68.78} & \cellcolor{ourcell}\textbf{68.75} & \cellcolor{ourcell}\textbf{31.66} & \cellcolor{ourcell}\textbf{62.34} \\ 
\cellcolor{ourcell}\textbf{Ours${_\mathrm Q}$} & \cellcolor{ourcell}\textbf{\TypeP[25\%]~\TypeQ[4b]} & \cellcolor{ourcell}\textbf{6.26GB} & \cellcolor{ourcell}\textbf{43.77} & \cellcolor{ourcell}\textbf{73.74} & \cellcolor{ourcell}\textbf{73.77} & \cellcolor{ourcell}\textbf{79.60} & \cellcolor{ourcell}\textbf{73.55} & \cellcolor{ourcell}\textbf{71.11} & \cellcolor{ourcell}\textbf{35.70} & \cellcolor{ourcell}\textbf{63.78} \\
\midrule

\multicolumn{11}{c}{\textbf{DeepSeek-V2-Lite}} \\
Baseline & -- & 31.41GB & 46.93 & 78.37 & 77.98 & 80.20 &  79.82 & 71.35 &  45.58 & 68.60 \\ 
Wanda & \TypeP[25\%] & 31.41GB & 46.84 & 76.64 & 58.44 & 79.65 & 79.39 & 71.51 & 53.84 & 66.62\\
Wanda & \TypeP[50\%] & 31.41GB & 31.40 & 50.84 & 35.41 & 64.15 & 63.18 & 59.27 & 40.81 & 49.29\\
Wanda & \TypeP[25\%]~\TypeQ[4b] & 7.70GB & 46.59 & 76.64 & 77.12 & 79.38 & 78.99 & 71.74 & 53.94 & 69.20 \\
MoNE & \TypeP[25\%] &  24.00GB & 46.67 & 74.62 & 43.00 & 79.76 & 78.47 & 71.43 & -- & 65.65  \\
MoNE & \TypeP[50\%] &  16.57GB & 37.20 & 67.17 & 36.80 & 75.30 & 73.39 & 67.88 & -- & 59.80  \\
MoNE & \TypeP[25\%]~\TypeQ[4b] & 6.00GB & 44.86 & 73.51 & 63.78 & 81.08 & 68.65 & 66.85 & 49.03 & 63.97 \\
\citeauthor{he2025towards} & \TypeP[50\%] & 16.57GB & 33.15 & 58.92 & 57.84 & 78.20 & 50.45 & 62.52 & 36.05 & 53.88 \\
\citeauthor{he2025towards} & \TypeP[25\%]~\TypeQ[4b] & 6.00GB & 44.86 & 73.51 & 63.78 & 81.08 & 68.65 & 66.85 & 49.03 & 63.97 \\
C-Prune & \TypeP[50\%] & 16.57GB & 24.40 & 35.65 & 41.14 & 58.92 & 55.87 & 51.22 & 29.52 & 42.39 \\
C-Prune & \TypeP[25\%]~\TypeQ[4b] & 6.00GB & 41.72 & 72.81 & 53.53 & 77.86 & 70.98 & 67.80 & 47.97 & 61.81 \\
MoE-I$^2$ & \TypeL[53.98\%]  & 15.39GB & 42.58 & 71.8 & {55.16} & -- & 76.79 & 67.64 & -- & 59.62 \\ 

\cellcolor{ourcell}\textbf{Ours} & \cellcolor{ourcell}{\TypeP[50\%]} & \cellcolor{ourcell}\textbf{16.57GB} & \cellcolor{ourcell}\textbf{42.49} & \cellcolor{ourcell}\textbf{73.78} & \cellcolor{ourcell}\textbf{74.59} & \cellcolor{ourcell}\textbf{78.40} & \cellcolor{ourcell}\textbf{71.77} & \cellcolor{ourcell}\textbf{69.22} & \cellcolor{ourcell}\textbf{44.16} & \cellcolor{ourcell}\textbf{64.92} \\ 
\cellcolor{ourcell}\textbf{Ours${_\mathrm Q}$} & \cellcolor{ourcell}{\TypeP[25\%]~\TypeQ[4b]}  & \cellcolor{ourcell}\textbf{6.00GB} & \cellcolor{ourcell}\textbf{47.61} & \cellcolor{ourcell}\textbf{76.35} & \cellcolor{ourcell}\textbf{76.25} & \cellcolor{ourcell}\textbf{79.54} & \cellcolor{ourcell}\textbf{75.29} & \cellcolor{ourcell}\textbf{69.30} & \cellcolor{ourcell}\textbf{48.91} & \cellcolor{ourcell}\textbf{67.61} \\ 

\bottomrule
\end{tabular}
\vspace{-0.35in}
}
\end{table*}

%% file: figures/9-memory-2-models.tex
\begin{figure}[t]
  \begin{center}
    \vspace{-0.2in}
    \centerline{\includegraphics[width=\columnwidth]{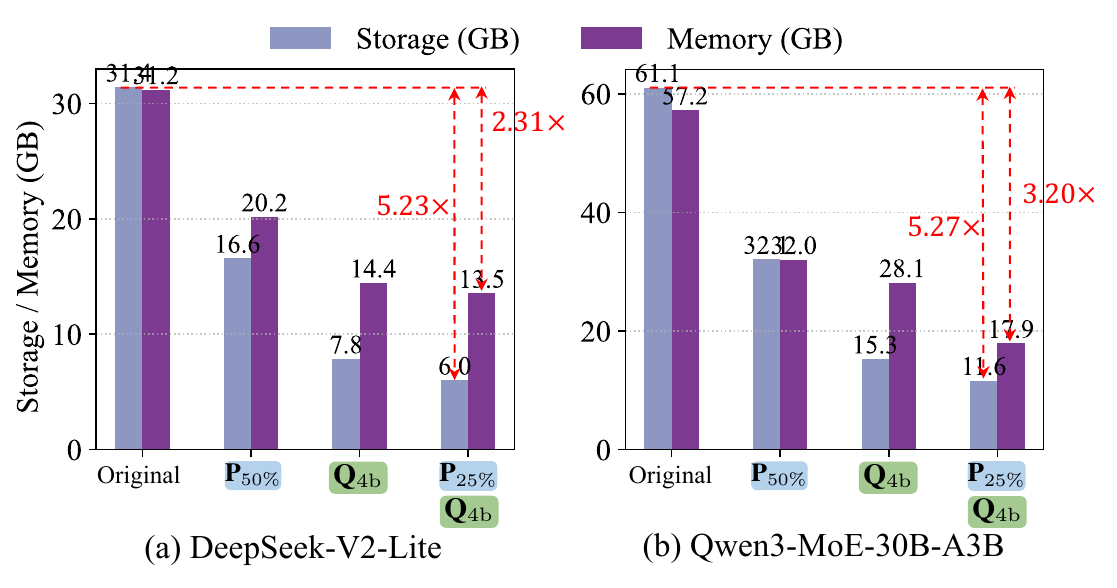}}
    \vspace{-0.10in}
    \caption{Comparison of storage and runtime memory usage (GB). }
    % \TypeP[p] denotes pruning with a sparsity of $p$\%, and \TypeQ[qb] denotes quantization to $q$-bits. All results are measured with 10 warm-up iterations and averaged over 50 runs. Pruned channel counts are aligned to multiples of 64 ($a=64$). }
    \vspace{-0.35in}
    \label{fig:speed-memory}
  \end{center}
\end{figure}

%% file: tables/ablation/prune_allocation_method.tex
% \begin{table}
% \centering
% \caption{Ablation study on sparsity allocation strategies. We report zero-shot results on GSM8K, HumanEval, and the average score of CommonSenseQA tasks.}
% % \resizebox{\columnwidth}{!}{
% \begin{tabular}{lccc}
% \toprule
% \textbf{Setting} & \textbf{GSM8K} & \textbf{HumanEval} & \textbf{CSQA} \\
% \midrule

% \multicolumn{4}{c}{\emph{Inter-layer sparsity allocation}} \\
% Non-Prune &  &  &  \\
% Uniform &  &  &  \\
% Loss-based &  &  &  \\
% Router weight  &  &  &  \\
% Token Count \\
% \cellcolor{ourcell}\textbf{Coverage} & \cellcolor{ourcell}{} & \cellcolor{ourcell}{} & \cellcolor{ourcell}{} \\
% \midrule

% \multicolumn{4}{c}{\emph{Intra-layer sparsity allocation}} \\
% Non-Prune &  &  &  \\
% Uniform &  &  &  \\
% Loss-based &  &  &  \\
% Router weight &  &  &  \\
% Token Count \\
% \cellcolor{ourcell}\textbf{Coverage} & \cellcolor{ourcell}{} & \cellcolor{ourcell}{} &  \cellcolor{ourcell}{} \\
% \bottomrule
% \end{tabular}
% % }
% \end{table}

% % router 统计要明确是 top-1 还是 top-k, 是对所有 token 还是只对被激活 token 做统计.

\begin{table}[t]
\centering
\vspace{-0.1in}
\caption{Comparison of inter- and intra-layer allocation strategies.}
\vspace{-0.05in}
\label{tab:sparsity_allocation}
\resizebox{0.93\columnwidth}{!}{
\begin{tabular}{l c c c}
\toprule
\textbf{Strategy}  & \textbf{ARC-c} & \textbf{GSM8K} & \textbf{HumanEval} \\
\midrule

\multicolumn{4}{c}{\textit{Inter-layer sparsity allocation}} \\
Non-prune               & 40.4 & 61.5 & 34.2 \\ 
Uniform                 & 37.3 & 57.8 & 30.5 \\
U-shaped                & 37.2 & 57.6 & 29.3 \\
Smoothed Loss           & 38.7 & 55.4 & 28.7 \\
\textbf{Coverage-based} & -- & -- & -- \\
\quad$\hookrightarrow$ Uniform        
                        & 38.0 & 57.0 & 29.3 \\
\quad$\hookrightarrow$ Raw loss   
                        & 38.4 & 56.1 & 30.5 \\
\quad$\hookrightarrow$ \cellcolor{ourcell}\textbf{Smoothed Loss} 
                        & \cellcolor{ourcell}\textbf{40.0} 
                        & \cellcolor{ourcell}\textbf{58.2}  
                        & \cellcolor{ourcell}\textbf{30.5} \\
\midrule

\multicolumn{4}{c}{\textit{Intra-layer sparsity allocation}} \\
Non-prune               & 40.4 & 61.5 & 34.2 \\ 
Uniform                 & 38.6 & 52.6 & 22.6 \\
Loss-based              & 37.0 & 48.6 & 16.5 \\
Router weight           & 38.2 & 57.3 & 22.6 \\
% Token count             & 38.4 & 52.8 & 20.12 \\
Topk-Channel            & 37.7 & 50.6 & 29.9 \\
\textbf{Coverage-based} & -- & -- & -- \\ 
\quad$\hookrightarrow$ Uniform         
                        & 37.9 & 53.8 & 23.2 \\
\quad$\hookrightarrow$ Raw loss     
                        & 36.1 & 45.7 & 12.2 \\
\quad$\hookrightarrow$ Router logits   
                        & 37.5 & 55.2 & 23.8 \\
% \quad$\hookrightarrow$ Token count     
%                         & 37.5 & 52.8 & 20.73 \\
\quad$\hookrightarrow$ \cellcolor{ourcell}\textbf{Attribution}   
                        & \cellcolor{ourcell}\textbf{40.0} 
                        & \cellcolor{ourcell}\textbf{58.2}  
                        & \cellcolor{ourcell}\textbf{30.5} \\
\bottomrule
\end{tabular}
\vspace{-0.3in}
}
\end{table}

%% file: tables/ablation/rebuttal_smoothing_compact.tex
\begin{wraptable}{r}{0.37\columnwidth}
\vspace{-0.8em}
\centering
\caption{Comparison of smoothing functions on Qwen1.5-MoE-A2.7B under \TypeP[50\%]. }
\label{tab:smoothing_main}
\vspace{-0.1em}
\resizebox{0.37\columnwidth}{!}{
\begin{tabular}{lc}
\toprule
\textbf{Smooth fn.} & \textbf{Avg} \\
\midrule
None                 & 52.40 \\
Log                  & 56.92 \\
Huber-style          & 55.92 \\
Clip                 & 56.69 \\
\textbf{Sqrt (ours)} & \textbf{58.27} \\
\bottomrule
\end{tabular}
}
\vspace{-2.0em}
\end{wraptable}

%% file: tables/ablation/rebuttal_aar_reallocation.tex
\begin{table}[h]
\centering
\vspace{-0.05in}
\caption{Comparison of two AAR residual reallocation strategies on Qwen1.5-MoE-A2.7B with different alignment block sizes $a$. CSQA Avg is the mean accuracy over PIQA, ARC-c, ARC-e, BoolQ, HellaSwag, and WinoGrande.}
\vspace{-0.05in}
\label{tab:aar_reallocation}
\resizebox{0.48\textwidth}{!}{
\begin{tabular}{lcccc}
\toprule
\textbf{Strategy} & $a$ & \textbf{GSM8K} & \textbf{HumanEval} & \textbf{CSQA Avg} \\
\midrule
l-r-c  & 64  & 56.29 & 26.83 & 65.12 \\
l-r-c  & 128 & 56.29 & 27.44 & 65.44 \\
l-r-c  & 256 & 57.29 & 24.39 & 65.47 \\
l-r-s           & 64  & 54.69 & 28.05 & 65.14 \\
l-r-s           & 128 & 55.89 & 26.83 & 65.04 \\
l-r-s           & 256 & 56.69 & 24.39 & 66.40 \\
\bottomrule
\end{tabular}
\vspace{-0.2in}
}
\end{table}

%% file: figures/4_coverage_keep_ratio_colorbar.tex
\begin{figure}[t]
  \begin{center}
    \vspace{-0.1in}
    \centerline{\includegraphics[width=\columnwidth]{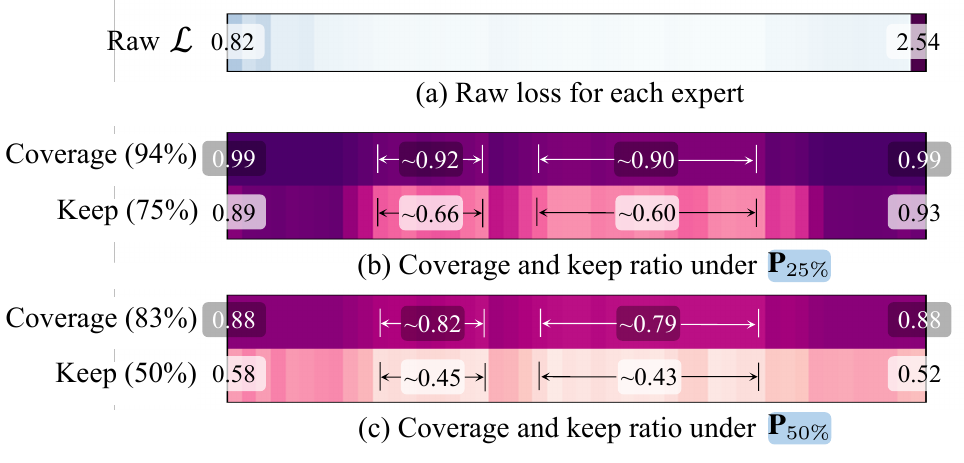}}
    \vspace{-0.05in}
    \caption{
Raw loss (top colorbar), score coverage ratio (purple colorbar) vs. channel keep ratio (channels retained after structured pruning, pink colorbar) for each layer on Qwen3-30B-A3B. }
    \vspace{-0.38in}
    \label{fig:coverage_keep_ratio_colorbar}
  \end{center}
\end{figure}

%% file: figures/rebuttal_routing_pq_compact.tex
\begin{center}
\vspace{-0.24in}
\centering
\begin{minipage}[t]{0.59\columnwidth}
\vspace{0pt}
\centering
\includegraphics[width=\linewidth]{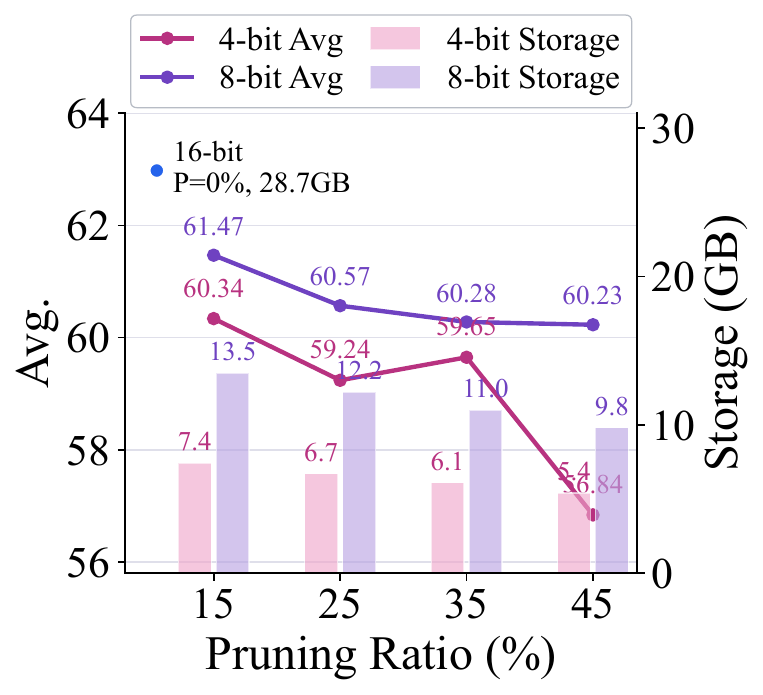}
\captionof{figure}{Combinations of pruning and quantization. The blue point is the 16-bit baseline at $P=0\%$.}
\label{fig:pq_tradeoff}
\end{minipage}
\hfill
\begin{minipage}[t]{0.36\columnwidth}
\vspace{0pt}
\centering
\scriptsize
\captionof{table}{Robustness under different top-$k$ routing strategies.}
\vspace{-0.08in}
\label{tab:routing_topk_main}
\resizebox{0.95\linewidth}{!}{
\begin{tabular}{ccc}
\toprule
\textbf{Top-$k$} & \textbf{P\%} & \textbf{Avg} \\
\midrule
\multicolumn{3}{c}{\textbf{Qwen1.5-MoE-A2.7B}} \\
4 & 0  & 62.98 \\
4 & 50 & 58.27 \\
2 & 0  & 62.98 \\
2 & 50 & 58.27 \\
1 & 0  & 60.56 \\
1 & 50 & 55.15 \\
\midrule
\multicolumn{3}{c}{\textbf{DeepSeek-V2-Lite}} \\
6 & 0  & 62.23 \\
6 & 50 & 59.08 \\
2 & 0  & 54.84 \\
2 & 50 & 50.12 \\
1 & 0  & 62.23 \\
1 & 50 & 59.08 \\
\bottomrule
\end{tabular}
}
\end{minipage}
\vspace{-0.12in}
\end{center}

%% file: sections/6_conclusion.tex
\section{Conclusion}

We propose an attribution-guided, expert-wise slimming framework for MoEs that reformulates pruning as maximizing channel-score coverage, which better captures internal redundancy and avoids allocating capacity to low-contribution structures. With alignment-aware redistribution, the pruned model remains kernel-compatible for low-bit quantization and achieves substantial compression while preserving accuracy. Experiments on modern MoEs demonstrate a practical path toward efficient MoE deployment.

% We propose an attribution-guided, expert-wise slimming framework for MoEs that reforms pruning objective to \textit{maximizing channel score coverage}, better capturing internal redundancy and excluding low-contribution structures. With alignment-aware redistribution, the pruned model integrates seamlessly with low-bit quantization, and delivers substantial compression while preserving accuracy. Experiments on modern MoEs demonstrate the practical value in MoE efficient deployment. 

% consistent performance across extensive knowledge and reasoning benchmarks. 

\section*{Impact Statement}
This paper presents work whose goal is to advance the field of machine learning. There are many potential societal consequences of our work, none of which we feel must be specifically highlighted here.

%% file: sections/10_appendix.tex
\newpage
\appendix
\onecolumn

\subsection*{Contents}
\begin{description}
    \item[\textbf{A}] \textbf{Algorithms} \dotfill \pageref{app:sec:algorithms}
        \begin{description}
            \item[A.1] Complete Process of Maximum Coverage Allocation Algorithm \dotfill \pageref{app:sec:detail_of_maximum_coverage_algo}
            \item[A.2] Hamilton Apportionment Redistribution \dotfill \pageref{app:sec:hamilton_apportionment}
        \end{description}

    \item[\textbf{B}] \textbf{Derivation and Proof} \dotfill \pageref{app:sec:derivation_proof}
        \begin{description}
            \item[B.1] Complete Proof of Attribution-based Loss Approximation \dotfill \pageref{app:sec:first_order_loss}
            \item[B.2] Derivation of Expected Redundant Channels \dotfill \pageref{app:sec:redundant_channels}
        \end{description}

    \item[\textbf{C}] \textbf{Experiments} \dotfill \pageref{app:sec:experiments}
        \begin{description}
            \item[C.1] Experimental Setup \dotfill \pageref{app:sec:exp:setup}
            \item[C.2] Overall Comparisons \dotfill \pageref{app:sec:overall_comparisons}
                \begin{description}
                    \item[C.2.1] Pareto Frontier of Channel-level vs.\ Expert-level Pruning Methods \dotfill \pageref{app:sec:rebuttal_channel_vs_expert}
                    \item[C.2.2] Wider Pruning--Quantization Combinations \dotfill \pageref{app:sec:rebuttal_pq_sweep}
                    \item[C.2.3] Speedup and Memory Usage with Different Alignment Granularity \dotfill \pageref{app:sec:exp:alignment_speedup_and_memory}
                    \item[C.2.4] Calibration Runtime Breakdown \dotfill \pageref{app:sec:calibration_runtime_breakdown}
                \end{description}
            \item[C.3] Further Ablation Studies on Proposed Methods \dotfill \pageref{app:sec:further_ablations}
                \begin{description}
                    \item[C.3.1] Channel Score Metric Selection \dotfill \pageref{app:sec:channel-score-metric}
                    \item[C.3.2] First-order vs.\ Second-order Attribution Score \dotfill \pageref{app:sec:rebuttal_second_order}
                    \item[C.3.3] Loss Smoothing \dotfill \pageref{app:sec:loss_smoothing}
                        \begin{description}
                            \item[C.3.3.1] Raw Loss vs.\ Smoothed Losses as the Target Coverage Ratio \dotfill \pageref{app:sec:raw_vs_smoothed_loss}
                            \item[C.3.3.2] Alternative Smoothing Functions for Layerwise Loss \dotfill \pageref{app:sec:rebuttal_smoothing}
                        \end{description}
                    \item[C.3.4] Hyperparameter Sensitivity in CBA and AAR \dotfill \pageref{app:sec:rebuttal_hyperparams}
                \end{description}
            \item[C.4] Sensitivity and Robustness Analysis \dotfill \pageref{app:sec:sensitivity_robustness}
                \begin{description}
                    \item[C.4.1] Sensitivity to Calibration Corpus \dotfill \pageref{app:sec:rebuttal_calibration}
                    \item[C.4.2] Robustness Across Routing Policies \dotfill \pageref{app:sec:rebuttal_routing}
                \end{description}
            \item[C.5] Visualizations \dotfill \pageref{app:sec:visualizations}
                \begin{description}
                    \item[C.5.1] Visualization of Loss, Scores and Sparsity Allocation at Expert-Level \dotfill \pageref{app:sec:visual-loss-qwen1.5}
                \end{description}
        \end{description}

    \item[\textbf{D}] \textbf{More Related Works} \dotfill \pageref{app:sec:related_work}
\end{description}

\hspace{0pt}

%%%%%%%%%%%%%%%%%%%%%%%%%%%%%%%%%
%%%%%%%%%%%%%%%%%%%%%%%%%%%%%%%%%
\section{Algorithms}
\label{app:sec:algorithms}

\subsection{Complete Process of Maximum Coverage Allocation Algorithm}
\label{app:sec:detail_of_maximum_coverage_algo}

This section provides full algorithmic details for the Coverage-Based Allocation (CBA) introduced in Section 4.2, including the concrete inter-layer and intra-layer instantiations and the associated search procedures.

\paragraph{Inter-layer Allocation. }
\label{sec:coverage-search-for-inter-layer-allocation}
We first allocate the overall channels to each layer based on layerwise saliency coverage ratio search. 
Given a target prune ratio $p\in (0,1)$, we have the overall remaining channels for the prune model $N_\mathrm{target} = (1-p)\,IEL$, where $I,E,L$ are the intermediate dimension, expert numbers in each layer, and number of layers, respectively. 

\textit{Step 1: Channel saliency preparation. } Let $c\in\mathcal{C}_\ell$ represent all the intermediate channels at layer $\ell$ (i.e., $|\mathcal{C_{\ell}}|=I \, E$). Each channel has a non-negative saliency score $s_{\ell,c} \ge 0$, which can be computed through various criteria, such as activation magnitude, weight norms, gradient norms, or simple combinations of them. We provide an ablation on the selection of channel saliency criteria in~\cref{app:sec:channel-score-metric}. 

% define the total score at layer $\ell$ as $\mathrm{S}^{tot}_\ell = \sum_{c\in \mathcal C_\ell} s_{\ell,c}$, and 
\textit{Step 2: Prefix sums calculation. } 
We sort channels layerwisely by their saliency scores in descending order, and rewrite $s_{\ell,(1)} \ge s_{\ell,(2)} \ge \dots \ge s_{\ell,(|\mathcal{C}_\ell|)}$ for the sorted scores. 
Next, we calculate the prefix sums of the leading $n$ channels at layer $\ell$ as $\mathcal{S}_\ell(n)$ by \cref{algo:prefix_sum}: 
\begin{equation}
\mathcal{S}_\ell(n) \;=\; \sum_{i=0}^{n} s_{\ell,(i)}, ~\text{where}~ 0 \le n \le |\mathcal{C}_\ell|. 
\end{equation}
Based on the prefix sums, the saliency coverage ratio of top-$n$ channels in layer $\ell$ can be obtained in $\mathcal O(1)$ time as $\rho_\ell(n) = \frac{\mathcal{S}_\ell(n)}{\mathrm{S}^{tot}_\ell}$, where $\mathrm{S}^{tot}_\ell = \sum_{c\in \mathcal C_\ell} s_{\ell,c}$ is the total saliency score. 
\label{app:sec:prefix_sum}
\input{algorithms/3_prefix_sum}

\textit{Step 3: Layerwise loss collection.} 
Inspired by previous studies that layers have diverse functionalities and redundancy~\cite{skean2025layer}, we use a small calibration dataset to collect layerwise loss by injecting a scaling as noise into a specific layer, and computing the Negative Log-Likelihood Loss (NLL) compared to the original model. Since the raw loss can have a large range, we conduct a simple smooth by square-root, and get the initial layerwise target saliency coverage ratio $\mathbf{w}\in\mathbb{R}^{L}$.

\textit{Step 4: Best coverage ratio by binary search.}
Given a target global pruning ratio $p$, our goal is to find the largest saliency coverage ratio ${\boldsymbol{\rho}}^\star \in [0,1]$ that covers as more high-saliency channels as possible, under the constraint of total remaining channels $N_\mathrm{target}$. 
We conduct an one-dimensional binary search to approach the supremum ${\boldsymbol{\rho}}^\star$. 

As shown in~\cref{algo:coverage_search}, we apply a coefficient $\alpha$ as the starting point (\#L5):
\begin{equation}
\label{eq:phi_alpha}
    \boldsymbol \rho (\alpha) = \min \big(\alpha\, \boldsymbol \phi, 1\big), 
\end{equation}
where $\boldsymbol \rho (\alpha) = \{\rho_1,\rho_2, \dots, \rho_L\}$. 
% Starting from the initial set of coverage ratio $\boldsymbol{\rho}_0=\{\rho_1,\rho_2, \dots, \rho_\ell\}$, where $\ell\in[1,L]$, 
Then, we accumulate the minimal number of channels required in each layer to reach the saliency coverage ratio as 
\begin{align}
\label{eq:N_alpha}
N( \boldsymbol{\rho} (\alpha)) &=\sum_{\ell\in [1,L]}  N_\ell(\rho_\ell)  \\
&= \sum_{\ell\in [1,L]} \min \left\{ n \,\middle|\, \mathcal{S}_\ell(n) \;\ge\; \rho_\ell \ \mathrm{S}^{tot}_\ell \right\}. 
\end{align}
Since $\mathcal{S}_\ell(n)$ is monotonic, we can quickly find $N_\ell(\rho)$ in $O(1)$ time. Algorithm is provided in \cref{algo:layer_min_channels}. 
\input{algorithms/2_layer_min_channels}

\textit{Step 5: Termination condition. }
If the gap between the current total number of preserved channels $N(\rho)$ and the target $N_{\mathrm{budget}}$ (\#L8) falls below the tolerance $\epsilon$ (to control pruning precision, $\epsilon$ is set to 0.01 as default), the loop is terminated right away (\#L9). In practice, we also impose a maximum iterations to prevent long searching. If not terminated, we perform a binary search over $\alpha \in [0,1]$ (\#L12--16), and finally get the supreme $\alpha^\star$ such that $N(\alpha^\star) \le N_{\mathrm{budget}}+\epsilon N^{tot}$, i.e., $p(\boldsymbol{\rho}^\star) \approx p$. 

% This procedure yields layer-wise pruning budget allocation $N_\ell^\star$ that satisfy the global pruning constraint $p$ while maximizing the global coverage ratio. In this way, layers that have more redundancy will have a smaller budget, on the contrary, layers that have dispersed channel scores can have more remaining channels. 

% 在层内：在给定本层通道预算的约束下，
% 按通道 score 排序，并使用同样的 coverage 思路选择要保留的 channel。

% \subsubsection{Weighted Coverage Allocation}
% - 在本层已有的 total budget n_\ell 下，用 w_{\ell,e} 对 expert 内通道能量进行重加权 or 分配，
%   二分搜索得到满足本层预算的最大 coverage，并导出每个 expert 的通道数。
% - 明确写出“配合 Section 4.2 的 coverage 框架，我们将 per-layer budget 细化为 per-expert channel budgets”。
\paragraph{Intra-Layer Allocation. } 
\label{sec:weighted_coverage_allocation}
After layerwise budget $N_{\ell}^{\star}$ is fixed, we perform the coverage search for intra-layer pruning for each expert. 

\input{algorithms/app_4_expert_coverage}
\textit{Step 1: Channel saliency reuse.} 
We reuse the channel saliency scores obtained above and denote as $s_{\ell, e, c}$ for expert $e$ at layer $\ell$. Next, since we have calculated the layerwise prefix sums, it only costs $<$0.1\textit{ms} to recompute the expert-wise prefix sums $\mathcal S_{\ell,e}(n)$. Time breakdown can be found in Appendix~\cref{app:sec:calibration_runtime_breakdown}. 

\textit{Step 2: Expert-wise importance estimator. }
We provide an efficient and accurate expert-wise loss approximation in \cref{sec:attribution_derivation}, taking place of the time-consuming expert-wise loss collection by ablating a specific expert one-by-one. Let $\boldsymbol \phi_{\ell}$ denote the expert-wise importance at layer $\ell$.
% where $\boldsymbol \phi_{\ell} =\{w_{\ell,1}, w_{\ell,2}, \dots, w_{\ell,E} \}$. 

\textit{Step 3: Best coverage ratio by binary search.} 
Similar to inter-layer saliency coverage search, given a target number of remaining channels $N_\ell^\star$, we start from an initial scaling factor $\alpha \in (0,1)$ and have
\begin{equation}
\boldsymbol\rho_{\ell}(\alpha) = \min\big(\alpha\,\boldsymbol \phi_{\ell}, 1\big), \qquad \forall \ell \in [1,L], 
\end{equation}
where $\boldsymbol{\rho}_{\ell}(\alpha) = \{\rho_{\ell, 1}, \rho_{\ell, 2}, \dots, \rho_{\ell, E}\}$. Next, the minimal channels required at layer $\ell$ to reach the coverage is
\begin{align}
N_\ell(\boldsymbol{\rho}_\ell(\alpha)) & = \sum_{e\in[1,E]} N_{\ell,e}(\rho_{\ell, e}) \\
& = \sum_{e\in[1,E]} \min \left\{ n \,\middle|\, \mathcal S_{\ell,e}(n) \ge \rho_{\ell,e}\, S_{\ell,e} \right\},
\end{align}
where $\mathrm S^{tot}_{\ell,e} = \sum_{c\in\mathcal C_{\ell, e}}s_{\ell, e, c}$ is the total saliency  score of expert $e$ at layer $\ell$. 
We then iteratively search $\alpha$ to adjust $\boldsymbol{\rho}_{\ell}$, comparing $N_\ell(\boldsymbol{\rho}_{\ell})$ against the target channel budget $N_\ell^\star$, until we find the optimal $\alpha^\star$ that 
$
\big|N_\ell(\boldsymbol{\rho}_{\ell}^\star) - N_\ell^\star\big| \le \varepsilon N^{tot}_{\ell},
$ 
where $\boldsymbol{\rho}_{\ell}^\star = \alpha^\star\mathbf{w}_{\ell}, \forall \ell\in[1, L]$. Each layer repeats the same binary search. The complete algorithm for intra-layer allocation can be found in~\cref{algo:expert_coverage}.

Combining inter-layer and intra-layer allocation, we obtain the final pruning budget for each expert $N_{\ell,e}^\star$ that satisfy the global pruning constraint $p$ while maximizing the saliency coverage ratio with the given layerwise/expert-wise importance. In this way, layers/experts that have more redundancy will have a smaller budget, on the contrary, layers/experts that have dispersed channel saliency  can have more remaining channels. 

As for the computational complexity of the overall allocation process, since $\mathcal S_{\ell}(n)$ and $\mathcal S_{\ell,e}(n)$ are non-decreasing in $n$, $N(\boldsymbol{\rho}(\alpha))$ and $N_\ell(\boldsymbol{\rho}_\ell(\alpha))$ are non-decreasing in $\alpha$, the binary search can be efficiently performed over $\alpha$ in $\mathcal O(1)$ time to find the optimal $\alpha^\star$ for global and for all the layers.

\subsection{Hamilton apportionment redistribution. }
\label{app:sec:hamilton_apportionment}
We regard the channel redistribution problem as the classical \textit{Hamilton apportionment}. 
Let \(N_{\ell,e}\) denote the allocated channels of expert \(e\) in layer \(l\) given by our maximized coverage algorithm introduced above, and let \(a\) be the supported GEMM block size (e.g., \(a=64,128,\dots\)). 
We additionally introduce a minimal channel threshold \(m\) to eliminate extremely small experts that carries little information due to too few channels left. 

\textit{Step 1: Minimal-channel trimming.}
We first trim experts whose allocated channels are smaller than \(m\):
\begin{equation}
\label{eq:trim_minimal}
\tilde N_{\ell,e} =
\begin{cases}
0, & N_{\ell,e} < m,\\
N_{\ell,e}, & N_{\ell,e} \ge m,
\end{cases}
\qquad
\mathcal A_\ell = \{e \mid \tilde N_{\ell,e} > 0\},
\end{equation}
where \(\mathcal A_\ell\) denotes the active experts after trimming. 

\textit{Step 2: Downward alignment.}
For each remaining expert \(e \in \mathcal A_\ell\), we round \(\tilde N_{l,e}\) down to the nearest multiple of \(a\), ensuring compatibility with low-bit GEMM kernels: 
\begin{equation}
\label{eq:k_base_minimal}
N^{\mathrm{base}}_{\ell,e}
= \left\lfloor \frac{\tilde N_{\ell,e}}{a} \right\rfloor \cdot a,
\qquad e \in \mathcal A_\ell,
\end{equation}
and set \(N^{\mathrm{base}}_{\ell,e}=0\) for trimmed experts \(e \notin \mathcal A_\ell\). 

\textit{Step 3: Compute remaining quota and available blocks.}
The channel budget released by trimming and alignment is collected and segmented as units of \(a\)-blocks. Therefore, the remaining quota to be redistributed , and the number of available blocks in layer $\ell$ can be expressed as
\begin{equation}
\label{eq:remaining_quota}
R_\ell =   N_\ell^\star - \sum_{e\in[1, E]} N_{\ell,e}^{\mathrm {base}}, \quad q_\ell = \left\lfloor \frac{R_\ell}{a} \right\rfloor.  
\end{equation} 

\textit{Step 4: Hamilton apportionment over experts. }
We redistribute the remaining quota in $q_\ell$ discrete \(a\)-blocks. 

The fractional remainder of each expert induced by downward alignment is:
\begin{equation}
\label{eq:remainder_from_rounding}
r_{\ell,e}
= \frac{\tilde N_{\ell,e} - N^{\mathrm{base}}_{\ell,e}}{a}
\in [0,1),
\quad e \in \mathcal A_\ell .
\end{equation}
To approach the original allocation derived by the expert importance, each expert can receive at most one additional block. 
We sort $r_{\ell, e}$ in descending order, and let \(\pi\) be a permutation of \(\mathcal A_\ell\) such that \(r_{\ell,\pi(1)} \ge r_{\ell,\pi(2)} \ge \cdots \ge r_{\ell,\pi(|\mathcal A_{\ell}|)} \). The largest $q_\ell$ experts can receive the additional $a$-block, which can be simply written as
\begin{equation}
\label{eq:hamilton_perm_cap1}
b_{\ell,e} =
\mathbb{I}\left[e \in \{\pi(1),\ldots,\pi(q_\ell)\}\right],
\end{equation}
Finally, the aligned channels are
\begin{equation}
\label{eq:k_final_minimal}
N'_{\ell,e} = N^{\mathrm{base}}_{\ell,e} + a \cdot b_{\ell,e}.
\end{equation}
The resulting aligned channels approaches the original layerwise allocation budget, satisfies expert capacity constraints, and guarantees that every expert has a channel dimension divisible by \(a\). It enables the pruned model to be stored and computed by low-bit quantization, yielding both effective compression and inference speedup without redundant zero padding on MoE models. 

\section{Derivation and Proof}
\label{app:sec:derivation_proof}

\subsection{Complete Proof of Attribution-based Loss Approximation in \cref{sec:method2:attribution_based_expert_wise_coverage}}
\label{app:sec:first_order_loss}

Let $h_\ell \in \mathbb{R}^{d}$ be the input hidden state of the MoE block in layer $\ell$, and let $z_{\ell,e} \;=\; f_{\ell,e}(h_\ell) \in \mathbb{R}^{d}$ denote the output of expert $e$ before gating. 
The output of MoE layer $\ell$ is the weighted sum of top-$k$ experts: 
\begin{equation}
y_\ell \;=\; \sum_{e \in \mathcal{E}_\ell} g_{\ell,e}(h_\ell)\, z_{\ell,e}.
\end{equation}
where $\mathcal{E}_\ell$ is the top-$k$ experts selected by the router at layer $\ell$, and $|\mathcal{E}_\ell| = k$ ($k$ is typically set as $1,2,4,8$ in modern MoE). 
$g_{\ell,e}(h_\ell) \ge 0$ is the router weight of expert $e$. 

We measure the contribution of expert $e$ at layer $\ell$  by the loss change when removing this expert. 
If the expert $e\in\mathcal{E}_\ell$ is ranked as top-$k$ by the router and selected for a specific token, removing it corresponds to replacing $z_{\ell,e}$ with zero, which will induce a perturbation in the layer output
\begin{align}
\label{eq:delta_y_ell}
\Delta y_\ell^{(e)}
&= \hat{y}_\ell^{(e)} - y_\ell \\
&= \sum_{e' \in \mathcal{E}_\ell \setminus \{e\}} g_{\ell,e'} z_{\ell,e'}
   \;-\; \sum_{e' \in \mathcal{E}_\ell} g_{\ell,e'} z_{\ell,e'} \\
&= -\, g_{\ell,e}\, z_{\ell,e}, 
\end{align}
where $\hat{y_\ell}^{(e)}$ is the layer output when removing expert $e$. 

Let $\mathcal{L}$ be the loss compared to the original layer's output. For any perturbation $\Delta y$ applied to the layer output $y_\ell$, the loss can be written by the first-order Taylor expansion as 
\begin{equation}
\label{eq:L_yl+delta_y}
\mathcal{L}(y_\ell + \Delta y)
\;=\;
\mathcal{L}(y_\ell)
+
\left( \frac{\partial \mathcal{L}}{\partial y_\ell} \right)^{\top} \Delta y
+
\mathcal{O}\bigl(\|\Delta y\|^2\bigr).
\end{equation}
In our case, removing expert $e$ at layer $\ell$ induces $\Delta y_\ell^{(e)}$ (\cref{eq:delta_y_ell}) to the block output, and the loss change is
\begin{equation}
\Delta \mathcal{L}^{(e)}
\;=\;
\mathcal{L}\bigl(y_\ell + \Delta y_\ell^{(e)}\bigr) - \mathcal{L}(y_\ell),
\end{equation}
which can be approximated by only keeping the first-order term in \cref{eq:L_yl+delta_y} as
\begin{equation}
\Delta \mathcal{L}^{(e)}
\;\approx\;
\left( \frac{\partial \mathcal{L}}{\partial y_\ell} \right)^{\top} \Delta y_\ell^{(e)}
= - \left( \frac{\partial \mathcal{L}}{\partial y_\ell} \right)^{\top} \bigl( g_{\ell,e} z_{\ell,e} \bigr).
\end{equation}

By the chain rule, the gradient w.r.t. the expert output $z_{\ell,e}$ is
\begin{equation}
\frac{\partial \mathcal{L}}{\partial z_{\ell,e}}
\;=\;
g_{\ell,e}\, \frac{\partial \mathcal{L}}{\partial y_\ell},
\end{equation}
so the final loss change can be estimated as
\begin{equation}
\Delta \mathcal{L}^{(e)}
\;\approx\;
- \left( \frac{\partial \mathcal{L}}{\partial z_{\ell,e}} \right)^{\top} z_{\ell,e}.
\end{equation}

The approximated loss is then used to measure the importance of experts at layer $\ell$ altogether. 

\subsection{Derivation of expected redundant channels in \cref{sec:method3:alignment_aware} Rationale}
\label{app:sec:redundant_channels}
From the pruning perspective, one can define a purely logical sparsity level as
\[
s_{\text{logical}}
= 1 - \frac{K}{D},
\]
where \(D\) is the original channel dimensionality and \(K\) is the number of channels retained after pruning. Under 4-bit quantization, however, parameters are physically stored and processed in fixed-size blocks. With a block size of 64, a linear layer with effective hidden size \(K\) is packed as
\[
\tilde{D}
= \left\lceil \frac{K}{64} \right\rceil \cdot 64,
\]
and the corresponding physical compression ratio becomes
\[
s_{\text{physical}}
= 1 - \frac{\tilde{D}}{D}.
\]
If we do not explicitly align \(K\) during pruning, each expert can waste between 0 and 63 channels at the storage level. Assuming that the residue \(K \bmod 64\) is approximately uniform in \(\{0,\dots,63\}\), the expected padding overhead per expert is
\[
\mathbb{E}[\tilde{D} - K]
= \frac{1}{64} \sum_{r=1}^{63} (64 - r)
= 31.5 \text{ channels}.
\]

For a Qwen3-style MoE block with hidden size \(D = 768\), \(E = 128\) experts and \(L = 64\) layers, this corresponds to roughly 
\[
\frac{31.5}{768} \approx 4.1\%
\]

\section{Experiments}
\label{app:sec:experiments}

\subsection{Experimental Setup}
\label{app:sec:exp:setup}

\paragraph{Models.}
We conduct experiments on the following representative open-source MoE LLMs that cover different scales and architectural choices:
DeepSeek-MoE-16B~\cite{deepseek-moe},
DeepSeek-V2-Lite~\cite{deepseekv2},
Qwen1.5-MoE-A2.7B~\cite{qwen1.5-moe},
and Qwen3-30B-A3B-Thinking~\cite{qwen3technicalreport}.

\paragraph{Compared Methods.}
\label{app:sec:compared-methods}
We compare our approach with advanced MoE compression baselines with various techniques: 
{EAC-MoE}~\cite{Chen2025EACMoEEA} performs joint pruning and quantization; we report its configurations with $\alpha=0.3$ (11\% sparsity) and $\alpha=0.7$ (38\% sparsity), and the corresponding average bitwidth of 3.03.
\citeauthor{he2025towards}~\cite{he2025towards} combines expert trimming and slimming; we report the 25\% layer or block drop setting together with 4-bit  AWQ quantization. 
{MoE-I$^2$}~\cite{yang2024moei2} jointly applies inter-expert pruning to remove redundant experts and intra-expert low-rank decomposition to reduce the parameter redundancy within remaining expert.
{PuzzleMoE}~\cite{zhao2025puzzlemoe} focuses on expert merging by 25\% or 50\%, and provides customized CUDA kernels for efficient inference.
{MoNE}~\cite{Zhang2025MoNERR} prunes MoE models by replacing redundant experts with lightweight counterparts. 
{C-Prune}~\cite{c-prune} addresses intra-layer and inter-layer expert redundancy in MoE LLMs via a two-stage framework of layer-wise expert clustering followed by global cluster pruning.
{Wanda}~\cite{wanda} is a training-free unstructured pruning method that scores each weight by the product of its magnitude and the corresponding input activation norm, requiring no retraining or weight reconstruction.
We report the results from the original papers under the closest comparable settings.

\paragraph{Pruning Settings. }
We use channel pruning as structural sparsification technique for easy implementation by mainstream inference engine. In the following experiments, we adopt two variants: 
\textbf{Ours} applies 50\% channel sparsity without quantization, and thus does not require alignment-aware redistribution. If an expert is assigned zero channel after pruning, we trim the expert and shrink the corresponding router dimension, so that the expert is never selected. 
Furthermore, \textbf{Ours$_\mathrm Q$} applies 25\% channel sparsity and further performs 4-bit quantization using BitsAndBytes NF4. 
Meanwhile, we enable Alignment-Aware Redistribution when applied quantization with granularity $a=128$ and enforce the minimum expert channel size $m=128$.
We select $a=128$ and $m=128$ based on a small grid search over feasible settings, constrained by linear layer shapes and quantized operator support.
We report throughput and peak memory trade-offs of the explored settings in Appendix~\cref{app:sec:exp:alignment_speedup_and_memory}, \cref{fig:speedup_memory_heatmap}, and choose the setting with best overall efficiency.

\paragraph{Calibration and Fine-tuning. }
We use C4~\cite{c4} as the calibration dataset for commonsense tasks. For reasoning benchmarks, we calibrate using samples drawn from GSM8K~\cite{gsm8k} or OpenCodeReasoning~\cite{ahmad2025opencodereasoning} depending on the task category. 
After pruning, we follow \citeauthor{yang2024moei2} to perform fine-tuning on Alpaca~\cite{alpaca} for 2 epochs.
We fine-tune the MoE blocks using DoRA~\cite{liu2024dora} with rank 32 and learning rate $1\mathrm{e}{-4}$, while adapting the routing module with rank 4 and learning rate $1\mathrm{e}{-6}$. 
We use AdamW with warmup ratio 0.1 and clip gradient exceeding 0.5, without weight decay. 
All training is conducted on 4$\times$H20 GPUs. The training cost is 12 GPU hours for Qwen1.5-MoE-A2.7B and 48 GPU hours for Qwen3-30B-A3B, and models of similar scale exhibit comparable training time. 

\paragraph{Benchmarks and Evaluation.}
We evaluate using two widely adopted toolkits: the LM Evaluation Harness\footnote{\url{https://github.com/EleutherAI/lm-evaluation-harness}} and OpenCompass\footnote{\url{https://github.com/open-compass/opencompass}}.
We report zero-shot performance on general reasoning and knowledge benchmarks, including ARC-C~\cite{arc_c}, ARC-E~\cite{arc_c}, HellaSwag~\cite{hellaswag}, PIQA~\cite{piqa}, BoolQ~\cite{boolq}, WinoGrande~\cite{winogrande}, and MMLU~\cite{mmlu}, and math/code benchmarks with 8-shot, including GSM8K~\cite{gsm8k}, HumanEval~\cite{humaneval}, MATH500~\cite{math500}, AIME25~\cite{aime25}, GPQA~\cite{gpqa}, and LiveCodeBench~\cite{jain2024livecodebench}. 
We follow the default task configurations and official evaluation protocols provided by the toolkits and report standard metrics, e.g., \textit{accuracy} for multiple-choice tasks, \textit{exact match} for math, and \textit{pass@1} for code generation. 
For long-context evaluations, we set the maximum sequence and output lengths as follows: AIME25 uses \texttt{MAX\_SEQ\_LEN=65536} and \texttt{MAX\_OUT\_LEN=32768}; MATH500 uses \texttt{MAX\_SEQ\_LEN=16384} and \texttt{MAX\_OUT\_LEN=4096}; LiveCodeBench\_v6\_academic uses \texttt{MAX\_SEQ\_LEN=32768} and \texttt{MAX\_OUT\_LEN=16384}.

\subsection{Overall Comparisons}
\label{app:sec:overall_comparisons}

\subsubsection{Pareto Frontier of Channel-level vs.\ Expert-level Pruning Methods}
\label{app:sec:rebuttal_channel_vs_expert}

\begin{figure}[h]
\centering
\includegraphics[width=0.5\textwidth]{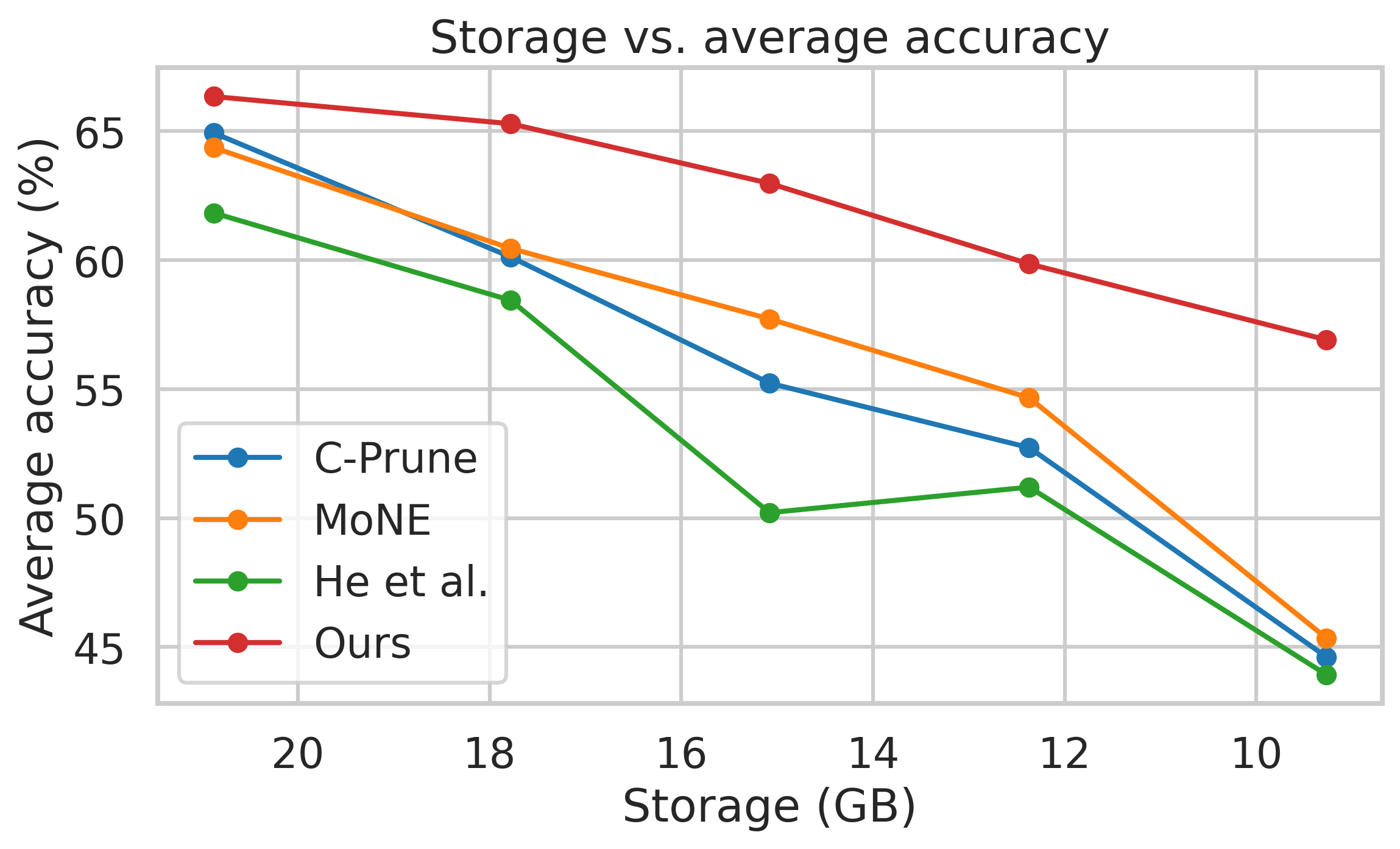}
\caption{Pareto frontier of average downstream-task accuracy versus compressed model storage (GB) for Qwen1.5-MoE-A2.7B. Our channel-level pruning consistently dominates expert-level baselines across the full compression range.}
\label{fig:pareto_frontier}
\end{figure}

We provide the full per-task accuracy for channel-level pruning and expert-level pruning baselines under matched storage budgets. Across moderate-to-aggressive budgets (25\%--75\%), channel-level pruning consistently stays on the Pareto frontier in~\cref{fig:pareto_frontier}. Under the mildest 13.3\% pruning setting, the channel budget is loose enough that expert-level pruning remains competitive, but the advantage of channel-level allocation becomes pronounced once the compression budget tightens.

\input{tables/ablation/rebuttal_channel_vs_expert_full}

\subsubsection{Wider Pruning--Quantization Combinations}
\label{app:sec:rebuttal_pq_sweep}

Based on our current experiments, \TypeP[25\%]~\TypeQ[4b] gives the best accuracy-efficiency tradeoff among the default deployment-oriented settings, but we do not claim it is a universal global optimum. We sweep a wider range of $P/Q$ combinations on Qwen1.5-MoE-A2.7B in~\cref{tab:app:pq_sweep}. Stronger compression gives lower storage but larger accuracy drop, while milder compression preserves accuracy better. The framework therefore supports flexible operating points depending on deployment constraints.

\input{tables/ablation/rebuttal_pq_sweep}

\subsubsection{Speedup and Memory Usage with Different Alignment Granularity}
\label{app:sec:exp:alignment_speedup_and_memory}

\cref{fig:speedup_memory_heatmap} reports throughput (tokens/s) and runtime peak memory of Qwen1.5-MoE-A2.7B as a function of the minimum kept-channel threshold $m$ (rows) and the alignment block size $a$ (columns) used in our Alignment-Aware Redistribution.

Across all settings, combining channel pruning with 4-bit quantization substantially reduces peak memory compared to the unpruned baseline, confirming that AAR successfully integrates structural sparsity with low-bit storage.
Larger block sizes $a$ align channel counts to coarser multiples, which enables larger and more regular GEMM kernels and is accordingly reflected in higher throughput.
However, coarser alignment reduces the degree to which each expert's channel count tracks the original CBA solution, so there is a natural trade-off between kernel efficiency and allocation fidelity.

The minimum-channel threshold $m$ controls the smallest permissible expert width after alignment.
Small $m$ allows very thin experts to survive, which can hurt throughput due to irregular kernel sizes, while large $m$ forces low-importance experts to retain more channels than necessary, slightly increasing memory.
The heatmap shows that $m=128$ with $a=128$ or $a=256$ achieves a favorable balance: throughput is near the maximum achievable value, and memory remains well below the unpruned baseline.
These are the default settings used throughout the main experiments.

\input{figures/6_speedup_memory_heatmap_qwen1.5}

\subsubsection{Calibration Runtime Breakdown}
\label{app:sec:calibration_runtime_breakdown}
\input{tables/ablation/calibration_time_breakdown}

We provide a time breakdown of the calibration process in Table~\ref{tab:build_masks_time_breakdown}, demonstrating the efficiency of the proposed method. Even for a large MoE model such as Qwen3-30B-A3B, the total calibration time remains within 10 seconds.

\subsection{Further Ablation Studies on Proposed Methods}
\label{app:sec:further_ablations}

\subsubsection{Channel Score Metric Selection}
\label{app:sec:channel-score-metric}

In this ablation, we only change the channel score definition ($s_{\ell,c}$) while keeping all other components of the pipeline fixed. 
\paragraph{Definition of metrics.} 
\begin{itemize}
\item \textbf{Weight Magnitude (channel-wise L2 norm).}
For expert $e$ in layer $\ell$ and a projection $\phi$ with weight matrix
$W_{\ell,e}^{(\phi)} \in \mathbb{R}^{O_\phi \times I_\phi}$, we define the importance of input channel $c$ by the L2 norm of the corresponding weight column:
\begin{equation}\label{app:eq:metric:weight}
  s_{\ell,e,c}^{(\phi, W)} = \lVert W^{(\phi)}_{\ell,e,c} \rVert_2^{O_{\phi}}
  = \Big( \sum_{o\in O_\phi} \big(W_{\ell,e,o,c}^{(\phi)}\big)^2 \Big)^{\frac{1}{2}}.
\end{equation}

 \item \textbf{Activation Magnitude (channel-wise L2 norm).}
For expert $e$ in layer $\ell$ and a projection $\phi$, let $A^{(\phi)}_{\ell,e,t,c}$ denote the activation of channel $c$ at token $t$. We compute the channel magnitude by an L2 norm over $\mathcal T$ tokens:
\begin{equation}\label{app:eq:metric:act}
  s_{\ell,e,c}^{(\phi, A)} = \lVert A^{(\phi)}_{\ell, e, c}  \rVert_2^{\mathcal T}
  = \Big( \sum_{t \in \mathcal{T}} {\big(A_{\ell,e,t,c}^{(\phi)}\big)}^2 \Big)^{\frac{1}{2}}. 
\end{equation}

\item \textbf{Weight$\times$Activation~\cite{wanda}.}
We follow Wanda to combine the magnitude of weight with per-channel activation. 
We compute the inner-product, and then reducing along the output dimension. 
\begin{equation}\label{app:eq:metric:wa}
s_{\ell,e,c}^{(\phi, \, WA)}
= \sum_{o\in O_{\phi}} \big( \big|W_{\ell,e,o,c}^{(\phi)}\big| \cdot \lVert A_{\ell,e,c}^{(\phi)} \rVert_2^{\mathcal T} \big). 
\end{equation}

\item \textbf{Gradient Saliency Map~\cite{song2019attribution}.}
We use a small calibration set, and collect the activation gradient by forward and backward propagation. 
Let $g_{\ell,e,t,c}^{(\phi,A)} = \nabla_{A_{\ell,e,t,c}^{(\phi)}} \mathcal{L}$ denote the gradient w.r.t.\ the activation of channel $c$ in expert $e$ and projection $\phi$ at token $t$.
We aggregate the gradients with an L2 norm over the token dimension:
\begin{equation}\label{app:eq:metric:grad}
  s_{\ell,e,c}^{(\phi, g)}
  = \big\lVert g_{\ell,e,c}^{(\phi,A)} \big\rVert_2^{\mathcal T}
  = \Big( \sum_{t \in \mathcal{T}} \big(g_{\ell,e,t,c}^{(\phi,A)}\big)^2 \Big)^{\frac{1}{2}}.
\end{equation}

\item \textbf{Activation$\times$Gradient Saliency Map~\cite{song2019attribution}.}
We use the element-wise product between activation and its gradient to score channel importance.
We compute the absolute value and then average over token dimension:
\begin{equation}\label{app:eq:metric:ag}
  s_{\ell,e,c}^{(\phi, \, gA)}
  = \frac{1}{|\mathcal{T}|} \sum_{t \in \mathcal{T}}
  \Big| A_{\ell,e,t,c}^{(\phi)} \cdot g_{\ell,e,t,c}^{(\phi,A)} \Big|. 
\end{equation}

\item \textbf{SNIP First-order Sensitivity (Weight$\times$Gradient) \cite{snip}.}
We follow SNIP to score channels by the first-order Taylor approximation, using the element-wise product between weight and its gradient. 
Let $g_{\ell,e,o,c}^{(\phi, W)} = \nabla_{W_{\ell,e,o,c}^{(\phi)}} \mathcal{L}$ denote the gradient w.r.t. the weight of input channel $c$ in expert $e$ and projection $\phi$ and output channel $o$, which  is obtained by backpropagating the loss on the calibration set.  
We compute the absolute value and then reduce along the output dimension:
\begin{equation}\label{app:eq:metric:wg}
  s_{\ell,e,c}^{(\phi, \, Wg)}
  = \sum_{o\in O_{\phi}}
  \Big| W_{\ell,e,o,c}^{(\phi)} \cdot g_{\ell,e,o,c}^{(\phi, W)} \Big|. 
\end{equation}
\end{itemize}

\input{tables/ablation/channel_score_metric}

As shown in \cref{tab:channel_score_metric}, the choice of channel scores has a non-trivial impact on performance: results on the commonsense task (ARC-c) varies moderately across metrics, whereas the gap becomes substantial on reasoning-heavy benchmarks (GSM8K and HumanEval). For example, on GSM8K, \textit{Activation} reaches 58.2, while \textit{Weight} drops to 25.9, and other weight or gradient based variants also lag behind. Overall, we use \textit{Activation} as the default channel scoring metric in all experiments.

\subsubsection{First-order vs.\ Second-order Attribution Score}
\label{app:sec:rebuttal_second_order}

Our default attribution-based score~$s_e^{(1)}$ is a first-order Taylor approximation, motivated by computational efficiency. To verify that this approximation does not compromise allocation quality, we additionally implement a lightweight but \emph{exact} second-order proxy~$s_e^{(2)}$ by perturbing each expert output with a scalar~$\alpha_e$ and benchmark both against the true ablated score $s_e^{(\mathrm{true})}=\mathcal{L}_e(0)-\mathcal{L}_e(1)$.
\cref{tab:second_order_proxy} shows that~$s_e^{(1)}$ already correlates highly with~$s_e^{(\mathrm{true})}$ (Pearson $0.959$; channel-allocation Pearson $0.966$), while~$s_e^{(2)}$ matches it almost exactly. The second-order proxy improves the end-to-end average by $+1.2$\%, at the cost of $\sim$17$\times$ longer calibration. The first-order score thus remains a strong default, and the second-order proxy serves as an enhanced variant whenever the additional calibration budget is acceptable.

\input{tables/ablation/rebuttal_second_order}

% \subsection{Reproduced and Quantized Baselines}
% \label{app:sec:rebuttal_reproduced_baselines}

% To complement~\cref{tab:overall-results-qwen} and~\cref{tab:overall-results-ds} in the main text, we reproduce publicly available baselines (MoNE, \citeauthor{he2025towards}, C-Prune) under matched settings to Ours and Ours$_Q$. Additionally, we apply the same NF4 4-bit quantization to compatible baselines (Wanda, MoNE, C-Prune). Results are reported in~\cref{tab:app:reproduced_baselines} and~\cref{tab:app:quantized_baselines}. Across all matched settings, Ours and Ours$_Q$ remain consistently ahead of the reproduced baselines.

% \input{tables/ablation/rebuttal_reproduced_baselines}
% \input{tables/ablation/rebuttal_quantized_baselines}

\subsubsection{Loss Smoothing}
\label{app:sec:loss_smoothing}

\paragraph{Raw Loss vs.\ Smoothed Losses as the Target Coverage Ratio}
\label{app:sec:raw_vs_smoothed_loss}

\cref{fig:loss_smooth_colorbar} illustrates how the square-root smoothing transforms the raw layerwise loss into a more balanced coverage target, and how that target translates into the final channel keep ratio.

The top colorbar shows the raw ablated loss per layer, which spans a wide dynamic range: a small number of critical layers dominate the signal while most layers contribute only modestly.
Directly using raw loss as the inter-layer importance signal would therefore concentrate the retained budget on a few layers and drastically under-budget the rest.
After applying square-root smoothing (middle colorbar), the dynamic range is compressed: the most sensitive layers are down-weighted, and moderately sensitive layers receive a proportionally larger share of the budget.
The resulting score-coverage targets are more uniformly distributed across layers, enabling a stable and globally balanced pruning allocation.

The bottom colorbar shows the actual channel keep ratio produced by the coverage-maximized allocation under this smoothed target.
Layers that are assigned a higher coverage ratio (darker color) retain more of their channels, while layers with highly concentrated scores can meet the same target with a smaller fraction.
Comparing the middle and bottom colorbars illustrates the decoupling between coverage target and channel count that is central to our method: a high coverage target does not imply a large channel budget when the score distribution is concentrated.

\input{figures/5_loss_smooth_colorbar}

\paragraph{Alternative Smoothing Functions for Layerwise Loss}
\label{app:sec:rebuttal_smoothing}

The square-root smoothing used in inter-layer allocation is not a theoretically essential component; rather, it is a simple realization of monotone-concave dynamic-range compression. Without smoothing, a few high-loss layers capture most of the channel budget while low-loss layers are over-pruned. Applying any monotone-concave transform reduces this imbalance by suppressing outlier values while preserving the relative ordering of layers.

We compare the default $\sqrt{\cdot}$ smoothing against three standard alternatives. Let $x \geq 0$ denote the raw layerwise loss, $\mu$ and $\sigma$ the mean and standard deviation of the losses across all layers.

\begin{itemize}[leftmargin=1.5em,itemsep=2pt]
  \item \textbf{Square-root (ours):}
    $g(x) = \sqrt{x}$.

  \item \textbf{Log smoothing:}
    $g(x) = \log(1 + \alpha x)$, \quad with $\alpha = 5$.

  \item \textbf{Huber-style smoothing:}
    \begin{equation*}
      g(x) =
      \begin{cases}
        x, & x \le \delta, \\
        \delta + \sqrt{\delta\,(x - \delta)}, & x > \delta,
      \end{cases}
      \qquad \delta = \mu + 0.5\,\sigma.
    \end{equation*}

  \item \textbf{Clip-based smoothing:}
    $g(x) = \operatorname{clip}(x,\; \mu - k\sigma,\; \mu + k\sigma)$, \quad with $k = 0.5$.
\end{itemize}

All four functions are monotone (preserving relative layer ordering) and concave (compressing the dynamic range). All smoothed variants substantially outperform the unsmoothed baseline as shown in~\cref{tab:app:smoothing}, supporting the need for dynamic-range compression. The square-root transform gives the best result while requiring no hyperparameter tuning.

\input{tables/ablation/rebuttal_smoothing}

\begin{figure}[h]
\centering
\vspace{-0.1in}
\includegraphics[width=0.44\textwidth]{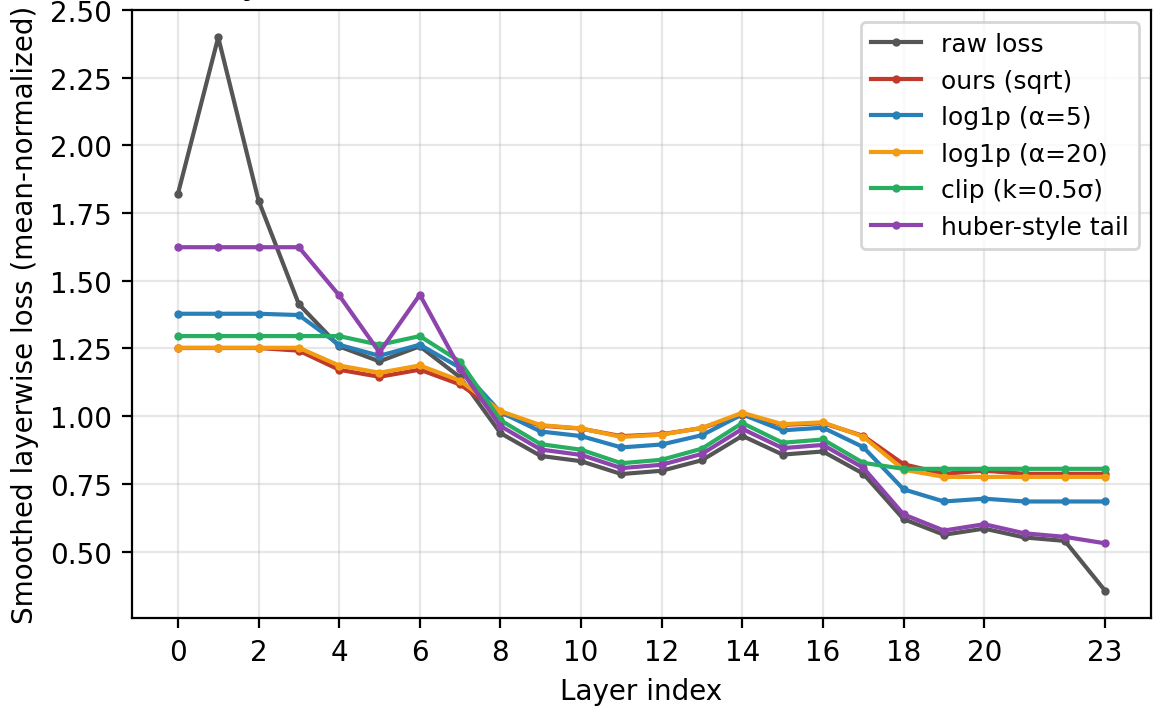}
\vspace{-0.05in}
\caption{Smoothed layerwise loss under different monotone-concave smoothing functions.}
\vspace{-0.1in}
\label{fig:smooth_fn}
\end{figure}

\subsubsection{Hyperparameter Sensitivity in CBA and AAR}
\label{app:sec:rebuttal_hyperparams}

For Coverage-Maximized Budget Allocation, we set the maximum number of binary-search iterations to 50. In practice, the search usually converges within 30 iterations, and the maximum only serves as a safeguard. For Alignment-Aware Redistribution, the minimum kept-channel threshold $m$ prevents overly thin experts and is constrained by hardware block size. We ablate $m\in\{64,128,256,512\}$ in~\cref{tab:app:hyperparam_m}; $m=128$ provides the best trade-off and is used by default. The residual reallocation strategy is summarized in the main text in~\cref{tab:aar_reallocation}, with full per-task results in~\cref{tab:app:aar_reallocation_full}.

\input{tables/ablation/rebuttal_hyperparams_m}
\input{tables/ablation/rebuttal_aar_reallocation_full}

\subsection{Sensitivity  and Robustness Analysis}
\label{app:sec:sensitivity_robustness}

\subsubsection{Sensitivity to Calibration Corpus}
\label{app:sec:rebuttal_calibration}

Our default setup follows common post-training compression practice, using C4 for general tasks, GSM8K for math, and OpenCodeReasoning for code. To examine the sensitivity systematically, we conduct an ablation on six calibration corpora: WikiText2, C4, Pile, RedPajama, GSM8K (train), and OpenCodeReasoning. Results in~\cref{tab:app:calibration_sensitivity} show that general tasks are relatively robust to general-domain corpora, while domain-specific tasks benefit from domain-matched calibration. The sensitivity is therefore structured rather than arbitrary.

\input{tables/ablation/rebuttal_calibration}

\subsubsection{Robustness Across Routing Policies}
\label{app:sec:rebuttal_routing}

\begin{figure}[t]
\centering
\vspace{-0.1in}
\includegraphics[width=0.9\textwidth]{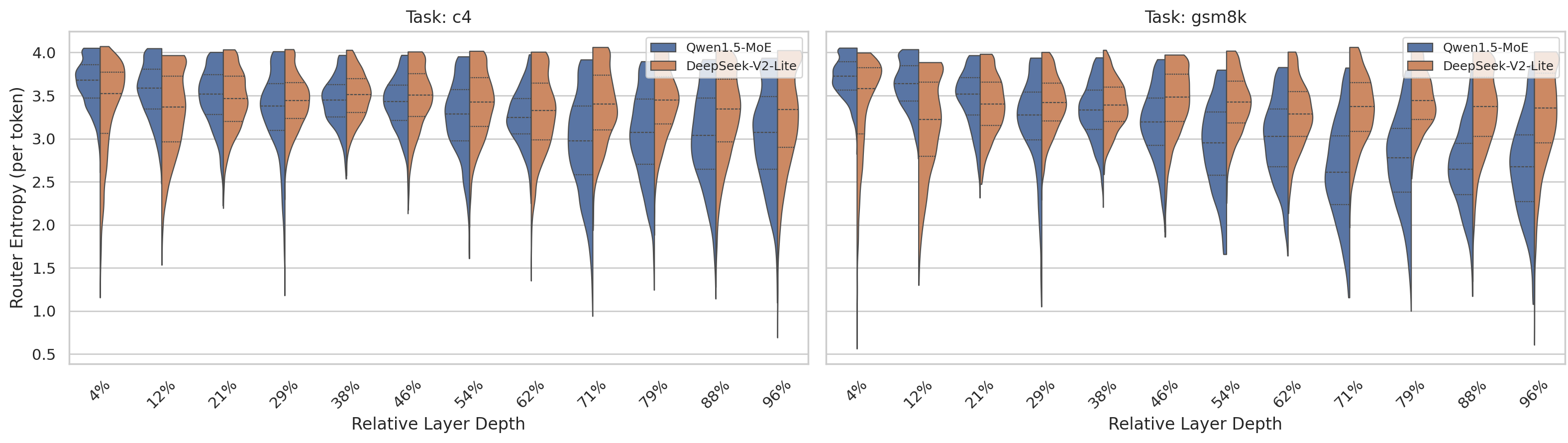}
\vspace{-0.05in}
\caption{Router entropy distributions across tasks, layer depths, and MoE models evaluated in the main text. The distributions illustrate the routing-dynamics variation used to evaluate robustness across architectures and tasks.}
\vspace{-0.1in}
\label{fig:router_entropy}
\end{figure}

Our method is not tied to a specific routing design. The main-text experiments cover Qwen-style standard top-$k$ routing and DeepSeek-style routing with load-balancing considerations. \cref{fig:router_entropy} visualizes router entropy across tasks, layer depths, and MoE models, showing that the evaluated settings cover different routing dynamics rather than a single homogeneous pattern.

\begin{figure}[t]
\centering
\includegraphics[width=0.7\textwidth]{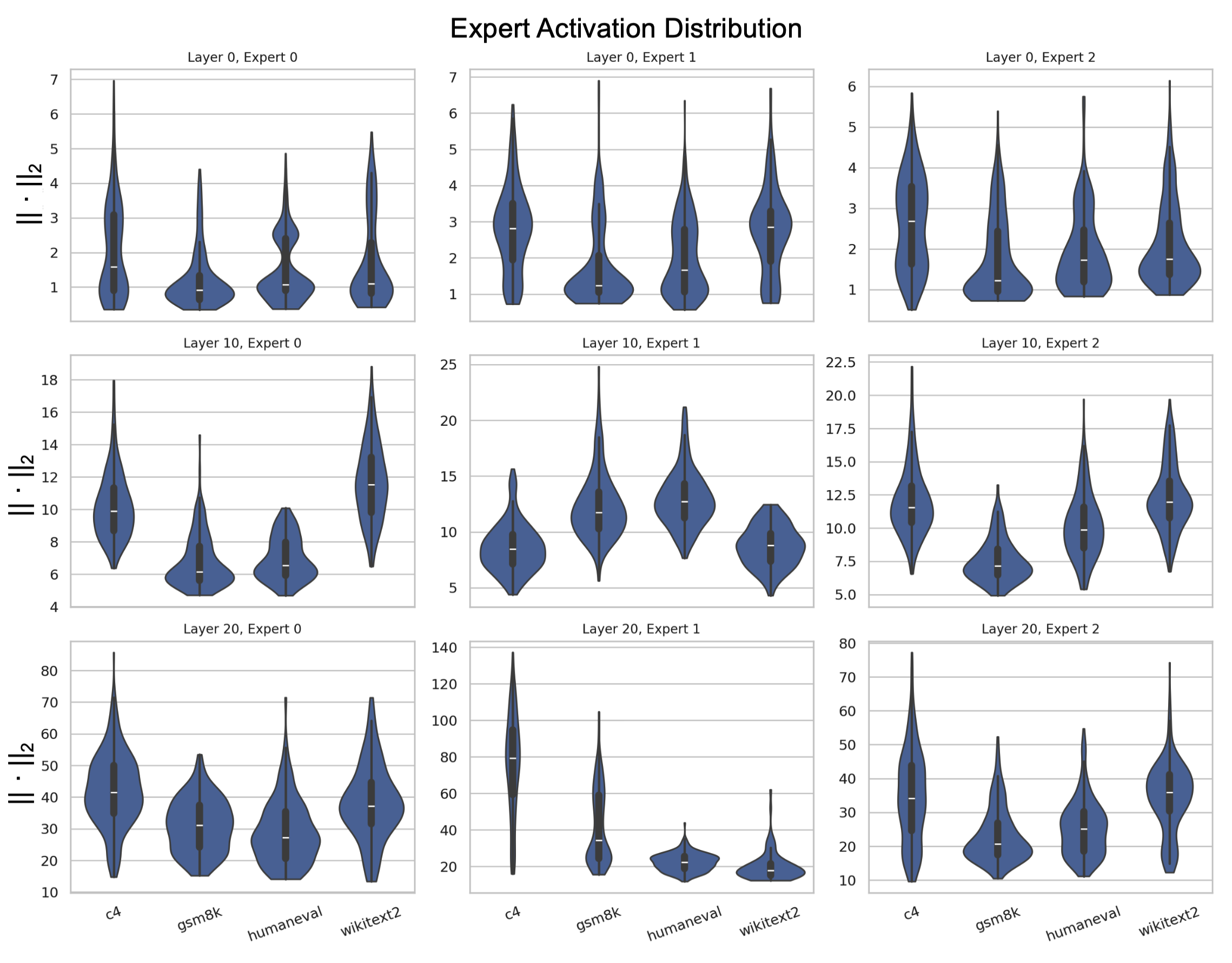}
\caption{Distribution of expert activation magnitudes across representative shallow, middle, and deep layers of Qwen1.5-MoE-A2.7B under different calibration corpora. The figure supports the calibration robustness analysis by showing that activation heterogeneity persists across data sources.}
\label{fig:expert_act_dist}
\end{figure}

\cref{fig:expert_act_dist} further reports expert activation magnitudes across representative shallow, middle, and deep layers of Qwen1.5-MoE-A2.7B under different calibration corpora. The distributions differ across both layers and corpora, confirming that expert heterogeneity is not an artifact of a single calibration source.

To further evaluate robustness under different routing budgets, we switch the activated experts from top-$2$ to top-$1$ on Qwen1.5-MoE-A2.7B and DeepSeek-V2-Lite. As expected, accuracy drops when fewer experts are activated, but our method still preserves most of the original performance at 50\% pruning under both top-1 and top-2 routing.

\input{tables/ablation/rebuttal_routing}

\subsection{Visualizations}
\label{app:sec:visualizations}

\subsubsection{Visualization of Loss, Scores and Sparsity Allocation at Expert-Level}
\label{app:sec:visual-loss-qwen1.5}

\cref{fig:qwen1.5-stacked-bar,fig:qwen3-stacked-bar} visualize the expert-level channel-score distribution and the resulting allocation for Qwen1.5-MoE-A2.7B and Qwen3-30B-A3B, respectively.
Each bar stack represents one expert: darker segments at the top correspond to high-scoring channels, while lighter segments reflect channels with smaller scores.
The fraction of dark segments therefore indicates how concentrated an expert's score is. Experts whose information is packed into a small number of channels exhibit darker, narrower stacks.

The red lines report the fraction of channels retained after coverage-maximized allocation.
Experts with a highly concentrated score distribution (tall dark segments, small light tails) are assigned a smaller channel budget, because a high coverage target can be met by keeping only the top channels.
Conversely, experts with flat, spread-out distributions require a larger kept-channel fraction to reach the same coverage threshold.
The yellow diamond markers show the attribution score of each expert, which controls the per-expert coverage target in our intra-layer allocation.
Experts with higher attribution scores receive a tighter coverage target (more channels retained) to avoid degrading high-contribution experts, whereas low-attribution experts are compressed more aggressively.

For Qwen3-30B-A3B, \cref{fig:qwen3-stacked-bar} further compares two channel-score metrics: activation-based scores (panels a, b) and weight-gradient-based scores (panels c, d). Both metrics lead to similar kept-channel allocations (red lines), confirming that the coverage-maximized allocation is robust to the choice of scoring metric.

\input{figures/7_qwen1.5-stacked-bar}

\input{figures/8_stacked_bar-qwen3}

\section{More Related Works}
\label{app:sec:related_work}

\noindent\textbf{Efficient MoE.}
MoE compression and acceleration have attracted increasing interest as models continue to grow in scale, from Mixtral 8$\times$7B with 8 activated experts~\cite{jiang2024mixtral}, to Qwen3-235B-A22B with 128 experts among which 8 are activated for each token~\cite{qwen3technicalreport}. 

Existing methods can be broadly categorized into four major techniques.
(1) \textbf{Expert trimming} removes a subset of experts through data driven selection, so that low contribution experts are never loaded or computed, reducing both memory footprint and computations~\cite{liu2024eep,bai2025diep,muzio2024seermoe,chowdhury2024aprovably,dong2025domain,lu2024notall,Zhang2025MoNERR}. (2) \textbf{Expert skipping} is a complementary approach that retains the full expert pool while skipping the computation of low importance experts at inference time, typically through routing thresholds or dynamic gating~\cite{liu2024eep,bai2025diep,lu2024notall,xu2025mcmoe,Chen2025EACMoEEA}. (3) \textbf{Expert slimming} compresses the internal structure and parameter of each expert by pruning, quantization, or low rank decomposition, while keeping the number of experts fixed~\cite{yang2024moei2,he2025towards,Lee2024STUNSP,xie2024moepruner,xu2025mcmoe,Chen2025CollaborativeCF,Chen2025EACMoEEA}. (4) \textbf{Expert merging} clusters experts with similar behavior or activation patterns and combines them into fewer experts by averaging, SVD based factorization, or pairwise merging strategies~\cite{zhang2024diversify,Li2025SubMoEEM,zhao2025puzzlemoe,c-prune}.

Despite the diverse aspects, most existing works only rank experts at the granularity of entire experts and do not explicitly analysis the redundancy within each expert. MoE-I$^2$~\cite{yang2024moei2} reduces the parameters via low rank decomposition and assigns higher ranks to more important experts while using lower ranks for less important ones. However, the speedup is limited: 
the fragmentation into small kernels makes it difficult to reach peak throughput of one larger kernel, introducing additional overhead in kernel launching, cache hit, and memory access. Chen et al.~ \cite{Chen2025CollaborativeCF} quantize all parameters to low bitwidth, compare the reconstruction error, and assign higher bitwidth to experts that are more sensitive to quantization. This strategy, however, only feasible to methods that have a small search space, e.g., 4/8 bitwidth, which is insufficient for fine grained expert-wise compression budget allocation. 

\noindent\textbf{Expert Importance.} 
A key driver behind MoE efficiency designs is the highly unbalanced contribution of different experts. This has motivated a variety of methods for measuring expert importance. 
(1) \textbf{Router based} statistics are widely used, including average gate scores, processed token counts (expert hit rates), and gate variations during fine tuning~\cite{he2025towards,Lee2024STUNSP,xie2024moepruner,muzio2024seermoe,chowdhury2024aprovably,dong2025domain,Li2025SubMoEEM,lu2024notall,xu2025mcmoe,Chen2025CollaborativeCF}. 
(2) \textbf{Activation based} metrics such as gate weighted outputs or activation saliency are also employed~\cite{dong2025domain,Li2025SubMoEEM,Zhang2025MoNERR,zhao2025puzzlemoe}. 
(3) \textbf{Loss or accuracy based} criteria measure the performance drop when removing a particular expert or a subset of experts. For example, they quantify the impact on reconstruction loss or downstream task performance after compression~\cite{liu2024eep,yang2024moei2,zhang2024diversify,lu2024notall}. 
(4) \textbf{Learnable method} is recently proposed, which learns a set of importance scalars that are jointly optimized during fine tuning~\cite{bai2025diep}. 

% 上述指标层间不可比
However, one limitation is that, router based and performance based statistics are often \textit{not comparable across layers}. Routers and activations in different layers may behave very different in decision patterns, or follow different distributions. Loss values can be depth dependent and unstable under different experimental setups, including the source of calibration data, the loss function, and the tokenization scheme. As a result, many previous methods resort to assigning a uniform compression ratio to all layers instead of performing cross layer importance comparison. 

% 粗粒度，只能排序重要性，不能精确分配稀疏率，并且比较耗时。
A second limitation is that most existing metrics exhibit \textit{a high dynamic range} and are primarily used to rank experts and entirely trim the least important $k$ experts, rather than to support precise allocation of compression ratios. Only a few works explore redundancy within experts. MoE-I$^2$ and Liu et al.~\cite{liu2024eep} remove a small group of experts at a time and compare the resulting loss increase or accuracy drop in order to infer expert importance within each layer. But such loss or accuracy based methods are only feasible for relatively small MoE models with a limited number of experts. When the search space over layers and experts grows, even greedy or genetic strategies incur prohibitively high computational cost, which severely restricts their applicability to modern large scale MoE architectures.

%% file: algorithms/3_prefix_sum.tex
\begin{algorithm}[h]
  \caption{Prefix sums $\mathcal{S}_g(n)$ for a group $g$ (a layer or an expert)}
  \label{algo:prefix_sum}
  \begin{algorithmic}[1]
    \STATE {\bfseries Input:} sorted channel scores $s_{g,(1)} \ge s_{g,(2)} \ge \cdots \ge s_{g,(|\mathcal C_g|)}$ for group $g$; total channel number $|\mathcal C_g|$
    \STATE {\bfseries Output:} prefix sums $\mathcal S_g(n)$ for $n = 1, \dots, |\mathcal C_g|$

    \STATE $\mathcal S_g(0) \leftarrow 0$
    \FOR{$n = 1$ {\bfseries to} $|\mathcal C_g|$}
      \STATE $\mathcal S_g(n) \leftarrow \mathcal S_g(n - 1) + s_{g,(n)}$
    \ENDFOR
    \STATE {\bfseries return} $\mathcal S_g(n)$ for all $n = 1, \dots, |\mathcal C_g|$
  \end{algorithmic}
\end{algorithm}

%% file: algorithms/2_layer_min_channels.tex
\begin{algorithm}[h]
  \caption{Binary search of minimal channels $N_g(\rho)$ for target coverage $\rho$ in group $g$}
  \label{algo:layer_min_channels}
  \begin{algorithmic}[1]
    \STATE {\bfseries Input:} prefix sums $\mathcal S_g(n)$; target coverage $\rho \in [0,1]$; total score $\mathrm S^{tot}_g$; channel number $|\mathcal C_g|$
    \STATE {\bfseries Output:} minimal number of channels $N_g(\rho)$

    \STATE $n_{\min} \leftarrow 1,\; n_{\max} \leftarrow |\mathcal C_g|$
    \WHILE{$n_{\min} < n_{\max}$}
      \STATE $n \leftarrow \left\lfloor (n_{\min} + n_{\max})/2 \right\rfloor$
      \IF{$\mathcal S_g(n) \ge \rho \, \mathrm S^{tot}_g$}
        \STATE $n_{\max} \leftarrow n$
      \ELSE
        \STATE $n_{\min} \leftarrow n + 1$
      \ENDIF
    \ENDWHILE
    \STATE $N_g(\rho) \leftarrow n_{\min}$
    \STATE {\bfseries return} $N_g(\rho)$
  \end{algorithmic}
\end{algorithm}

%% file: algorithms/app_4_expert_coverage.tex
\begin{algorithm}[h]
  \caption{Intra-layer coverage search at layer $\ell$ with an expert-wise importance prior}
  \label{algo:expert_coverage}
  \begin{algorithmic}[1]
    \STATE {\bfseries Input:} experts $\mathcal E_\ell=\{1,\ldots,E\}$ at layer $\ell$; non-negative importance prior $\{\phi_{\ell,e}\}_{e\in\mathcal E_\ell}$; prefix sums $\{\mathcal S_{\ell,e}(n),\, \mathrm S^{tot}_{\ell,e}\}_{e\in\mathcal E_\ell}$; per-expert channel counts $\{|\mathcal C_{\ell,e}|\}_{e\in\mathcal E_\ell}$; layer budget $N^\star_\ell$; tolerance $\varepsilon$
    \STATE {\bfseries Output:} expert-wise channel budgets $\{N^\star_{\ell,e}\}_{e\in\mathcal E_\ell}$

    \STATE $N^{tot}_\ell \leftarrow \sum_{e\in\mathcal E_\ell} |\mathcal C_{\ell,e}|$
    \STATE $\alpha_{\min} \leftarrow 0,\;\alpha_{\max} \leftarrow 1$

    \WHILE{$\alpha_{\min} < \alpha_{\max}$}
      \STATE $\alpha \leftarrow (\alpha_{\min} + \alpha_{\max})/2$
      \FOR{each expert $e\in\mathcal E_\ell$}
        \STATE $\rho_{\ell,e} \leftarrow \min(\alpha \phi_{\ell,e},\, 1)$
        \STATE $N_{\ell,e}(\rho_{\ell,e}) \leftarrow \min\{n \mid \mathcal S_{\ell,e}(n) \ge \rho_{\ell,e}\, \mathrm S^{tot}_{\ell,e}\}$ \hfill (\cref{algo:layer_min_channels})
      \ENDFOR
      \STATE $N_\ell(\boldsymbol\rho) \leftarrow \sum_{e\in\mathcal E_\ell} N_{\ell,e}(\rho_{\ell,e})$

      \IF{$\bigl| N_\ell(\boldsymbol\rho) - N^\star_\ell \bigr| \le \varepsilon \, N^{tot}_\ell$}
        \STATE $N^\star_{\ell,e} \leftarrow N_{\ell,e}(\rho_{\ell,e})$, $\forall e\in\mathcal E_\ell$
        \STATE {\bfseries break}
      \ENDIF

      \IF{$N_\ell(\boldsymbol\rho) > N^\star_\ell$}
        \STATE $\alpha_{\max} \leftarrow \alpha$
      \ELSE
        \STATE $\alpha_{\min} \leftarrow \alpha$
      \ENDIF
    \ENDWHILE

    \STATE {\bfseries return} $\{N^\star_{\ell,e}\}_{e\in\mathcal E_\ell}$
  \end{algorithmic}
\end{algorithm}

%% file: tables/ablation/rebuttal_channel_vs_expert_full.tex
\begin{table}[t]
\centering
\caption{Per-task accuracy of channel-level (Ours) vs.\ expert-level pruning baselines on Qwen1.5-MoE-A2.7B at matched storage budgets. Our method performs channel-level structural pruning, whereas the competing baselines adopt expert-level pruning.}
\label{tab:app:channel_vs_expert_pareto_full}
\resizebox{1.0\textwidth}{!}{
\begin{tabular}{lcccccccccc}
\toprule
\textbf{Method} & \textbf{Expert-level} & \textbf{Prune ratio (\%)} & \textbf{Storage (GB)} & \textbf{ARC-c} & \textbf{ARC-e} & \textbf{HellaSwag} & \textbf{PIQA} & \textbf{BoolQ} & \textbf{WinoGrande} & \textbf{Avg} \\
\midrule
C-Prune   & \cmark & 13.3 & 23.59 & 44.71 & 69.15 & 77.26 & 80.36 & 79.48 & 68.90 & \textbf{69.98} \\
MoNE      & \cmark & 13.3 & 23.59 & 41.55 & 63.68 & 77.07 & 80.03 & 76.51 & 68.51 & 67.89 \\
He et al. & \cmark & 13.3 & 23.59 & 40.54 & 65.95 & 63.78 & 82.70 & 74.05 & 67.57 & 65.77 \\
Ours      & \xmark & 13.3 & 23.59 & 42.70 & 65.95 & 65.59 & 80.90 & 76.40 & 70.81 & 67.06 \\
\midrule
C-Prune   & \cmark & 25.0 & 20.88 & 40.00 & 62.70 & 63.06 & 78.92 & 77.12 & 67.75 & 64.93 \\
MoNE      & \cmark & 25.0 & 20.88 & 40.44 & 60.73 & 64.14 & 81.20 & 71.53 & 68.11 & 64.36 \\
He et al. & \cmark & 25.0 & 20.88 & 37.30 & 59.64 & 61.80 & 81.08 & 67.93 & 63.24 & 61.83 \\
\textbf{Ours} & \xmark & \textbf{25.0} & \textbf{20.88} & \textbf{42.16} & \textbf{65.41} & \textbf{63.24} & \textbf{79.10} & \textbf{78.20} & \textbf{69.91} & \textbf{66.34} \\
\midrule
C-Prune   & \cmark & 38.3 & 17.78 & 34.59 & 58.02 & 60.72 & 76.22 & 69.55 & 61.62 & 60.12 \\
MoNE      & \cmark & 38.3 & 17.78 & 35.50 & 52.61 & 64.32 & 80.36 & 64.50 & 65.41 & 60.45 \\
He et al. & \cmark & 38.3 & 17.78 & 34.41 & 55.86 & 58.74 & 77.12 & 64.86 & 59.64 & 58.44 \\
\textbf{Ours} & \xmark & \textbf{38.3} & \textbf{17.78} & \textbf{40.72} & \textbf{64.14} & \textbf{63.24} & \textbf{78.92} & \textbf{75.68} & \textbf{69.01} & \textbf{65.29} \\
\midrule
C-Prune   & \cmark & 50.0 & 15.08 & 30.63 & 50.09 & 57.12 & 74.23 & 62.16 & 57.12 & 55.23 \\
MoNE      & \cmark & 50.0 & 15.08 & 32.07 & 51.71 & 60.54 & 76.58 & 63.60 & 61.80 & 57.72 \\
He et al. & \cmark & 50.0 & 15.08 & 25.95 & 45.41 & 44.86 & 69.37 & 63.06 & 52.61 & 50.21 \\
\textbf{Ours} & \xmark & \textbf{50.0} & \textbf{15.08} & \textbf{36.76} & \textbf{65.95} & \textbf{61.98} & \textbf{75.32} & \textbf{69.37} & \textbf{68.47} & \textbf{62.98} \\
\midrule
C-Prune   & \cmark & 61.7 & 12.37 & 28.29 & 44.50 & 53.87 & 69.73 & 65.95 & 54.05 & 52.73 \\
MoNE      & \cmark & 61.7 & 12.37 & 30.27 & 46.13 & 55.68 & 73.15 & 63.42 & 59.28 & 54.66 \\
He et al. & \cmark & 61.7 & 12.37 & 27.57 & 45.05 & 47.21 & 72.07 & 61.26 & 54.05 & 51.20 \\
\textbf{Ours} & \xmark & \textbf{61.7} & \textbf{12.37} & \textbf{33.15} & \textbf{58.20} & \textbf{60.00} & \textbf{73.87} & \textbf{67.39} & \textbf{66.49} & \textbf{59.85} \\
\midrule
C-Prune   & \cmark & 75.0 & 9.27 & 23.42 & 35.32 & 42.52 & 59.46 & 54.41 & 52.61 & 44.62 \\
MoNE      & \cmark & 75.0 & 9.27 & 25.95 & 36.58 & 41.62 & 61.44 & 52.97 & 53.51 & 45.35 \\
He et al. & \cmark & 75.0 & 9.27 & 21.80 & 36.04 & 37.48 & 60.36 & 54.23 & 53.69 & 43.93 \\
\textbf{Ours} & \xmark & \textbf{75.0} & \textbf{9.27} & \textbf{32.61} & \textbf{52.07} & \textbf{56.22} & \textbf{69.37} & \textbf{67.57} & \textbf{63.60} & \textbf{56.91} \\
\bottomrule
\end{tabular}
}
\end{table}

%% file: tables/ablation/rebuttal_pq_sweep.tex
\begin{table}[t]
\centering
\caption{Storage and downstream accuracy under different combinations of pruning ratio $P$ and quantization bitwidth $Q$ on Qwen1.5-MoE-A2.7B. The default Ours$_Q$ configuration is highlighted.}
\label{tab:app:pq_sweep}
\resizebox{1.0\textwidth}{!}{
\begin{tabular}{cccccccccccc}
\toprule
\textbf{P (\%)} & \textbf{Q (bits)} & \textbf{Storage (GB)} & \textbf{PIQA} & \textbf{ARC-c} & \textbf{ARC-e} & \textbf{BoolQ} & \textbf{HellaSwag} & \textbf{WinoGrande} & \textbf{GSM8K} & \textbf{HumanEval} & \textbf{Avg} \\
\midrule
0  & 16 & 28.70 & 80.79 & 40.41 & 69.44 & 70.57 & 77.17 & 69.77 & 61.50 & 34.20 & 62.98 \\
15 & 8  & 13.50 & 80.04 & 41.52 & 65.07 & 77.64 & 65.07 & 70.26 & 61.08 & 31.10 & 61.47 \\
25 & 8  & 12.25 & 79.04 & 41.52 & 64.67 & 76.85 & 63.67 & 69.66 & 62.35 & 26.83 & 60.57 \\
35 & 8  & 11.01 & 78.24 & 39.32 & 61.68 & 75.25 & 63.87 & 67.47 & 62.28 & 34.15 & 60.28 \\
45 & 8  & 9.76  & 76.65 & 39.32 & 64.07 & 72.85 & 62.28 & 69.46 & 62.48 & 34.76 & 60.23 \\
15 & 4  & 7.36  & 79.24 & 38.72 & 63.87 & 76.45 & 64.27 & 70.86 & 61.26 & 28.05 & 60.34 \\
\textbf{25} & \textbf{4} & \textbf{6.71} & \textbf{79.04} & \textbf{40.92} & \textbf{60.88} & \textbf{76.85} & \textbf{63.27} & \textbf{69.26} & \textbf{56.24} & \textbf{27.44} & \textbf{59.24} \\
35 & 4  & 6.07  & 78.24 & 37.92 & 62.28 & 77.05 & 62.87 & 70.26 & 56.89 & 31.71 & 59.65 \\
45 & 4  & 5.43  & 74.65 & 35.13 & 62.48 & 71.86 & 61.28 & 70.46 & 52.61 & 26.22 & 56.84 \\
\bottomrule
\end{tabular}
}
\end{table}

%% file: figures/6_speedup_memory_heatmap_qwen1.5.tex
\begin{figure}[t]
  \begin{center}
    \centerline{\includegraphics[width=0.5\columnwidth]{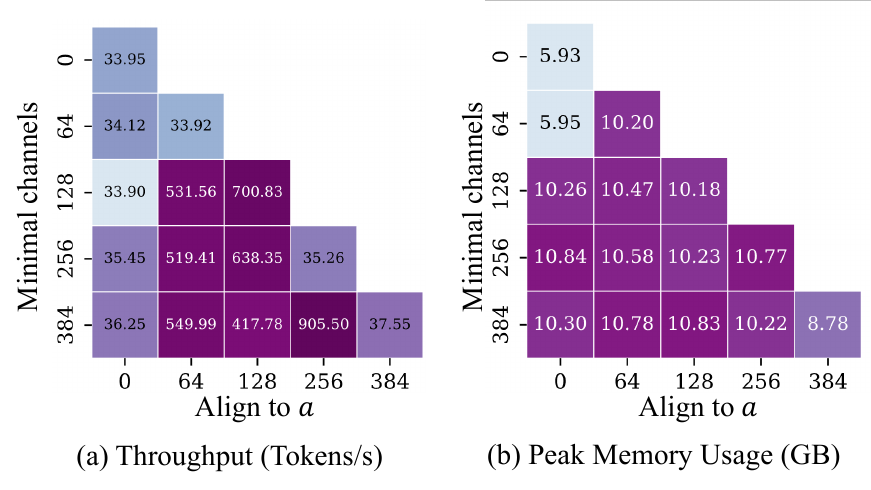}}
    \caption{
    Throughput and runtime memory usage of Qwen1.5-MoE-A2.7B with different minimal channel numbers and alignment granularity. 
    }
    \vspace{-0.2in}
    \label{fig:speedup_memory_heatmap}
  \end{center}
\end{figure}

%% file: tables/ablation/calibration_time_breakdown.tex
\begin{table}[htbp]
\centering
\caption{Time breakdown of generating prune allocation of 50\% sparsity.}
\label{tab:build_masks_time_breakdown}

% \footnotesize
\resizebox{0.45\columnwidth}{!}{
\begin{tabular}{l r r}
\toprule
\textbf{Stage} & \textbf{Time (ms)} & \textbf{Time (\%)} \\
\midrule

%%%%%%%%%%%%%%%%%%%
\multicolumn{3}{c}{\textbf{Qwen1.5-MoE-A2.7B}} \\

% Overall
Overall generating prune plan & 2078.31  & 100\% \\
\addlinespace[2pt]

% Inter-layer planning
~~ \textit{-- Inter-layer coverage search} & 63.00  & 3.03\% \\
~~~ \vbar{3} ~\gray{Smooth weights} & \gray{30.35}  & \gray{1.46\%} \\
~~~~~ \gray{Compute prefix sum} & \gray{14.16}  & \gray{0.68\%} \\
~~~~~ \gray{Binary search} & \gray{17.90}  & \gray{0.86\%} \\
\addlinespace[2pt]

% Intra-layer planning
~~ \textit{-- Intra-layer coverage search} & 1836.49  & 88.36\% \\
~~~ \vbar{2} ~\gray{Recompute prefix sum} & \gray{0.08}  & \gray{$<$0.01\%} \\
~~~~~ \gray{Binary Search (24 MoE layers)} & \gray{1836.41}  & \gray{88.36\%} \\
~~~~~ \gray{-- per layer} & \gray{76.52}  & \gray{3.68\%} \\
\addlinespace[2pt]

% Alignment-aware redistribution
~~ \textit{-- Alignment-aware redistribution} & 178.82  & 8.60\% \\
~~~ \vbar{4} ~\gray{Compute $K^\mathrm{base}$} & \gray{52.84}  & \gray{2.54\%} \\
~~~~~ \gray{Compute headroom} & \gray{14.96}  & \gray{0.72\%} \\
~~~~~ \gray{Allocate chunks} & \gray{110.14}  & \gray{5.30\%} \\
~~~~~ \gray{Clamp to $I$} & \gray{0.88}  & \gray{0.04\%} \\

\midrule

%%%%%%%%%%%%%%%%%%%
\multicolumn{3}{c}{\textbf{Deepseek-MoE-16B}} \\

% Overall
Overall generating prune plan & 2914.31  & 100\% \\
\addlinespace[2pt]

% Inter-layer planning
~~ \textit{-- Inter-layer coverage search} & 73.81  & 2.53\% \\
~~~ \vbar{3} ~\gray{Smooth weights} & \gray{30.48}  & \gray{1.05\%} \\
~~~~~ \gray{Compute prefix sum} & \gray{15.57}  & \gray{0.53\%} \\
~~~~~ \gray{Binary search} & \gray{27.14}  & \gray{0.93\%} \\
\addlinespace[2pt]

% Intra-layer planning
~~ \textit{-- Intra-layer coverage search} & 2647.70  & 90.85\% \\
~~~ \vbar{2} ~\gray{Recompute prefix sum} & \gray{0.10}  & \gray{$<$0.01\%} \\
~~~~~ \gray{Binary Search (27 MoE layers)} & \gray{2647.60}  & \gray{90.85\%} \\
~~~~~ \gray{-- per layer} & \gray{98.06}  & \gray{3.36\%} \\
\addlinespace[2pt]

% Alignment-aware redistribution
~~ \textit{-- Alignment-aware redistribution} & 192.80  & 6.62\% \\
~~~ \vbar{4} ~\gray{Compute $K^\mathrm{base}$} & \gray{56.12}  & \gray{1.93\%} \\
~~~~~ \gray{Compute headroom} & \gray{17.16}  & \gray{0.59\%} \\
~~~~~ \gray{Allocate chunks} & \gray{118.57}  & \gray{4.07\%} \\
~~~~~ \gray{Clamp to $I$} & \gray{0.95}  & \gray{0.03\%} \\

\midrule

%%%%%%%%%%%%%%%%%%%
\multicolumn{3}{c}{\textbf{Qwen3-30B-A3B}} \\

% Overall
Overall generating prune plan (total) & 10063.12 & 100\% \\
\addlinespace[2pt]

% Inter-layer planning
~~ \textit{-- Inter-layer coverage search} & 87.43 & 0.87\% \\
~~~ \vbar{3} ~\gray{Smooth weights} & \gray{29.36} & \gray{0.29\%} \\
~~~~~ \gray{Compute prefix sum} & \gray{19.91} & \gray{0.20\%} \\
~~~~~ \gray{Binary search} & \gray{37.15} & \gray{0.37\%} \\
\addlinespace[2pt]

% Intra-layer planning
~~ \textit{-- Intra-layer coverage search} & 9619.28 & 95.59\% \\
~~~ \vbar{2} ~\gray{Recompute prefix sum} & \gray{0.09} & \gray{$<$0.01\%} \\
~~~~~ \gray{Binary Search (MoE 48 layers)} & \gray{9619.19} & \gray{95.59\%} \\
~~~~~ \gray{-- per layer} & \gray{200.40} & \gray{1.99\%} \\
\addlinespace[2pt]

% Alignment-aware redistribution
~~ \textit{-- Alignment-aware redistribution} & 356.41 & 3.54\% \\
~~~ \vbar{4} ~\gray{Compute $K^\mathrm{base}$} & \gray{20.44} & \gray{0.20\%} \\
~~~~~ \gray{Compute headroom} & \gray{17.02} & \gray{0.17\%} \\
~~~~~ \gray{Allocate chunks} & \gray{317.40} & \gray{3.15\%} \\
~~~~~ \gray{Clamp to $I$} & \gray{1.55} & \gray{0.02\%} \\

\bottomrule
\end{tabular}
}
\end{table}

%% file: tables/ablation/channel_score_metric.tex
% \begin{table}[h]
% \centering
% \caption{Ablation on channel score definition $s_{\ell,c}$. Inter-layer and intra-layer allocation use the same maximum-coverage procedure, while only the channel saliency scores differ.}
% \label{tab:channel_score_metric}
% \resizebox{0.48\columnwidth}{!}{
% \begin{tabular}{lccc}
% \toprule
% \textbf{$s_{\ell,c}$} & \textbf{ARC-c} & \textbf{GSM8K} & \textbf{HumanEval} \\
% \midrule
% Weight & 38.0 & 25.9 & 18.9 \\
% \textbf{Activation} & \textbf{40.0} & \textbf{58.2} & \textbf{30.5} \\
% Gradient & 39.0 & 51.7 & 26.8 \\
% Weight$\times$Activation & 38.2 & 33.7 & 28.7 \\
% Activation$\times$Gradient & 38.7 & 46.3 & 26.2 \\
% Weight$\times$Gradient & 39.2 & 41.5 & 24.4 \\
% \bottomrule
% \end{tabular}
% }
% \end{table}

\begin{table}[h]
\centering
\caption{Ablation on channel score definition $s_{\ell,c}$. Inter-layer and intra-layer allocation use the same maximum-coverage procedure, while only the channel saliency scores differ.}
\label{tab:channel_score_metric}
\resizebox{0.70\columnwidth}{!}{
\begin{tabular}{lccc}
\toprule
\textbf{$s_{\ell,c}$} & \textbf{ARC-c} & \textbf{GSM8K} & \textbf{HumanEval} \\
\midrule
Weight (\cref{app:eq:metric:weight}) & 38.0 & 25.9 & 18.9 \\
\textbf{Activation} (\cref{app:eq:metric:act}) & \textbf{40.0} & \textbf{58.2} & \textbf{30.5} \\
Gradient Saliency (\cref{app:eq:metric:grad}) & 39.0 & 51.7 & 26.8 \\
Weight$\times$Activation (\cref{app:eq:metric:wa}) & 38.2 & 33.7 & 28.7 \\
Activation$\times$Gradient (\cref{app:eq:metric:ag}) & 38.7 & 46.3 & 26.2 \\
SNIP First-order Sensitivity (Weight$\times$Gradient) (\cref{app:eq:metric:wg}) & 39.2 & 41.5 & 24.4 \\
\bottomrule
\end{tabular}
}
\end{table}

%% file: tables/ablation/rebuttal_second_order.tex
\begin{table}[h]
\centering
\caption{Comparison of the first-order attribution score $s_e^{(1)}$ used in our main results and an exact second-order proxy $s_e^{(2)}$ on Qwen1.5-MoE-A2.7B under \TypeP[25\%]~\TypeQ[4b]. The second-order proxy is also benchmarked against the true ablated score $s_e^{(\mathrm{true})}$ computed by directly removing each expert.}
\label{tab:second_order_proxy}
% \vspace{4pt}
% \begin{minipage}[t]{0.57\linewidth}
\centering
\resizebox{0.6\linewidth}{!}{
\begin{tabular}{lcccc}
\toprule
\textbf{Comparison} & \textbf{Pearson} $\uparrow$ & \textbf{L1} $\downarrow$ & \textbf{Channel Pearson} $\uparrow$ & \textbf{Channel Diff} $\downarrow$ \\
\midrule
$s_e^{(1)}$ vs. $s_e^{(\mathrm{true})}$ & 0.959 & 0.058 & 0.966 & 23.4 \\
$s_e^{(2)}$ vs. $s_e^{(\mathrm{true})}$ & 1.000 & 0.000 & 1.000 & 0.13 \\
\bottomrule
\end{tabular}
}
% \end{minipage}

\vspace{6pt}
\resizebox{0.95\linewidth}{!}{
\begin{tabular}{lcccccccccc}
\toprule
\textbf{Score} & \textbf{PIQA} & \textbf{ARC-c} & \textbf{ARC-e} & \textbf{BoolQ} & \textbf{HellaSwag} & \textbf{WinoGrande} & \textbf{GSM8K} & \textbf{MMLU} & \textbf{Avg} \\
\midrule
First-order $s_e^{(1)}$ (default) & 74.45 & 35.93 & \textbf{66.07} & 70.46 & 61.28 & 69.26 & \textbf{58.24} & 51.43 & 60.89 \\
Second-order $s_e^{(2)}$          & \textbf{78.04} & \textbf{36.13} & 64.67 & \textbf{76.45} & \textbf{61.30} & \textbf{70.26} & 56.49 & \textbf{53.40} & \textbf{62.09} \\
\bottomrule
\end{tabular}
}
\end{table}

%% file: figures/5_loss_smooth_colorbar.tex
\begin{figure}[ht]
  \begin{center}
    \centerline{\includegraphics[width=0.67\columnwidth]{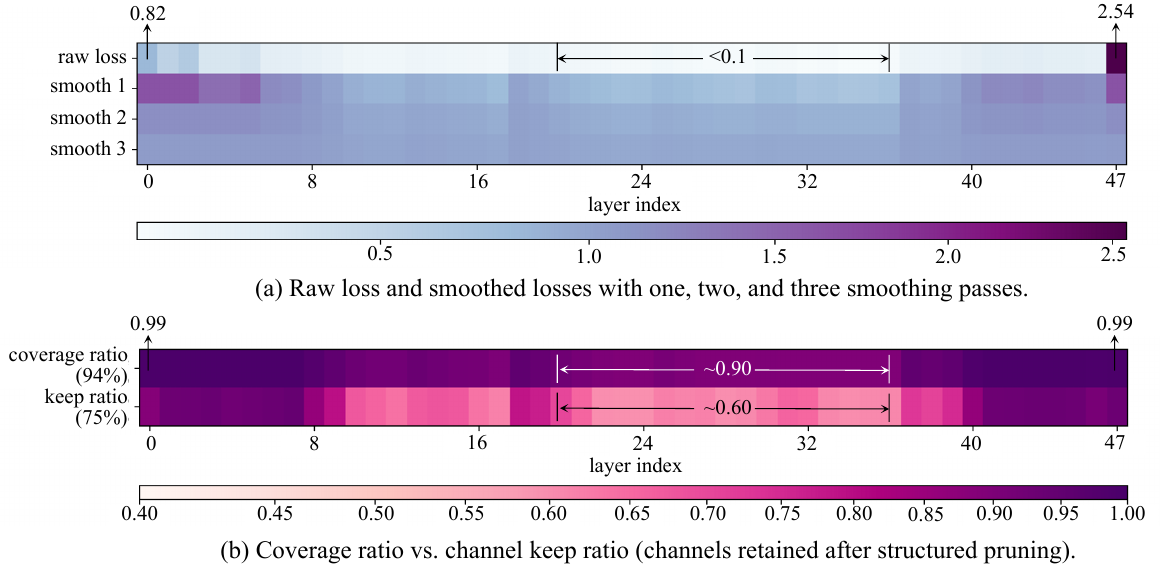}}
    \caption{
 Losses (raw and smoothed), coverage ratio and channel keep ratio after pruning. 
    }
    \vspace{-0.2in}
    \label{fig:loss_smooth_colorbar}
  \end{center}
\end{figure}

%% file: tables/ablation/rebuttal_smoothing.tex
\begin{table}[h]
\centering
\caption{Downstream accuracy with channel allocations derived from different monotone-concave smoothing functions of the layerwise loss on Qwen1.5-MoE-A2.7B under \TypeP[50\%].}
\label{tab:app:smoothing}
\resizebox{1.0\textwidth}{!}{
\begin{tabular}{llccccccccc}
\toprule
\textbf{Smooth fn.} & \textbf{Setting} & \textbf{PIQA} & \textbf{ARC-c} & \textbf{ARC-e} & \textbf{BoolQ} & \textbf{HellaSwag} & \textbf{WinoGrande} & \textbf{GSM8K} & \textbf{HumanEval} & \textbf{Avg} \\
\midrule
None             & --                       & 72.06 & 34.73 & 54.29 & 70.46 & 58.48 & 64.07 & 46.82 & 18.29 & 52.40 \\
Log              & $g(x){=}\log(1{+}5x)$    & 74.85 & 32.34 & 60.88 & 68.66 & 61.07 & 68.86 & 56.40 & 32.32 & 56.92 \\
Huber-style      & $\delta{=}\mu{+}0.5\sigma$ & 73.45 & 32.14 & 56.89 & 67.66 & 61.65 & 69.86 & 56.40 & 29.27 & 55.92 \\
Clip             & $k{=}0.5$                & 74.45 & 35.33 & 59.88 & 68.66 & 62.08 & 68.66 & 55.80 & 28.66 & 56.69 \\
\textbf{Sqrt (ours)} & \textbf{--}          & \textbf{74.45} & \textbf{35.93} & \textbf{66.07} & \textbf{70.46} & \textbf{61.28} & \textbf{69.26} & \textbf{58.24} & \textbf{30.50} & \textbf{58.27} \\
\bottomrule
\end{tabular}
}
\end{table}

%% file: tables/ablation/rebuttal_hyperparams_m.tex
\begin{table}[t]
\centering
\caption{Ablation on the minimum-channel threshold $m$ in AAR on Qwen1.5-MoE-A2.7B under \TypeP[25\%]~\TypeQ[4b].}
\label{tab:app:hyperparam_m}
\resizebox{0.7\textwidth}{!}{
\begin{tabular}{lccccccccc}
\toprule
$m$ & \textbf{PIQA} & \textbf{ARC-c} & \textbf{ARC-e} & \textbf{BoolQ} & \textbf{HellaSwag} & \textbf{GSM8K} & \textbf{HumanEval} & \textbf{WinoGrande} & \textbf{Avg} \\
\midrule
64       & 78.44 & 40.52 & 63.67 & 77.25 & 63.17 & 54.29 & 27.44 & 67.66 & 59.06 \\
\textbf{128} & \textbf{78.72} & \textbf{41.51} & \textbf{63.48} & \textbf{77.12} & \textbf{63.62} & \textbf{56.88} & \textbf{26.83} & \textbf{68.16} & \textbf{59.54} \\
256      & 78.64 & 41.32 & 63.87 & 77.25 & 63.47 & 56.69 & 26.83 & 68.26 & 59.54 \\
512      & 79.04 & 39.72 & 63.07 & 77.45 & 63.67 & 55.49 & 26.22 & 68.46 & 59.14 \\
\bottomrule
\end{tabular}
}
\end{table}

%% file: tables/ablation/rebuttal_aar_reallocation_full.tex
\begin{table}[t]
\centering
\vspace{-0.1in}
\caption{Comparison of two AAR residual reallocation strategies on Qwen1.5-MoE-A2.7B with different alignment block sizes $a$. \textit{l-r-c}: \emph{largest removed channels}, \textit{l-r-s}: \emph{largest removed scores}.}
\vspace{-0.1in}
\label{tab:app:aar_reallocation_full}
\resizebox{0.97\textwidth}{!}{
\begin{tabular}{lcccccccccc}
\toprule
\textbf{Strategy} & $a$ & \textbf{GSM8K} & \textbf{HumanEval} & \textbf{PIQA} & \textbf{ARC-c} & \textbf{ARC-e} & \textbf{BoolQ} & \textbf{HellaSwag} & \textbf{WinoGrande} & \textbf{Avg} \\
\midrule
l-r-c  & 64  & 56.29 & 26.83 & 78.44 & 40.52 & 63.67 & 77.25 & 63.17 & 67.66 & 59.23 \\
l-r-c  & 128 & 56.29 & 27.44 & 78.72 & 41.51 & 63.48 & 77.12 & 63.62 & 68.16 & 59.54 \\
l-r-c  & 256 & 57.29 & 24.39 & 78.64 & 41.32 & 63.87 & 77.25 & 63.47 & 68.26 & 59.31 \\
l-r-s           & 64  & 54.69 & 28.05 & 78.04 & 40.72 & 63.67 & 77.45 & 63.47 & 67.47 & 59.20 \\
l-r-s           & 128 & 55.89 & 26.83 & 78.64 & 39.92 & 61.88 & 76.85 & 63.27 & 69.66 & 59.12 \\
l-r-s           & 256 & 56.69 & 24.39 & 79.64 & 41.52 & 66.07 & 75.65 & 64.07 & 71.46 & 59.94 \\
\bottomrule
\end{tabular}
\vspace{-0.15in}
}
\end{table}

%% file: tables/ablation/rebuttal_calibration.tex
\begin{table}[h]
\centering
\caption{Sensitivity of Ours$_Q$ to the choice of calibration corpus on Qwen1.5-MoE-A2.7B under \TypeP[25\%]~\TypeQ[4b].}
\label{tab:app:calibration_sensitivity}
\resizebox{0.8\textwidth}{!}{
\begin{tabular}{lccccccc}
\toprule
\textbf{Calibration data} & \textbf{HellaSwag} & \textbf{WinoGrande} & \textbf{PIQA} & \textbf{ARC-c} & \textbf{MMLU} & \textbf{GSM8K} & \textbf{HumanEval} \\
\midrule
WikiText2          & 57.83 & \textbf{70.26} & 74.85 & 36.11 & 51.59 & 27.15 & 2.44 \\
C4                 & \textbf{61.28} & 69.26 & 74.45 & 35.94 & 52.46 & 9.18 & 3.05 \\
Pile               & 58.68 & 65.27 & 75.65 & \textbf{39.92} & \textbf{54.18} & 42.51 & 16.46 \\
GSM8K (train)      & 61.22 & 65.45 & \textbf{76.05} & 37.13 & 49.35 & \textbf{61.08} & 15.85 \\
OpenCodeReasoning  & 58.25 & 62.28 & 74.85 & 33.95 & 52.56 & 54.49 & \textbf{29.27} \\
RedPajama          & 60.61 & 65.88 & 73.65 & 36.51 & 52.43 & 23.35 & 3.05 \\
\bottomrule
\end{tabular}
}
\end{table}

%% file: tables/ablation/rebuttal_routing.tex
\begin{table}[t]
\centering
\caption{Evaluation under different top-$k$ routing strategies for Qwen1.5-MoE-A2.7B and DeepSeek-V2-Lite, with and without 50\% structural pruning.}
\label{tab:app:routing_topk}
\resizebox{1.0\textwidth}{!}{
\begin{tabular}{lcccccccccc c}
\toprule
\textbf{Model} & \textbf{Top-$k$} & \textbf{Prune (\%)} & \textbf{PIQA} & \textbf{ARC-c} & \textbf{ARC-e} & \textbf{BoolQ} & \textbf{HellaSwag} & \textbf{WinoGrande} & \textbf{GSM8K} & \textbf{HumanEval} & \textbf{Avg} \\
\midrule
\multirow{6}{*}{Qwen1.5-MoE-A2.7B}
 & 4 &  0 & 80.79 & 40.41 & 69.44 & 70.57 & 77.17 & 69.77 & 61.50 & 34.20 & 62.98 \\
 & 4 & 50 & 74.45 & 35.93 & 66.07 & 70.46 & 61.28 & 69.26 & 58.24 & 30.50 & 58.27 \\
 & 2 &  0 & 83.03 & 42.12 & 65.27 & 77.45 & 64.87 & 65.07 & 52.50 & 34.15 & 60.56 \\
 & 2 & 50 & 72.65 & 35.13 & 63.27 & 66.67 & 60.28 & 67.66 & 50.50 & 25.00 & 55.15 \\
 & 1 &  0 & 79.24 & 40.12 & 66.87 & 69.26 & 61.08 & 65.07 & 36.93 & 20.12 & 54.84 \\
 & 1 & 50 & 71.86 & 33.53 & 59.28 & 66.07 & 56.09 & 63.07 & 29.74 & 21.34 & 50.12 \\
\midrule
\multirow{6}{*}{DeepSeek-V2-Lite}
 & 6 &  0 & 80.20 & 46.93 & 78.37 & 79.82 & 77.98 & 71.35 & 30.90 & 32.30 & 62.23 \\
 & 6 & 50 & 78.40 & 42.49 & 73.78 & 71.77 & 74.59 & 69.22 & 33.70 & 28.70 & 59.08 \\
 & 2 &  0 & 79.45 & 42.66 & 68.30 & 75.15 & 62.82 & 65.56 & 25.64 & 25.61 & 55.65 \\
 & 2 & 50 & 75.34 & 35.42 & 61.25 & 65.36 & 57.34 & 63.01 & 24.85 & 18.29 & 50.11 \\
 & 1 &  0 & 73.65 & 33.53 & 61.88 & 65.07 & 54.69 & 60.68 &  6.39 & 10.98 & 45.86 \\
 & 1 & 50 & 71.06 & 29.74 & 56.49 & 61.48 & 52.89 & 57.29 &  8.38 &  7.93 & 43.16 \\
\bottomrule
\end{tabular}
}
\end{table}

%% file: figures/7_qwen1.5-stacked-bar.tex
\begin{figure}[ht]
  \begin{center}
    \centerline{\includegraphics[width=0.8\columnwidth]{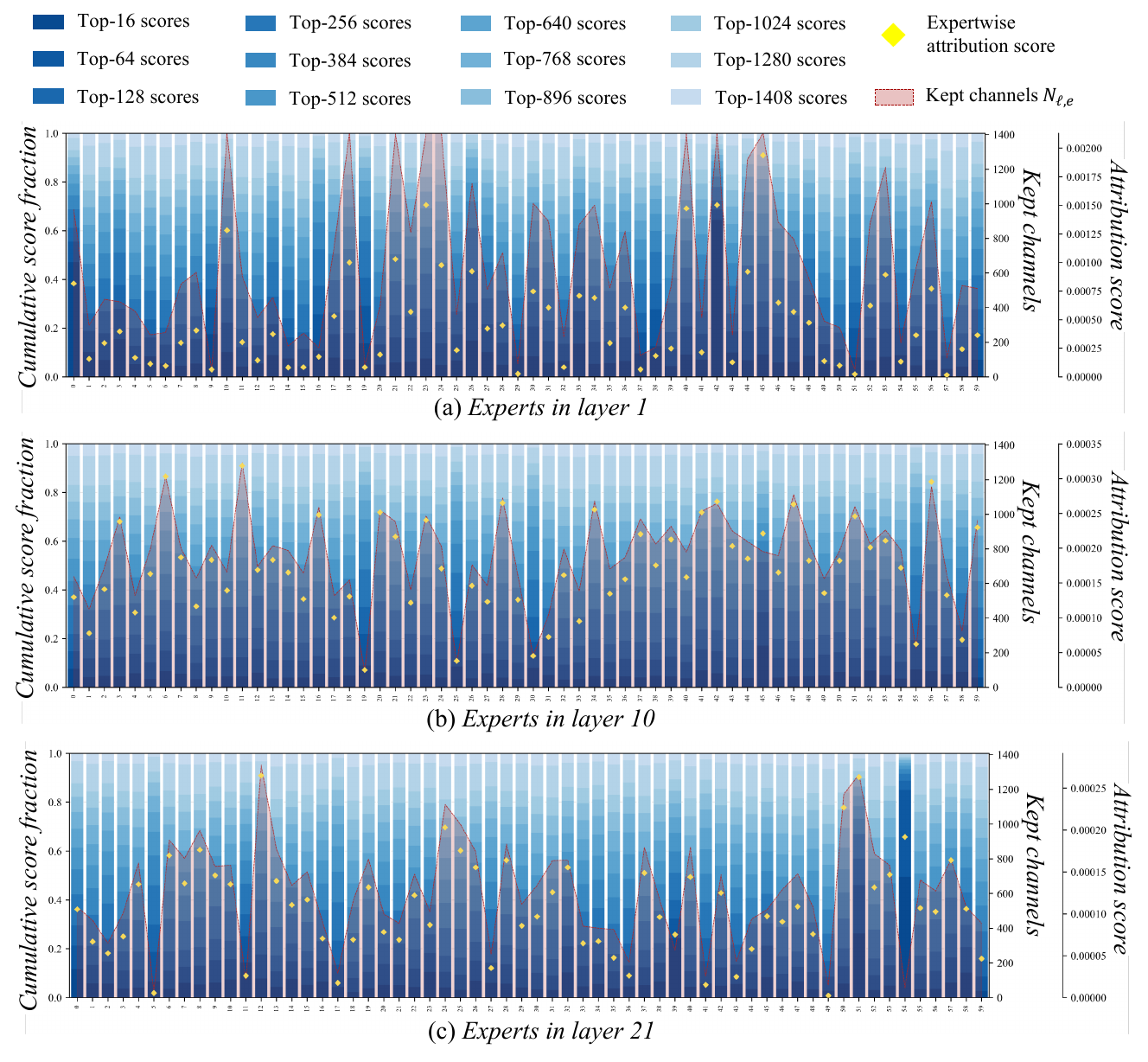}}
    \caption{Cumulative scores fraction (blue stacked bars), kepted channels (red lines) and attribution score (yellow diamond markers) for each expert in specific layers in Qwen1.5-MoE-A2.7B. }
    \vspace{-0.2in}
    \label{fig:qwen1.5-stacked-bar}
  \end{center}
\end{figure}

%% file: figures/8_stacked_bar-qwen3.tex
\begin{figure}[ht]
  \begin{center}
    \centerline{\includegraphics[width=\columnwidth]{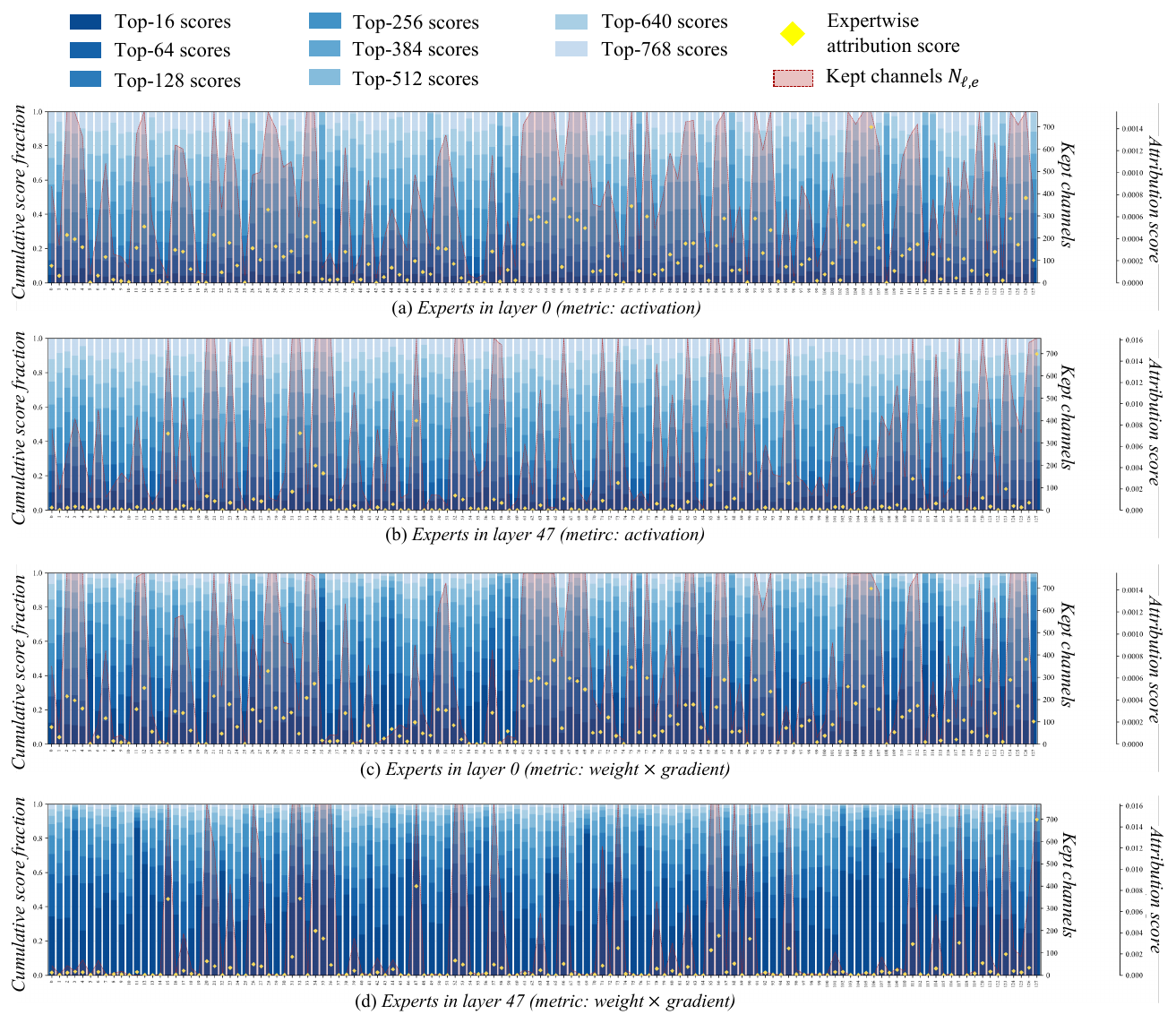}}
    \vspace{-0.05in}
    \caption{Cumulative channel-score fractions (blue stacked bars), kept channels (red lines), and attribution scores (yellow diamonds) for experts within a representative layer of Qwen3-30B-A3B. Panels (a) and (b) use activation-based scores, while (c) and (d) use weight gradient scores (weight times gradient). The channel-score concentration pattern appears under both metrics. Although experts exhibit substantial heterogeneity, our coverage-maximized budget allocation yields similar kept-channel allocations across metrics (red). }
    \vspace{-0.3in}
    \label{fig:qwen3-stacked-bar}
  \end{center}
\end{figure}

%% file: references.bib
@article{huang2025modes,
  title={MoDES: Accelerating Mixture-of-Experts Multimodal Large Language Models via Dynamic Expert Skipping},
  author={Huang, Yue and Wang, Zhiyuan and Yuan, Zhengrong and Ding, Yifu and Gong, Runpei and Guo, Jinyang and Liu, Xianglong and Zhang, Jie},
  journal={arXiv preprint arXiv:2511.15690},
  year={2025}
}

@inproceedings{guo2024joint,
  title={Compressing Large Language Models by Joint Sparsification and Quantization},
  author={Guo, Jinyang and Wu, Jing and Wang, Zhiyuan and Liu, Jiaheng and Yang, Ge and Ding, Yifu and Gong, Runpei and Qin, Haotong and Liu, Xianglong},
  booktitle={International Conference on Machine Learning},
  year={2024}
}

@article{gong2025lowbit,
  title={A survey of low-bit large language models: Basics, systems, and algorithms},
  author={Gong, Runpei and Ding, Yifu and Wang, Zhiyuan and Lv, Cheng and Zheng, Xingyu and Du, Jie and Yong, Yuanyuan and Gu, Shiming and Qin, Haotong and others},
  journal={Neural Networks},
  pages={107856},
  year={2025}
}

@inproceedings{lv2026llmcplus,
  title={{LLMC+}: Benchmarking Vision-Language Model Compression with a Plug-and-play Toolkit},
  author={Lv, Cheng and Zhang, Bei and Yong, Yuanyuan and Gong, Runpei and Huang, Yue and Gu, Shiming and Wu, Jing and Shi, Yuqiu and Guo, Jinyang and others},
  booktitle={AAAI Conference on Artificial Intelligence},
  year={2026}
}

@inproceedings{chen2024dbllm,
  title={{DB-LLM}: Accurate Dual-Binarization for Efficient {LLM}s},
  author={Chen, Hong and Lv, Cheng and Ding, Lei and Qin, Haotong and Zhou, Xudong and Ding, Yifu and Liu, Xianglong and Zhang, Mingyuan and Guo, Jinyang and others},
  booktitle={Findings of the Association for Computational Linguistics: ACL},
  year={2024}
}

@article{jiang2024mixtral,
  title={Mixtral of experts},
  author={Jiang, Albert Q and Sablayrolles, Alexandre and Roux, Antoine and Mensch, Arthur and Savary, Blanche and Bamford, Chris and Chaplot, Devendra Singh and Casas, Diego de las and Hanna, Emma Bou and Bressand, Florian and others},
  journal={arXiv preprint arXiv:2401.04088},
  year={2024}
}

@article{bai2025diep,
  title={DiEP: Adaptive Mixture-of-Experts Compression through Differentiable Expert Pruning},
  author={Bai, Sikai and Li, Haoxi and Zhang, Jie and Hong, Zicong and Guo, Song},
  journal={Advances in neural information processing systems},
  year={2025}
}

@inproceedings{yang2024moei2,
    title = "{M}o{E}-I$^2$: Compressing Mixture of Experts Models through Inter-Expert Pruning and Intra-Expert Low-Rank Decomposition",
    author = "Yang, Cheng  and       Sui, Yang  and       Xiao, Jinqi  and       Huang, Lingyi  and       Gong, Yu  and       Duan, Yuanlin  and       Jia, Wenqi  and       Yin, Miao  and       Cheng, Yu  and       Yuan, Bo",
    booktitle = "Findings of the Association for Computational Linguistics: EMNLP 2024",
    month = nov,
    year = "2024",
    address = "Miami, Florida, USA",
    publisher = "Association for Computational Linguistics",
    pages = "10456--10466",
}

@inproceedings{zhang2024diversify,
  author    = {Zeliang Zhang and Xiaodong Liu and Hao Cheng and Chenliang Xu and Jianfeng Gao},
  year      = {2024},
  title     = {Diversifying the Expert Knowledge for Task-Agnostic Pruning in Sparse Mixture-of-Experts},
  booktitle = {arXiv.org},
  doi       = {10.48550/arXiv.2407.09590},
}

@article{
  he2025towards,
  title={Towards Efficient Mixture of Experts: A Holistic Study of Compression Techniques},
  author={Shwai He and Daize Dong and Liang Ding and Ang Li},
  journal={Transactions on Machine Learning Research},
  issn={2835-8856},
  year={2025},
  url={https://openreview.net/forum?id=HTpMOl6xSI},
  note={}
}

@inproceedings{Lee2024STUNSP,
  title={Stun: Structured-then-unstructured pruning for scalable moe pruning},
  author={Lee, Jaeseong and Hwang, Seung-won and Qiao, Aurick and Campos, Daniel F and Yao, Zhewei and He, Yuxiong},
  booktitle={Proceedings of the 63rd Annual Meeting of the Association for Computational Linguistics (Volume 1: Long Papers)},
  pages={13660--13676},
  year={2025}
}

@inproceedings{xie2024moepruner,
  author    = {Yanyue Xie and Zhi Zhang and Ding Zhou and Cong Xie and Ziang Song and Xin Liu and Yanzhi Wang and Xue Lin and An Xu},
  year      = {2024},
  title     = {MoE-Pruner: Pruning Mixture-of-Experts Large Language Model using the Hints from Its Router},
  booktitle = {arXiv.org},
  doi       = {10.48550/arXiv.2410.12013},
}

@inproceedings{muzio2024seermoe,
  author    = {Alexandre Muzio and Alex Sun and Churan He},
  year      = {2024},
  title     = {SEER-MoE: Sparse Expert Efficiency through Regularization for Mixture-of-Experts},
  booktitle = {arXiv.org},
  doi       = {10.48550/arXiv.2404.05089},
}

@inproceedings{chowdhury2024aprovably,
  author    = {Mohammed Nowaz Rabbani Chowdhury and Meng Wang and K. E. Maghraoui and Naigang Wang and Pin-Yu Chen and Christopher Carothers},
  year      = {2024},
  title     = {A Provably Effective Method for Pruning Experts in Fine-tuned Sparse Mixture-of-Experts},
  booktitle = {International Conference on Machine Learning},
  doi       = {10.48550/arXiv.2405.16646},
}

@inproceedings{dong2025domain,
  author    = {Zican Dong and Han Peng and Peiyu Liu and Wayne Xin Zhao and Dong Wu and Feng Xiao and Zhifeng Wang},
  year      = {2025},
  title     = {Domain-Specific Pruning of Large Mixture-of-Experts Models with Few-shot Demonstrations},
}

@Inproceedings{Li2025SubMoEEM,
 author = {Lujun Li and Qiyuan Zhu and Jiacheng Wang and Wei Li and Hao Gu and Sirui Han and Yike Guo},
 booktitle = {arXiv.org},
 title = {Sub-MoE: Efficient Mixture-of-Expert LLMs Compression via Subspace Expert Merging},
 year = {2025}
}

@inproceedings{lu2024notall,
  author    = {Xudong Lu and Qi Liu and Yuhui Xu and Aojun Zhou and Siyuan Huang and Bo Zhang and Junchi Yan and Hongsheng Li},
  year      = {2024},
  title     = {Not All Experts are Equal: Efficient Expert Pruning and Skipping for Mixture-of-Experts Large Language Models},
  booktitle = {Annual Meeting of the Association for Computational Linguistics},
  doi       = {10.48550/arXiv.2402.14800},
}

@inproceedings{mc-moe,
  author    = {Wei Huang and Yue Liao and Jianhui Liu and Ruifei He and Haoru Tan and Shiming Zhang and Hongsheng Li and Si Liu and Xiaojuan Qi},
  year      = {2024},
  title     = {MC-MoE: Mixture Compressor for Mixture-of-Experts LLMs Gains More},
  booktitle = {arXiv.org},
  doi       = {10.48550/arXiv.2410.06270},
}

@Inproceedings{Chen2025CollaborativeCF,
 author = {Yixiao Chen and Yanyue Xie and Ruining Yang and Wei Jiang and Wei Wang and Yong He and Yue Chen and Pu Zhao and Yanzhi Wang},
 title = {Collaborative Compression for Large-Scale MoE Deployment on Edge},
 year = {2025}
}

@Inproceedings{Chen2025EACMoEEA,
 author = {Yuanteng Chen and Yuantian Shao and Peisong Wang and Jian Cheng},
 booktitle = {Annual Meeting of the Association for Computational Linguistics},
 title = {EAC-MoE: Expert-Selection Aware Compressor for Mixture-of-Experts Large Language Models},
 year = {2025}
}

@Inproceedings{Zhang2025MoNERR,
 author = {Geng Zhang and Yuxuan Han and Yuxuan Lou and Wangbo Zhao and Yiqi Zhang and Yang You},
 title = {MoNE: Replacing Redundant Experts with Lightweight Novices for Structured Pruning of MoE},
 year = {2025}
}

@article{zhao2025puzzlemoe,
  title={PuzzleMoE: Efficient Compression of Large Mixture-of-Experts Models via Sparse Expert Merging and Bit-packed inference},
  author={Zhao, Yushu and Wang, Zheng and Zhang, Minjia},
  journal={arXiv preprint arXiv:2511.04805},
  year={2025}
}

@article{c-prune,
  title={Cluster-Driven Expert Pruning for Mixture-of-Experts Large Language Models},
  author={Guo, Hongcheng and Yao, Juntao and Wang, Boyang and Du, Junjia and Cao, Shaosheng and Di, Donglin and Zhang, Shun and Li, Zhoujun},
  journal={arXiv preprint arXiv:2504.07807},
  year={2025}
}

@inproceedings{liu2024eep,
  author    = {Enhao Liu and Junyi Zhu and Zinan Lin and Xuefei Ning and Matthew B. Blaschko and Shengen Yan and Guohao Dai and Huazhong Yang and Yu Wang},
  year      = {2024},
  title     = {Efficient Expert Pruning for Sparse Mixture-of-Experts Language Models: Enhancing Performance and Reducing Inference Costs},
  booktitle = {arXiv.org},
  doi       = {10.48550/arXiv.2407.00945},
}

@misc{qwen3technicalreport,
      title={Qwen3 Technical Report}, 
      author={Qwen-Team},
      year={2025},
      eprint={2505.09388},
      archivePrefix={arXiv},
      primaryClass={cs.CL},
      url={https://arxiv.org/abs/2505.09388}, 
}

@Inproceedings{arc_c,
 author = {Peter Clark and Isaac Cowhey and Oren Etzioni and Tushar Khot and Ashish Sabharwal and Carissa Schoenick and Oyvind Tafjord},
 booktitle = {arXiv.org},
 title = {Think you have Solved Question Answering? Try ARC, the AI2 Reasoning Challenge},
 year = {2018}
}

@Inproceedings{hellaswag,
 author = {Rowan Zellers and Ari Holtzman and Yonatan Bisk and Ali Farhadi and Yejin Choi},
 booktitle = {Annual Meeting of the Association for Computational Linguistics},
 pages = {4791-4800},
 title = {HellaSwag: Can a Machine Really Finish Your Sentence?},
 year = {2019}
}

@Inproceedings{piqa,
 author = {Yonatan Bisk and Rowan Zellers and Ronan Le Bras and Jianfeng Gao and Yejin Choi},
 booktitle = {AAAI Conference on Artificial Intelligence},
 title = {PIQA: Reasoning about Physical Commonsense in Natural Language},
 year = {2019}
}

@Inproceedings{boolq,
 author = {Christopher Clark and Kenton Lee and Ming-Wei Chang and T. Kwiatkowski and Michael Collins and Kristina Toutanova},
 booktitle = {North American Chapter of the Association for Computational Linguistics},
 title = {BoolQ: Exploring the Surprising Difficulty of Natural Yes/No Questions},
 year = {2019}
}

@Article{mmlu,
 author = {Dan Hendrycks and Collin Burns and Steven Basart and Andy Zou and Mantas Mazeika and D. Song and J. Steinhardt},
 booktitle = {International Conference on Learning Representations},
 journal = {ArXiv},
 title = {Measuring Massive Multitask Language Understanding},
 volume = {abs/2009.03300},
 year = {2020}
}

@article{winogrande,
  author = {Keisuke Sakaguchi and Ronan Le Bras and Chandra Bhagavatula and Yejin Choi},
  year = {undefined},
  title = {WinoGrande: An Adversarial Winograd Schema Challenge at Scale},
  journal = {Proceedings of the AAAI Conference on Artificial Intelligence},
}

@misc{qwen1.5-moe,
    title = {Qwen1.5-MoE: Matching 7B Model Performance with 1/3 Activated Parameters},
    url = {https://qwenlm.github.io/blog/qwen-moe/},
    author = {Qwen Team},
    month = {February},
    year = {2024}
}

@Article{deepseek-moe,
 author = {Damai Dai and Chengqi Deng and Chenggang Zhao and R. Xu and Huazuo Gao and Deli Chen and Jiashi Li and Wangding Zeng and Xingkai Yu and Yu Wu and Zhenda Xie and Y. K. Li and Panpan Huang and Fuli Luo and Chong Ruan and Zhifang Sui and W. Liang},
 booktitle = {Annual Meeting of the Association for Computational Linguistics},
 pages = {1280-1297},
 title = {DeepSeekMoE: Towards Ultimate Expert Specialization in Mixture-of-Experts Language Models},
 year = {2024}
}

@misc{deepseekv2,
      title={DeepSeek-V2: A Strong, Economical, and Efficient Mixture-of-Experts Language Model}, 
      author={DeepSeek-AI},
      year={2024},
      eprint={2405.04434},
      archivePrefix={arXiv},
      primaryClass={cs.CL}
}

@inproceedings{gsm8k,
  author    = {K. Cobbe and Vineet Kosaraju and Mo Bavarian and Mark Chen and Heewoo Jun and Lukasz Kaiser and Matthias Plappert and Jerry Tworek and Jacob Hilton and Reiichiro Nakano and Christopher Hesse and John Schulman},
  year      = {2021},
  title     = {Training Verifiers to Solve Math Word Problems},
  booktitle = {arXiv.org},
}

@article{humaneval,
  title={Evaluating Large Language Models Trained on Code},
  author={Mark Chen and Jerry Tworek and Heewoo Jun and Qiming Yuan and Henrique Ponde de Oliveira Pinto and Jared Kaplan and Harri Edwards and Yuri Burda and Nicholas Joseph and Greg Brockman and Alex Ray and Raul Puri and Gretchen Krueger and Michael Petrov and Heidy Khlaaf and Girish Sastry and Pamela Mishkin and Brooke Chan and Scott Gray and Nick Ryder and Mikhail Pavlov and Alethea Power and Lukasz Kaiser and Mohammad Bavarian and Clemens Winter and Philippe Tillet and Felipe Petroski Such and Dave Cummings and Matthias Plappert and Fotios Chantzis and Elizabeth Barnes and Ariel Herbert-Voss and William Hebgen Guss and Alex Nichol and Alex Paino and Nikolas Tezak and Jie Tang and Igor Babuschkin and Suchir Balaji and Shantanu Jain and William Saunders and Christopher Hesse and Andrew N. Carr and Jan Leike and Josh Achiam and Vedant Misra and Evan Morikawa and Alec Radford and Matthew Knight and Miles Brundage and Mira Murati and Katie Mayer and Peter Welinder and Bob McGrew and Dario Amodei and Sam McCandlish and Ilya Sutskever and Wojciech Zaremba},
  year={2021},
  eprint={2107.03374},
  archivePrefix={arXiv},
  primaryClass={cs.LG}
}

@article{jain2024livecodebench,
  title={Livecodebench: Holistic and contamination free evaluation of large language models for code},
  author={Jain, Naman and Han, King and Gu, Alex and Li, Wen-Ding and Yan, Fanjia and Zhang, Tianjun and Wang, Sida and Solar-Lezama, Armando and Sen, Koushik and Stoica, Ion},
  journal={arXiv preprint arXiv:2403.07974},
  year={2024}
}

@misc{alpaca,
  author = {Rohan Taori and Ishaan Gulrajani and Tianyi Zhang and Yann Dubois and Xuechen Li and Carlos Guestrin and Percy Liang and Tatsunori B. Hashimoto },
  title = {Stanford Alpaca: An Instruction-following LLaMA model},
  year = {2023},
  publisher = {GitHub},
  journal = {GitHub repository},
  howpublished = {\url{https://github.com/tatsu-lab/stanford_alpaca}},
}

@Article{c4,
 author = {Colin Raffel and Noam M. Shazeer and Adam Roberts and Katherine Lee and Sharan Narang and Michael Matena and Yanqi Zhou and Wei Li and Peter J. Liu},
 booktitle = {Journal of machine learning research},
 journal = {J. Mach. Learn. Res.},
 pages = {140:1-140:67},
 title = {Exploring the Limits of Transfer Learning with a Unified Text-to-Text Transformer},
 volume = {21},
 year = {2019}
}

@article{ahmad2025opencodereasoning,
  title={Opencodereasoning: Advancing data distillation for competitive coding},
  author={Ahmad, Wasi Uddin and Narenthiran, Sean and Majumdar, Somshubra and Ficek, Aleksander and Jain, Siddhartha and Huang, Jocelyn and Noroozi, Vahid and Ginsburg, Boris},
  journal={arXiv preprint arXiv:2504.01943},
  year={2025}
}

@article{liu2024dora,
  title={DoRA: Weight-Decomposed Low-Rank Adaptation},
  author={Liu, Shih-Yang and Wang, Chien-Yi and Yin, Hongxu and Molchanov, Pavlo and Wang, Yu-Chiang Frank and Cheng, Kwang-Ting and Chen, Min-Hung},
  journal={arXiv preprint arXiv:2402.09353},
  year={2024}
}

@article{math500,
      title={Let's Verify Step by Step}, 
      author={Lightman, Hunter and Kosaraju, Vineet and Burda, Yura and Edwards, Harri and Baker, Bowen and Lee, Teddy and Leike, Jan and Schulman, John and Sutskever, Ilya and Cobbe, Karl},
      journal={arXiv preprint arXiv:2305.20050},
      year={2023}
}

@article{song2019attribution,
  title={Deep model transferability from attribution maps},
  author={Song, Jie and Chen, Yixin and Wang, Xinchao and Shen, Chengchao and Song, Mingli},
  journal={Advances in Neural Information Processing Systems},
  volume={32},
  year={2019}
}

@inproceedings{snip,
  author    = {Namhoon Lee and Thalaiyasingam Ajanthan and Philip H. S. Torr},
  year      = {2018},
  title     = {SNIP: Single-shot Network Pruning based on Connection Sensitivity},
  booktitle = {International Conference on Learning Representations},
}

@article{wanda,
  title={A simple and effective pruning approach for large language models},
  author={Sun, Mingjie and Liu, Zhuang and Bair, Anna and Kolter, J Zico},
  journal={arXiv preprint arXiv:2306.11695},
  year={2023}
}

@article{skean2025layer,
  title={Layer by layer: Uncovering hidden representations in language models},
  author={Skean, Oscar and Arefin, Md Rifat and Zhao, Dan and Patel, Niket and Naghiyev, Jalal and LeCun, Yann and Shwartz-Ziv, Ravid},
  journal={arXiv preprint arXiv:2502.02013},
  year={2025}
}

@article{deepseek-r1,
  title={Deepseek-r1: Incentivizing reasoning capability in llms via reinforcement learning},
  author={Guo, Daya and Yang, Dejian and Zhang, Haowei and Song, Junxiao and Zhang, Ruoyu and Xu, Runxin and Zhu, Qihao and Ma, Shirong and Wang, Peiyi and Bi, Xiao and others},
  journal={arXiv preprint arXiv:2501.12948},
  year={2025}
}

@article{openmoe,
  title={OpenMoE: An Early Effort on Open Mixture-of-Experts Language Models},
  author={Xue, Fuzhao and Zheng, Zian and Fu, Yao and Ni, Jinjie and Zheng, Zangwei and Zhou, Wangchunshu and You, Yang},
  journal={arXiv preprint arXiv:2402.01739},
  year={2024}
}

@article{llm-pruner,
  title={Llm-pruner: On the structural pruning of large language models},
  author={Ma, Xinyin and Fang, Gongfan and Wang, Xinchao},
  journal={Advances in neural information processing systems},
  volume={36},
  pages={21702--21720},
  year={2023}
}

@inproceedings{disp-llm,
  author    = {Shangqian Gao and Chi-Heng Lin and Ting Hua and Tang Zheng and Yilin Shen and Hongxia Jin and Yen-Chang Hsu},
  year      = {2024},
  title     = {DISP-LLM: Dimension-Independent Structural Pruning for Large Language Models},
  booktitle = {Neural Information Processing Systems},
  doi       = {10.48550/arXiv.2410.11988},
}

@inproceedings{an2024fluctuation,
  title={Fluctuation-based adaptive structured pruning for large language models},
  author={An, Yongqi and Zhao, Xu and Yu, Tao and Tang, Ming and Wang, Jinqiao},
  booktitle={Proceedings of the AAAI Conference on Artificial Intelligence},
  volume={38},
  number={10},
  pages={10865--10873},
  year={2024}
}

@article{cao2024condense,
  title={Condense, Don't Just Prune: Enhancing Efficiency and Performance in MoE Layer Pruning},
  author={Cao, Mingyu and Li, Gen and Ji, Jie and Zhang, Jiaqi and Ma, Xiaolong and Liu, Shiwei and Yin, Lu},
  journal={arXiv preprint arXiv:2412.00069},
  year={2024}
}

@misc{anon_icml2026_hidden_intermediate_prune_distill,
  title        = {Orchestrating Hidden-Intermediate Pruning-and-Distill for MoEs Slimming},
  author={Anonymous}, 
  howpublished = {Anonymous ICML 2026 submission (under review)},
  year         = {2026},
  note         = {Anonymized concurrent submission. The anonymized PDF is provided in the supplementary material.}
}

@misc{aime25,
      title={American Invitational Mathematics Examination (AIME) 2025}, 
      author={Zhang, Yifan and Math-AI, Team},
      year={2025},
}

@misc{gpqa,
      title={GPQA: A Graduate-Level Google-Proof Q\&A Benchmark}, 
      author={David Rein and Betty Li Hou and Asa Cooper Stickland and Jackson Petty and Richard Yuanzhe Pang and Julien Dirani and Julian Michael and Samuel R. Bowman},
      year={2023},
      eprint={2311.12022},
      archivePrefix={arXiv},
      primaryClass={cs.AI},
      url={https://arxiv.org/abs/2311.12022}, 
}

@misc{xu2025mcmoe,
      title={MCMoE: Completing Missing Modalities with Mixture of Experts for Incomplete Multimodal Action Quality Assessment}, 
      author={Huangbiao Xu and Huanqi Wu and Xiao Ke and Junyi Wu and Rui Xu and Jinglin Xu},
      year={2025},
      eprint={2511.17397},
      archivePrefix={arXiv},
      primaryClass={cs.CV},
      url={https://arxiv.org/abs/2511.17397}, 
}
